\documentclass[12pt]{article}
\usepackage[english]{babel}
\usepackage[utf8]{inputenc}
\usepackage[T1]{fontenc}
\usepackage{lmodern}

\usepackage{indentfirst}

\usepackage{chngcntr}
\counterwithin{figure}{section}

\usepackage{amsmath, amssymb, amsthm}
\usepackage{mathtools}
\usepackage{wrapfig}
\usepackage[margin=1in]{geometry} 
\usepackage{setspace}
\usepackage{graphicx}
\usepackage{array}
\usepackage{caption}
\usepackage{subcaption}
\usepackage{tabularx}
\usepackage{booktabs}
\usepackage{threeparttable}
\usepackage{amssymb}
\usepackage{comment}
\usepackage{multirow}

\usepackage{titling}
\usepackage{authblk} 
\usepackage{pdfpages}
\RequirePackage[mathlines]{lineno}

\usepackage{url}
\usepackage{hyperref}
\usepackage{xcolor}
\hypersetup{
    colorlinks=true,
    linkcolor=purple,    
    urlcolor=purple,
    citecolor=blue
}

\newtheorem{remark}{Remark}

\usepackage[round]{natbib}
\providecommand{\Result}[1]{%
  \ifcsname paperresult@#1\endcsname%
    \csname paperresult@#1\endcsname%
  \else%
    \PackageError{paper-results}{Unknown result key #1}{Regenerate results.tex.}%
  \fi}
\expandafter\def\csname paperresult@run.id\endcsname{sp\_full}
\expandafter\def\csname paperresult@sample.n_pages\endcsname{5,118,347}
\expandafter\def\csname paperresult@sample.n_domains\endcsname{505,974}
\expandafter\def\csname paperresult@sample.n_crawls\endcsname{71}
\expandafter\def\csname paperresult@sample.first_crawl\endcsname{CC-MAIN-2019-04}
\expandafter\def\csname paperresult@sample.last_crawl\endcsname{CC-MAIN-2026-34}
\expandafter\def\csname paperresult@sample.start_date\endcsname{2019-01-15}
\expandafter\def\csname paperresult@sample.end_date\endcsname{2026-08-20}
\expandafter\def\csname paperresult@sample.date_range\endcsname{January 2019 to August 2026}
\expandafter\def\csname paperresult@outcome.definition\endcsname{fully AI-generated or AI-edited text}

\providecommand{\Result}[1]{%
  \ifcsname paperresult@#1\endcsname%
    \csname paperresult@#1\endcsname%
  \else%
    \PackageError{paper-results}{Unknown result key #1}{Regenerate results.tex.}%
  \fi}
\expandafter\def\csname paperresult@overall.first.mean\endcsname{1.2}
\expandafter\def\csname paperresult@overall.first.lo95\endcsname{1.0}
\expandafter\def\csname paperresult@overall.first.hi95\endcsname{1.3}
\expandafter\def\csname paperresult@overall.first.raw_rate\endcsname{6.9}
\expandafter\def\csname paperresult@overall.latest.mean\endcsname{23.6}
\expandafter\def\csname paperresult@overall.latest.lo95\endcsname{22.9}
\expandafter\def\csname paperresult@overall.latest.hi95\endcsname{24.4}
\expandafter\def\csname paperresult@overall.latest.raw_rate\endcsname{25.1}
\expandafter\def\csname paperresult@overall.rate.baseline.rate_pp_year\endcsname{0.2}
\expandafter\def\csname paperresult@overall.rate.baseline.lo95_pp_year\endcsname{0.2}
\expandafter\def\csname paperresult@overall.rate.baseline.hi95_pp_year\endcsname{0.2}
\expandafter\def\csname paperresult@overall.rate.baseline.start_date\endcsname{January 2019}
\expandafter\def\csname paperresult@overall.rate.baseline.end_date\endcsname{February 2023}
\expandafter\def\csname paperresult@overall.rate.early.rate_pp_year\endcsname{4.6}
\expandafter\def\csname paperresult@overall.rate.early.lo95_pp_year\endcsname{4.4}
\expandafter\def\csname paperresult@overall.rate.early.hi95_pp_year\endcsname{4.7}
\expandafter\def\csname paperresult@overall.rate.early.start_date\endcsname{February 2023}
\expandafter\def\csname paperresult@overall.rate.early.end_date\endcsname{January 2025}
\expandafter\def\csname paperresult@overall.rate.late.rate_pp_year\endcsname{8.1}
\expandafter\def\csname paperresult@overall.rate.late.lo95_pp_year\endcsname{7.8}
\expandafter\def\csname paperresult@overall.rate.late.hi95_pp_year\endcsname{8.5}
\expandafter\def\csname paperresult@overall.rate.late.start_date\endcsname{January 2025}
\expandafter\def\csname paperresult@overall.rate.late.end_date\endcsname{August 2026}
\expandafter\def\csname paperresult@overall.rate.difference.rate_difference_pp_year\endcsname{3.6}
\expandafter\def\csname paperresult@overall.rate.difference.lo95_pp_year\endcsname{3.3}
\expandafter\def\csname paperresult@overall.rate.difference.hi95_pp_year\endcsname{3.9}

\providecommand{\Result}[1]{%
  \ifcsname paperresult@#1\endcsname%
    \csname paperresult@#1\endcsname%
  \else%
    \PackageError{paper-results}{Unknown result key #1}{Regenerate results.tex.}%
  \fi}
\expandafter\def\csname paperresult@centrality.latest.top.page_share\endcsname{2.7}
\expandafter\def\csname paperresult@centrality.latest.top.mean\endcsname{11.6}
\expandafter\def\csname paperresult@centrality.latest.top.lo95\endcsname{11.2}
\expandafter\def\csname paperresult@centrality.latest.top.hi95\endcsname{11.9}
\expandafter\def\csname paperresult@centrality.latest.upper_mid.page_share\endcsname{3.6}
\expandafter\def\csname paperresult@centrality.latest.upper_mid.mean\endcsname{13.2}
\expandafter\def\csname paperresult@centrality.latest.upper_mid.lo95\endcsname{12.7}
\expandafter\def\csname paperresult@centrality.latest.upper_mid.hi95\endcsname{13.6}
\expandafter\def\csname paperresult@centrality.latest.mid.page_share\endcsname{8.5}
\expandafter\def\csname paperresult@centrality.latest.mid.mean\endcsname{16.6}
\expandafter\def\csname paperresult@centrality.latest.mid.lo95\endcsname{16.1}
\expandafter\def\csname paperresult@centrality.latest.mid.hi95\endcsname{17.1}
\expandafter\def\csname paperresult@centrality.latest.lower_mid.page_share\endcsname{14.5}
\expandafter\def\csname paperresult@centrality.latest.lower_mid.mean\endcsname{22.5}
\expandafter\def\csname paperresult@centrality.latest.lower_mid.lo95\endcsname{21.8}
\expandafter\def\csname paperresult@centrality.latest.lower_mid.hi95\endcsname{23.1}
\expandafter\def\csname paperresult@centrality.latest.tail.page_share\endcsname{29.8}
\expandafter\def\csname paperresult@centrality.latest.tail.mean\endcsname{23.3}
\expandafter\def\csname paperresult@centrality.latest.tail.lo95\endcsname{22.5}
\expandafter\def\csname paperresult@centrality.latest.tail.hi95\endcsname{24.1}
\expandafter\def\csname paperresult@centrality.latest.bottom.page_share\endcsname{40.9}
\expandafter\def\csname paperresult@centrality.latest.bottom.mean\endcsname{27.5}
\expandafter\def\csname paperresult@centrality.latest.bottom.lo95\endcsname{26.5}
\expandafter\def\csname paperresult@centrality.latest.bottom.hi95\endcsname{28.4}
\expandafter\def\csname paperresult@centrality.growth.top.weight_share_t1\endcsname{2.7}
\expandafter\def\csname paperresult@centrality.growth.top.domain_share\endcsname{0.001}
\expandafter\def\csname paperresult@centrality.growth.top.share_of_increase\endcsname{0.7}
\expandafter\def\csname paperresult@centrality.growth.top.share_of_increase_lo95\endcsname{0.6}
\expandafter\def\csname paperresult@centrality.growth.top.share_of_increase_hi95\endcsname{0.7}
\expandafter\def\csname paperresult@centrality.growth.top.excess_share_of_increase\endcsname{-2.1}
\expandafter\def\csname paperresult@centrality.growth.top.excess_share_of_increase_lo95\endcsname{-2.1}
\expandafter\def\csname paperresult@centrality.growth.top.excess_share_of_increase_hi95\endcsname{-2.0}
\expandafter\def\csname paperresult@centrality.growth.upper_mid.weight_share_t1\endcsname{3.6}
\expandafter\def\csname paperresult@centrality.growth.upper_mid.domain_share\endcsname{0.009}
\expandafter\def\csname paperresult@centrality.growth.upper_mid.share_of_increase\endcsname{1.5}
\expandafter\def\csname paperresult@centrality.growth.upper_mid.share_of_increase_lo95\endcsname{1.5}
\expandafter\def\csname paperresult@centrality.growth.upper_mid.share_of_increase_hi95\endcsname{1.6}
\expandafter\def\csname paperresult@centrality.growth.upper_mid.excess_share_of_increase\endcsname{-2.0}
\expandafter\def\csname paperresult@centrality.growth.upper_mid.excess_share_of_increase_lo95\endcsname{-2.1}
\expandafter\def\csname paperresult@centrality.growth.upper_mid.excess_share_of_increase_hi95\endcsname{-2.0}
\expandafter\def\csname paperresult@centrality.growth.mid.weight_share_t1\endcsname{8.5}
\expandafter\def\csname paperresult@centrality.growth.mid.domain_share\endcsname{0.09}
\expandafter\def\csname paperresult@centrality.growth.mid.share_of_increase\endcsname{5.4}
\expandafter\def\csname paperresult@centrality.growth.mid.share_of_increase_lo95\endcsname{5.2}
\expandafter\def\csname paperresult@centrality.growth.mid.share_of_increase_hi95\endcsname{5.5}
\expandafter\def\csname paperresult@centrality.growth.mid.excess_share_of_increase\endcsname{-3.2}
\expandafter\def\csname paperresult@centrality.growth.mid.excess_share_of_increase_lo95\endcsname{-3.3}
\expandafter\def\csname paperresult@centrality.growth.mid.excess_share_of_increase_hi95\endcsname{-3.1}
\expandafter\def\csname paperresult@centrality.growth.lower_mid.weight_share_t1\endcsname{14.6}
\expandafter\def\csname paperresult@centrality.growth.lower_mid.domain_share\endcsname{0.9}
\expandafter\def\csname paperresult@centrality.growth.lower_mid.share_of_increase\endcsname{13.6}
\expandafter\def\csname paperresult@centrality.growth.lower_mid.share_of_increase_lo95\endcsname{13.4}
\expandafter\def\csname paperresult@centrality.growth.lower_mid.share_of_increase_hi95\endcsname{13.8}
\expandafter\def\csname paperresult@centrality.growth.lower_mid.excess_share_of_increase\endcsname{-1.0}
\expandafter\def\csname paperresult@centrality.growth.lower_mid.excess_share_of_increase_lo95\endcsname{-1.2}
\expandafter\def\csname paperresult@centrality.growth.lower_mid.excess_share_of_increase_hi95\endcsname{-0.8}
\expandafter\def\csname paperresult@centrality.growth.tail.weight_share_t1\endcsname{29.9}
\expandafter\def\csname paperresult@centrality.growth.tail.domain_share\endcsname{9}
\expandafter\def\csname paperresult@centrality.growth.tail.share_of_increase\endcsname{30.2}
\expandafter\def\csname paperresult@centrality.growth.tail.share_of_increase_lo95\endcsname{29.8}
\expandafter\def\csname paperresult@centrality.growth.tail.share_of_increase_hi95\endcsname{30.6}
\expandafter\def\csname paperresult@centrality.growth.tail.excess_share_of_increase\endcsname{0.3}
\expandafter\def\csname paperresult@centrality.growth.tail.excess_share_of_increase_lo95\endcsname{-0.2}
\expandafter\def\csname paperresult@centrality.growth.tail.excess_share_of_increase_hi95\endcsname{0.7}
\expandafter\def\csname paperresult@centrality.growth.bottom.weight_share_t1\endcsname{40.6}
\expandafter\def\csname paperresult@centrality.growth.bottom.domain_share\endcsname{90}
\expandafter\def\csname paperresult@centrality.growth.bottom.share_of_increase\endcsname{48.6}
\expandafter\def\csname paperresult@centrality.growth.bottom.share_of_increase_lo95\endcsname{48.1}
\expandafter\def\csname paperresult@centrality.growth.bottom.share_of_increase_hi95\endcsname{49.2}
\expandafter\def\csname paperresult@centrality.growth.bottom.excess_share_of_increase\endcsname{8.0}
\expandafter\def\csname paperresult@centrality.growth.bottom.excess_share_of_increase_lo95\endcsname{7.5}
\expandafter\def\csname paperresult@centrality.growth.bottom.excess_share_of_increase_hi95\endcsname{8.6}
\expandafter\def\csname paperresult@centrality.growth.lowest_two.weight_share_t1\endcsname{70.5}
\expandafter\def\csname paperresult@centrality.growth.lowest_two.share_of_increase\endcsname{78.8}
\expandafter\def\csname paperresult@centrality.growth.lowest_two.excess_share_of_increase\endcsname{8.3}
\expandafter\def\csname paperresult@centrality.growth.baseline_date\endcsname{December 2021}
\expandafter\def\csname paperresult@centrality.growth.latest_date\endcsname{August 2026}
\expandafter\def\csname paperresult@centrality.growth.lowest_two.absolute_excess\endcsname{8.3}

\providecommand{\Result}[1]{%
  \ifcsname paperresult@#1\endcsname%
    \csname paperresult@#1\endcsname%
  \else%
    \PackageError{paper-results}{Unknown result key #1}{Regenerate results.tex.}%
  \fi}
\expandafter\def\csname paperresult@centrality.decomp.overall0\endcsname{1.9}
\expandafter\def\csname paperresult@centrality.decomp.overall1\endcsname{23.6}
\expandafter\def\csname paperresult@centrality.decomp.overall_change\endcsname{21.7}
\expandafter\def\csname paperresult@centrality.decomp.total_within\endcsname{21.2}
\expandafter\def\csname paperresult@centrality.decomp.total_composition\endcsname{0.5}
\expandafter\def\csname paperresult@centrality.decomp.low_contribution\endcsname{17.1}
\expandafter\def\csname paperresult@centrality.decomp.low_share_of_increase\endcsname{78.8}
\expandafter\def\csname paperresult@centrality.decomp.low_within\endcsname{15.7}
\expandafter\def\csname paperresult@centrality.decomp.low_composition\endcsname{1.4}
\expandafter\def\csname paperresult@centrality.decomp.low_within_share_of_overall\endcsname{72.3}
\expandafter\def\csname paperresult@centrality.decomp.average_low_page_share\endcsname{65.3}
\expandafter\def\csname paperresult@centrality.decomp.low_within_rate\endcsname{24.0}
\expandafter\def\csname paperresult@centrality.decomp.other_within_rate\endcsname{16.0}
\expandafter\def\csname paperresult@centrality.decomp.within_rate_difference\endcsname{8.0}
\expandafter\def\csname paperresult@centrality.decomp.low_share_of_within\endcsname{73.9}
\expandafter\def\csname paperresult@centrality.decomp.low_page_share0\endcsname{60.2}
\expandafter\def\csname paperresult@centrality.decomp.low_page_share1\endcsname{70.5}
\expandafter\def\csname paperresult@centrality.decomp.low_share_export_mean\endcsname{78.8}
\expandafter\def\csname paperresult@centrality.decomp.low_share_conservative_lower\endcsname{77.9}
\expandafter\def\csname paperresult@centrality.decomp.baseline_date\endcsname{December 2021}
\expandafter\def\csname paperresult@centrality.decomp.latest_date\endcsname{August 2026}

\providecommand{\Result}[1]{%
  \ifcsname paperresult@#1\endcsname%
    \csname paperresult@#1\endcsname%
  \else%
    \PackageError{paper-results}{Unknown result key #1}{Regenerate results.tex.}%
  \fi}
\expandafter\def\csname paperresult@domain.latest.com.mean\endcsname{29.5}
\expandafter\def\csname paperresult@domain.latest.com.lo95\endcsname{28.7}
\expandafter\def\csname paperresult@domain.latest.com.hi95\endcsname{30.2}
\expandafter\def\csname paperresult@domain.latest.net.mean\endcsname{20.4}
\expandafter\def\csname paperresult@domain.latest.net.lo95\endcsname{19.5}
\expandafter\def\csname paperresult@domain.latest.net.hi95\endcsname{21.4}
\expandafter\def\csname paperresult@domain.latest.other.mean\endcsname{20.2}
\expandafter\def\csname paperresult@domain.latest.other.lo95\endcsname{19.3}
\expandafter\def\csname paperresult@domain.latest.other.hi95\endcsname{21.0}
\expandafter\def\csname paperresult@domain.latest.org.mean\endcsname{14.6}
\expandafter\def\csname paperresult@domain.latest.org.lo95\endcsname{14.1}
\expandafter\def\csname paperresult@domain.latest.org.hi95\endcsname{15.0}
\expandafter\def\csname paperresult@domain.latest.gov.mean\endcsname{5.8}
\expandafter\def\csname paperresult@domain.latest.gov.lo95\endcsname{5.1}
\expandafter\def\csname paperresult@domain.latest.gov.hi95\endcsname{6.5}
\expandafter\def\csname paperresult@domain.latest.edu.mean\endcsname{6.1}
\expandafter\def\csname paperresult@domain.latest.edu.lo95\endcsname{5.4}
\expandafter\def\csname paperresult@domain.latest.edu.hi95\endcsname{6.9}
\expandafter\def\csname paperresult@domain.baseline.com.mean\endcsname{2.4}
\expandafter\def\csname paperresult@domain.baseline.com.lo95\endcsname{2.2}
\expandafter\def\csname paperresult@domain.baseline.com.hi95\endcsname{2.5}
\expandafter\def\csname paperresult@domain.baseline.net.mean\endcsname{1.5}
\expandafter\def\csname paperresult@domain.baseline.net.lo95\endcsname{1.4}
\expandafter\def\csname paperresult@domain.baseline.net.hi95\endcsname{1.6}
\expandafter\def\csname paperresult@domain.baseline.other.mean\endcsname{1.6}
\expandafter\def\csname paperresult@domain.baseline.other.lo95\endcsname{1.4}
\expandafter\def\csname paperresult@domain.baseline.other.hi95\endcsname{1.7}
\expandafter\def\csname paperresult@domain.baseline.org.mean\endcsname{1.8}
\expandafter\def\csname paperresult@domain.baseline.org.lo95\endcsname{1.7}
\expandafter\def\csname paperresult@domain.baseline.org.hi95\endcsname{1.9}
\expandafter\def\csname paperresult@domain.baseline.gov.mean\endcsname{1.8}
\expandafter\def\csname paperresult@domain.baseline.gov.lo95\endcsname{1.6}
\expandafter\def\csname paperresult@domain.baseline.gov.hi95\endcsname{2.0}
\expandafter\def\csname paperresult@domain.baseline.edu.mean\endcsname{1.2}
\expandafter\def\csname paperresult@domain.baseline.edu.lo95\endcsname{1.1}
\expandafter\def\csname paperresult@domain.baseline.edu.hi95\endcsname{1.4}
\expandafter\def\csname paperresult@domain.growth.com.mean\endcsname{27.1}
\expandafter\def\csname paperresult@domain.growth.com.lo95\endcsname{26.4}
\expandafter\def\csname paperresult@domain.growth.com.hi95\endcsname{27.8}
\expandafter\def\csname paperresult@domain.growth.net.mean\endcsname{18.9}
\expandafter\def\csname paperresult@domain.growth.net.lo95\endcsname{17.9}
\expandafter\def\csname paperresult@domain.growth.net.hi95\endcsname{19.9}
\expandafter\def\csname paperresult@domain.growth.other.mean\endcsname{18.6}
\expandafter\def\csname paperresult@domain.growth.other.lo95\endcsname{17.8}
\expandafter\def\csname paperresult@domain.growth.other.hi95\endcsname{19.4}
\expandafter\def\csname paperresult@domain.growth.org.mean\endcsname{12.8}
\expandafter\def\csname paperresult@domain.growth.org.lo95\endcsname{12.3}
\expandafter\def\csname paperresult@domain.growth.org.hi95\endcsname{13.2}
\expandafter\def\csname paperresult@domain.growth.gov.mean\endcsname{4.0}
\expandafter\def\csname paperresult@domain.growth.gov.lo95\endcsname{3.3}
\expandafter\def\csname paperresult@domain.growth.gov.hi95\endcsname{4.7}
\expandafter\def\csname paperresult@domain.growth.edu.mean\endcsname{4.9}
\expandafter\def\csname paperresult@domain.growth.edu.lo95\endcsname{4.2}
\expandafter\def\csname paperresult@domain.growth.edu.hi95\endcsname{5.7}
\expandafter\def\csname paperresult@domain.standardized.com.mean\endcsname{26.4}
\expandafter\def\csname paperresult@domain.standardized.com.lo95\endcsname{25.6}
\expandafter\def\csname paperresult@domain.standardized.com.hi95\endcsname{27.3}
\expandafter\def\csname paperresult@domain.standardized.net.mean\endcsname{21.0}
\expandafter\def\csname paperresult@domain.standardized.net.lo95\endcsname{20.0}
\expandafter\def\csname paperresult@domain.standardized.net.hi95\endcsname{21.9}
\expandafter\def\csname paperresult@domain.standardized.other.mean\endcsname{22.6}
\expandafter\def\csname paperresult@domain.standardized.other.lo95\endcsname{21.9}
\expandafter\def\csname paperresult@domain.standardized.other.hi95\endcsname{23.4}
\expandafter\def\csname paperresult@domain.standardized.org.mean\endcsname{13.4}
\expandafter\def\csname paperresult@domain.standardized.org.lo95\endcsname{12.9}
\expandafter\def\csname paperresult@domain.standardized.org.hi95\endcsname{13.9}
\expandafter\def\csname paperresult@domain.standardized.gov.mean\endcsname{7.5}
\expandafter\def\csname paperresult@domain.standardized.gov.lo95\endcsname{6.6}
\expandafter\def\csname paperresult@domain.standardized.gov.hi95\endcsname{8.5}
\expandafter\def\csname paperresult@domain.standardized.edu.mean\endcsname{7.0}
\expandafter\def\csname paperresult@domain.standardized.edu.lo95\endcsname{6.1}
\expandafter\def\csname paperresult@domain.standardized.edu.hi95\endcsname{7.9}
\expandafter\def\csname paperresult@domain.shared.com.mean\endcsname{27.5}
\expandafter\def\csname paperresult@domain.shared.com.lo95\endcsname{27.0}
\expandafter\def\csname paperresult@domain.shared.com.hi95\endcsname{28.0}
\expandafter\def\csname paperresult@domain.shared.net.mean\endcsname{21.8}
\expandafter\def\csname paperresult@domain.shared.net.lo95\endcsname{21.0}
\expandafter\def\csname paperresult@domain.shared.net.hi95\endcsname{22.7}
\expandafter\def\csname paperresult@domain.shared.other.mean\endcsname{23.5}
\expandafter\def\csname paperresult@domain.shared.other.lo95\endcsname{23.0}
\expandafter\def\csname paperresult@domain.shared.other.hi95\endcsname{24.1}
\expandafter\def\csname paperresult@domain.shared.org.mean\endcsname{13.9}
\expandafter\def\csname paperresult@domain.shared.org.lo95\endcsname{13.5}
\expandafter\def\csname paperresult@domain.shared.org.hi95\endcsname{14.3}
\expandafter\def\csname paperresult@domain.shared.gov.mean\endcsname{7.8}
\expandafter\def\csname paperresult@domain.shared.gov.lo95\endcsname{6.9}
\expandafter\def\csname paperresult@domain.shared.gov.hi95\endcsname{8.7}
\expandafter\def\csname paperresult@domain.shared.edu.mean\endcsname{7.2}
\expandafter\def\csname paperresult@domain.shared.edu.lo95\endcsname{6.4}
\expandafter\def\csname paperresult@domain.shared.edu.hi95\endcsname{8.1}
\expandafter\def\csname paperresult@domain.gap.gov.actual.mean\endcsname{23.6}
\expandafter\def\csname paperresult@domain.gap.gov.actual.lo95\endcsname{22.7}
\expandafter\def\csname paperresult@domain.gap.gov.actual.hi95\endcsname{24.6}
\expandafter\def\csname paperresult@domain.gap.gov.standardized.mean\endcsname{18.9}
\expandafter\def\csname paperresult@domain.gap.gov.standardized.lo95\endcsname{17.9}
\expandafter\def\csname paperresult@domain.gap.gov.standardized.hi95\endcsname{20.0}
\expandafter\def\csname paperresult@domain.gap.gov.shared.mean\endcsname{19.7}
\expandafter\def\csname paperresult@domain.gap.gov.shared.lo95\endcsname{18.8}
\expandafter\def\csname paperresult@domain.gap.gov.shared.hi95\endcsname{20.7}
\expandafter\def\csname paperresult@domain.gap.gov.reduction.mean\endcsname{4.7}
\expandafter\def\csname paperresult@domain.gap.gov.reduction.lo95\endcsname{4.5}
\expandafter\def\csname paperresult@domain.gap.gov.reduction.hi95\endcsname{4.9}
\expandafter\def\csname paperresult@domain.gap.gov.fraction_remaining.mean\endcsname{80.1}
\expandafter\def\csname paperresult@domain.gap.gov.fraction_remaining.lo95\endcsname{78.6}
\expandafter\def\csname paperresult@domain.gap.gov.fraction_remaining.hi95\endcsname{81.5}
\expandafter\def\csname paperresult@domain.gap.edu.actual.mean\endcsname{23.3}
\expandafter\def\csname paperresult@domain.gap.edu.actual.lo95\endcsname{22.3}
\expandafter\def\csname paperresult@domain.gap.edu.actual.hi95\endcsname{24.3}
\expandafter\def\csname paperresult@domain.gap.edu.standardized.mean\endcsname{19.5}
\expandafter\def\csname paperresult@domain.gap.edu.standardized.lo95\endcsname{18.4}
\expandafter\def\csname paperresult@domain.gap.edu.standardized.hi95\endcsname{20.5}
\expandafter\def\csname paperresult@domain.gap.edu.shared.mean\endcsname{20.3}
\expandafter\def\csname paperresult@domain.gap.edu.shared.lo95\endcsname{19.3}
\expandafter\def\csname paperresult@domain.gap.edu.shared.hi95\endcsname{21.2}
\expandafter\def\csname paperresult@domain.gap.edu.reduction.mean\endcsname{3.8}
\expandafter\def\csname paperresult@domain.gap.edu.reduction.lo95\endcsname{3.7}
\expandafter\def\csname paperresult@domain.gap.edu.reduction.hi95\endcsname{4.0}
\expandafter\def\csname paperresult@domain.gap.edu.fraction_remaining.mean\endcsname{83.6}
\expandafter\def\csname paperresult@domain.gap.edu.fraction_remaining.lo95\endcsname{82.5}
\expandafter\def\csname paperresult@domain.gap.edu.fraction_remaining.hi95\endcsname{84.6}
\expandafter\def\csname paperresult@domain.gap.org.actual.mean\endcsname{14.9}
\expandafter\def\csname paperresult@domain.gap.org.actual.lo95\endcsname{14.4}
\expandafter\def\csname paperresult@domain.gap.org.actual.hi95\endcsname{15.4}
\expandafter\def\csname paperresult@domain.gap.org.standardized.mean\endcsname{13.0}
\expandafter\def\csname paperresult@domain.gap.org.standardized.lo95\endcsname{12.5}
\expandafter\def\csname paperresult@domain.gap.org.standardized.hi95\endcsname{13.5}
\expandafter\def\csname paperresult@domain.gap.org.shared.mean\endcsname{13.6}
\expandafter\def\csname paperresult@domain.gap.org.shared.lo95\endcsname{13.2}
\expandafter\def\csname paperresult@domain.gap.org.shared.hi95\endcsname{14.0}
\expandafter\def\csname paperresult@domain.gap.org.reduction.mean\endcsname{1.9}
\expandafter\def\csname paperresult@domain.gap.org.reduction.lo95\endcsname{1.8}
\expandafter\def\csname paperresult@domain.gap.org.reduction.hi95\endcsname{1.9}
\expandafter\def\csname paperresult@domain.gap.org.fraction_remaining.mean\endcsname{87.6}
\expandafter\def\csname paperresult@domain.gap.org.fraction_remaining.lo95\endcsname{87.0}
\expandafter\def\csname paperresult@domain.gap.org.fraction_remaining.hi95\endcsname{88.1}
\expandafter\def\csname paperresult@domain.coverage.com\endcsname{100.0}
\expandafter\def\csname paperresult@domain.coverage.net\endcsname{99.1}
\expandafter\def\csname paperresult@domain.coverage.other\endcsname{99.8}
\expandafter\def\csname paperresult@domain.coverage.org\endcsname{99.3}
\expandafter\def\csname paperresult@domain.coverage.gov\endcsname{22.5}
\expandafter\def\csname paperresult@domain.coverage.edu\endcsname{16.6}
\expandafter\def\csname paperresult@domain.shared_reference_share\endcsname{14.1}
\expandafter\def\csname paperresult@domain.date.baseline\endcsname{December 2021}
\expandafter\def\csname paperresult@domain.date.latest\endcsname{August 2026}
\expandafter\def\csname paperresult@domain.date.earlier\endcsname{February--March 2024}
\expandafter\def\csname paperresult@domain.earlier.com.mean\endcsname{7.0}
\expandafter\def\csname paperresult@domain.earlier.com.lo95\endcsname{6.8}
\expandafter\def\csname paperresult@domain.earlier.com.hi95\endcsname{7.2}
\expandafter\def\csname paperresult@domain.earlier.net.mean\endcsname{5.1}
\expandafter\def\csname paperresult@domain.earlier.net.lo95\endcsname{4.8}
\expandafter\def\csname paperresult@domain.earlier.net.hi95\endcsname{5.3}
\expandafter\def\csname paperresult@domain.earlier.other.mean\endcsname{4.3}
\expandafter\def\csname paperresult@domain.earlier.other.lo95\endcsname{4.1}
\expandafter\def\csname paperresult@domain.earlier.other.hi95\endcsname{4.5}
\expandafter\def\csname paperresult@domain.earlier.org.mean\endcsname{3.9}
\expandafter\def\csname paperresult@domain.earlier.org.lo95\endcsname{3.8}
\expandafter\def\csname paperresult@domain.earlier.org.hi95\endcsname{4.1}
\expandafter\def\csname paperresult@domain.earlier.gov.mean\endcsname{2.5}
\expandafter\def\csname paperresult@domain.earlier.gov.lo95\endcsname{2.3}
\expandafter\def\csname paperresult@domain.earlier.gov.hi95\endcsname{2.8}
\expandafter\def\csname paperresult@domain.earlier.edu.mean\endcsname{2.1}
\expandafter\def\csname paperresult@domain.earlier.edu.lo95\endcsname{1.9}
\expandafter\def\csname paperresult@domain.earlier.edu.hi95\endcsname{2.3}

\providecommand{\Result}[1]{%
  \ifcsname paperresult@#1\endcsname%
    \csname paperresult@#1\endcsname%
  \else%
    \PackageError{paper-results}{Unknown result key #1}{Regenerate results.tex.}%
  \fi}
\expandafter\def\csname paperresult@language.share.english.first\endcsname{39.0}
\expandafter\def\csname paperresult@language.share.nonenglish.first\endcsname{35.8}
\expandafter\def\csname paperresult@language.share.bilingual.first\endcsname{23.9}
\expandafter\def\csname paperresult@language.share.unknown.first\endcsname{1.2}
\expandafter\def\csname paperresult@language.share.english.latest\endcsname{35.9}
\expandafter\def\csname paperresult@language.share.nonenglish.latest\endcsname{37.3}
\expandafter\def\csname paperresult@language.share.bilingual.latest\endcsname{25.0}
\expandafter\def\csname paperresult@language.share.unknown.latest\endcsname{1.7}
\expandafter\def\csname paperresult@language.level.english.start2023.mean\endcsname{2.7}
\expandafter\def\csname paperresult@language.level.english.start2023.lo95\endcsname{2.6}
\expandafter\def\csname paperresult@language.level.english.start2023.hi95\endcsname{2.8}
\expandafter\def\csname paperresult@language.level.nonenglish.start2023.mean\endcsname{2.1}
\expandafter\def\csname paperresult@language.level.nonenglish.start2023.lo95\endcsname{1.9}
\expandafter\def\csname paperresult@language.level.nonenglish.start2023.hi95\endcsname{2.3}
\expandafter\def\csname paperresult@language.level.bilingual.start2023.mean\endcsname{0.5}
\expandafter\def\csname paperresult@language.level.bilingual.start2023.lo95\endcsname{0.4}
\expandafter\def\csname paperresult@language.level.bilingual.start2023.hi95\endcsname{0.5}
\expandafter\def\csname paperresult@language.level.english.start2025.mean\endcsname{15.9}
\expandafter\def\csname paperresult@language.level.english.start2025.lo95\endcsname{15.6}
\expandafter\def\csname paperresult@language.level.english.start2025.hi95\endcsname{16.2}
\expandafter\def\csname paperresult@language.level.nonenglish.start2025.mean\endcsname{9.4}
\expandafter\def\csname paperresult@language.level.nonenglish.start2025.lo95\endcsname{8.5}
\expandafter\def\csname paperresult@language.level.nonenglish.start2025.hi95\endcsname{10.2}
\expandafter\def\csname paperresult@language.level.bilingual.start2025.mean\endcsname{5.8}
\expandafter\def\csname paperresult@language.level.bilingual.start2025.lo95\endcsname{5.5}
\expandafter\def\csname paperresult@language.level.bilingual.start2025.hi95\endcsname{6.1}
\expandafter\def\csname paperresult@language.level.english.latest.mean\endcsname{32.8}
\expandafter\def\csname paperresult@language.level.english.latest.lo95\endcsname{32.2}
\expandafter\def\csname paperresult@language.level.english.latest.hi95\endcsname{33.4}
\expandafter\def\csname paperresult@language.level.nonenglish.latest.mean\endcsname{17.0}
\expandafter\def\csname paperresult@language.level.nonenglish.latest.lo95\endcsname{15.4}
\expandafter\def\csname paperresult@language.level.nonenglish.latest.hi95\endcsname{18.4}
\expandafter\def\csname paperresult@language.level.bilingual.latest.mean\endcsname{20.2}
\expandafter\def\csname paperresult@language.level.bilingual.latest.lo95\endcsname{19.3}
\expandafter\def\csname paperresult@language.level.bilingual.latest.hi95\endcsname{21.0}
\expandafter\def\csname paperresult@language.contrast.english_vs_nonenglish.start2023.pp.mean\endcsname{0.6}
\expandafter\def\csname paperresult@language.contrast.english_vs_nonenglish.start2023.pp.lo95\endcsname{0.4}
\expandafter\def\csname paperresult@language.contrast.english_vs_nonenglish.start2023.pp.hi95\endcsname{0.8}
\expandafter\def\csname paperresult@language.contrast.english_vs_nonenglish.start2023.ratio.mean\endcsname{1.26}
\expandafter\def\csname paperresult@language.contrast.english_vs_nonenglish.start2023.ratio.lo95\endcsname{1.17}
\expandafter\def\csname paperresult@language.contrast.english_vs_nonenglish.start2023.ratio.hi95\endcsname{1.41}
\expandafter\def\csname paperresult@language.contrast.english_vs_nonenglish.start2023.abs\endcsname{0.6}
\expandafter\def\csname paperresult@language.contrast.bilingual_vs_english.start2023.pp.mean\endcsname{-2.2}
\expandafter\def\csname paperresult@language.contrast.bilingual_vs_english.start2023.pp.lo95\endcsname{-2.3}
\expandafter\def\csname paperresult@language.contrast.bilingual_vs_english.start2023.pp.hi95\endcsname{-2.1}
\expandafter\def\csname paperresult@language.contrast.bilingual_vs_english.start2023.ratio.mean\endcsname{0.18}
\expandafter\def\csname paperresult@language.contrast.bilingual_vs_english.start2023.ratio.lo95\endcsname{0.16}
\expandafter\def\csname paperresult@language.contrast.bilingual_vs_english.start2023.ratio.hi95\endcsname{0.20}
\expandafter\def\csname paperresult@language.contrast.bilingual_vs_english.start2023.abs\endcsname{2.2}
\expandafter\def\csname paperresult@language.contrast.bilingual_vs_nonenglish.start2023.pp.mean\endcsname{-1.7}
\expandafter\def\csname paperresult@language.contrast.bilingual_vs_nonenglish.start2023.pp.lo95\endcsname{-1.8}
\expandafter\def\csname paperresult@language.contrast.bilingual_vs_nonenglish.start2023.pp.hi95\endcsname{-1.4}
\expandafter\def\csname paperresult@language.contrast.bilingual_vs_nonenglish.start2023.ratio.mean\endcsname{0.23}
\expandafter\def\csname paperresult@language.contrast.bilingual_vs_nonenglish.start2023.ratio.lo95\endcsname{0.20}
\expandafter\def\csname paperresult@language.contrast.bilingual_vs_nonenglish.start2023.ratio.hi95\endcsname{0.26}
\expandafter\def\csname paperresult@language.contrast.bilingual_vs_nonenglish.start2023.abs\endcsname{1.7}
\expandafter\def\csname paperresult@language.contrast.english_vs_nonenglish.start2025.pp.mean\endcsname{6.5}
\expandafter\def\csname paperresult@language.contrast.english_vs_nonenglish.start2025.pp.lo95\endcsname{5.7}
\expandafter\def\csname paperresult@language.contrast.english_vs_nonenglish.start2025.pp.hi95\endcsname{7.4}
\expandafter\def\csname paperresult@language.contrast.english_vs_nonenglish.start2025.ratio.mean\endcsname{1.68}
\expandafter\def\csname paperresult@language.contrast.english_vs_nonenglish.start2025.ratio.lo95\endcsname{1.56}
\expandafter\def\csname paperresult@language.contrast.english_vs_nonenglish.start2025.ratio.hi95\endcsname{1.86}
\expandafter\def\csname paperresult@language.contrast.english_vs_nonenglish.start2025.abs\endcsname{6.5}
\expandafter\def\csname paperresult@language.contrast.bilingual_vs_english.start2025.pp.mean\endcsname{-10.1}
\expandafter\def\csname paperresult@language.contrast.bilingual_vs_english.start2025.pp.lo95\endcsname{-10.4}
\expandafter\def\csname paperresult@language.contrast.bilingual_vs_english.start2025.pp.hi95\endcsname{-9.8}
\expandafter\def\csname paperresult@language.contrast.bilingual_vs_english.start2025.ratio.mean\endcsname{0.36}
\expandafter\def\csname paperresult@language.contrast.bilingual_vs_english.start2025.ratio.lo95\endcsname{0.35}
\expandafter\def\csname paperresult@language.contrast.bilingual_vs_english.start2025.ratio.hi95\endcsname{0.38}
\expandafter\def\csname paperresult@language.contrast.bilingual_vs_english.start2025.abs\endcsname{10.1}
\expandafter\def\csname paperresult@language.contrast.bilingual_vs_nonenglish.start2025.pp.mean\endcsname{-3.7}
\expandafter\def\csname paperresult@language.contrast.bilingual_vs_nonenglish.start2025.pp.lo95\endcsname{-4.5}
\expandafter\def\csname paperresult@language.contrast.bilingual_vs_nonenglish.start2025.pp.hi95\endcsname{-2.8}
\expandafter\def\csname paperresult@language.contrast.bilingual_vs_nonenglish.start2025.ratio.mean\endcsname{0.61}
\expandafter\def\csname paperresult@language.contrast.bilingual_vs_nonenglish.start2025.ratio.lo95\endcsname{0.56}
\expandafter\def\csname paperresult@language.contrast.bilingual_vs_nonenglish.start2025.ratio.hi95\endcsname{0.68}
\expandafter\def\csname paperresult@language.contrast.bilingual_vs_nonenglish.start2025.abs\endcsname{3.7}
\expandafter\def\csname paperresult@language.contrast.english_vs_nonenglish.latest.pp.mean\endcsname{15.8}
\expandafter\def\csname paperresult@language.contrast.english_vs_nonenglish.latest.pp.lo95\endcsname{14.4}
\expandafter\def\csname paperresult@language.contrast.english_vs_nonenglish.latest.pp.hi95\endcsname{17.4}
\expandafter\def\csname paperresult@language.contrast.english_vs_nonenglish.latest.ratio.mean\endcsname{1.93}
\expandafter\def\csname paperresult@language.contrast.english_vs_nonenglish.latest.ratio.lo95\endcsname{1.78}
\expandafter\def\csname paperresult@language.contrast.english_vs_nonenglish.latest.ratio.hi95\endcsname{2.13}
\expandafter\def\csname paperresult@language.contrast.english_vs_nonenglish.latest.abs\endcsname{15.8}
\expandafter\def\csname paperresult@language.contrast.bilingual_vs_english.latest.pp.mean\endcsname{-12.7}
\expandafter\def\csname paperresult@language.contrast.bilingual_vs_english.latest.pp.lo95\endcsname{-13.4}
\expandafter\def\csname paperresult@language.contrast.bilingual_vs_english.latest.pp.hi95\endcsname{-11.9}
\expandafter\def\csname paperresult@language.contrast.bilingual_vs_english.latest.ratio.mean\endcsname{0.61}
\expandafter\def\csname paperresult@language.contrast.bilingual_vs_english.latest.ratio.lo95\endcsname{0.59}
\expandafter\def\csname paperresult@language.contrast.bilingual_vs_english.latest.ratio.hi95\endcsname{0.64}
\expandafter\def\csname paperresult@language.contrast.bilingual_vs_english.latest.abs\endcsname{12.7}
\expandafter\def\csname paperresult@language.contrast.bilingual_vs_nonenglish.latest.pp.mean\endcsname{3.2}
\expandafter\def\csname paperresult@language.contrast.bilingual_vs_nonenglish.latest.pp.lo95\endcsname{1.6}
\expandafter\def\csname paperresult@language.contrast.bilingual_vs_nonenglish.latest.pp.hi95\endcsname{4.9}
\expandafter\def\csname paperresult@language.contrast.bilingual_vs_nonenglish.latest.ratio.mean\endcsname{1.19}
\expandafter\def\csname paperresult@language.contrast.bilingual_vs_nonenglish.latest.ratio.lo95\endcsname{1.09}
\expandafter\def\csname paperresult@language.contrast.bilingual_vs_nonenglish.latest.ratio.hi95\endcsname{1.31}
\expandafter\def\csname paperresult@language.contrast.bilingual_vs_nonenglish.latest.abs\endcsname{3.2}
\expandafter\def\csname paperresult@language.rate.english.early.mean\endcsname{6.7}
\expandafter\def\csname paperresult@language.rate.english.early.lo95\endcsname{6.6}
\expandafter\def\csname paperresult@language.rate.english.early.hi95\endcsname{6.8}
\expandafter\def\csname paperresult@language.fold.english.early.mean\endcsname{5.89}
\expandafter\def\csname paperresult@language.fold.english.early.lo95\endcsname{5.65}
\expandafter\def\csname paperresult@language.fold.english.early.hi95\endcsname{6.15}
\expandafter\def\csname paperresult@language.rate.nonenglish.early.mean\endcsname{3.7}
\expandafter\def\csname paperresult@language.rate.nonenglish.early.lo95\endcsname{3.4}
\expandafter\def\csname paperresult@language.rate.nonenglish.early.hi95\endcsname{4.0}
\expandafter\def\csname paperresult@language.fold.nonenglish.early.mean\endcsname{4.41}
\expandafter\def\csname paperresult@language.fold.nonenglish.early.lo95\endcsname{4.24}
\expandafter\def\csname paperresult@language.fold.nonenglish.early.hi95\endcsname{4.60}
\expandafter\def\csname paperresult@language.rate.bilingual.early.mean\endcsname{2.7}
\expandafter\def\csname paperresult@language.rate.bilingual.early.lo95\endcsname{2.6}
\expandafter\def\csname paperresult@language.rate.bilingual.early.hi95\endcsname{2.8}
\expandafter\def\csname paperresult@language.fold.bilingual.early.mean\endcsname{11.87}
\expandafter\def\csname paperresult@language.fold.bilingual.early.lo95\endcsname{10.97}
\expandafter\def\csname paperresult@language.fold.bilingual.early.hi95\endcsname{12.90}
\expandafter\def\csname paperresult@language.rate.english.late.mean\endcsname{10.8}
\expandafter\def\csname paperresult@language.rate.english.late.lo95\endcsname{10.5}
\expandafter\def\csname paperresult@language.rate.english.late.hi95\endcsname{11.2}
\expandafter\def\csname paperresult@language.fold.english.late.mean\endcsname{2.06}
\expandafter\def\csname paperresult@language.fold.english.late.lo95\endcsname{2.03}
\expandafter\def\csname paperresult@language.fold.english.late.hi95\endcsname{2.10}
\expandafter\def\csname paperresult@language.rate.nonenglish.late.mean\endcsname{4.8}
\expandafter\def\csname paperresult@language.rate.nonenglish.late.lo95\endcsname{4.3}
\expandafter\def\csname paperresult@language.rate.nonenglish.late.hi95\endcsname{5.3}
\expandafter\def\csname paperresult@language.fold.nonenglish.late.mean\endcsname{1.80}
\expandafter\def\csname paperresult@language.fold.nonenglish.late.lo95\endcsname{1.75}
\expandafter\def\csname paperresult@language.fold.nonenglish.late.hi95\endcsname{1.85}
\expandafter\def\csname paperresult@language.rate.bilingual.late.mean\endcsname{9.2}
\expandafter\def\csname paperresult@language.rate.bilingual.late.lo95\endcsname{8.7}
\expandafter\def\csname paperresult@language.rate.bilingual.late.hi95\endcsname{9.7}
\expandafter\def\csname paperresult@language.fold.bilingual.late.mean\endcsname{3.50}
\expandafter\def\csname paperresult@language.fold.bilingual.late.lo95\endcsname{3.33}
\expandafter\def\csname paperresult@language.fold.bilingual.late.hi95\endcsname{3.68}
\expandafter\def\csname paperresult@language.rate.english.change.mean\endcsname{4.1}
\expandafter\def\csname paperresult@language.rate.english.change.lo95\endcsname{3.7}
\expandafter\def\csname paperresult@language.rate.english.change.hi95\endcsname{4.5}
\expandafter\def\csname paperresult@language.rate.nonenglish.change.mean\endcsname{1.1}
\expandafter\def\csname paperresult@language.rate.nonenglish.change.lo95\endcsname{0.8}
\expandafter\def\csname paperresult@language.rate.nonenglish.change.hi95\endcsname{1.4}
\expandafter\def\csname paperresult@language.rate.bilingual.change.mean\endcsname{6.5}
\expandafter\def\csname paperresult@language.rate.bilingual.change.lo95\endcsname{6.1}
\expandafter\def\csname paperresult@language.rate.bilingual.change.hi95\endcsname{7.0}
\expandafter\def\csname paperresult@language.rate.gap.early.mean\endcsname{3.0}
\expandafter\def\csname paperresult@language.rate.gap.early.lo95\endcsname{2.7}
\expandafter\def\csname paperresult@language.rate.gap.early.hi95\endcsname{3.4}
\expandafter\def\csname paperresult@language.rate.gap.late.mean\endcsname{6.0}
\expandafter\def\csname paperresult@language.rate.gap.late.lo95\endcsname{5.5}
\expandafter\def\csname paperresult@language.rate.gap.late.hi95\endcsname{6.5}
\expandafter\def\csname paperresult@language.rate.gapchange.all.mean\endcsname{3.0}
\expandafter\def\csname paperresult@language.rate.gapchange.all.lo95\endcsname{2.8}
\expandafter\def\csname paperresult@language.rate.gapchange.all.hi95\endcsname{3.2}
\expandafter\def\csname paperresult@language.rate.gapchange.all.prob\endcsname{above 0.999}
\expandafter\def\csname paperresult@language.rate.gapchange.shared_own.mean\endcsname{3.1}
\expandafter\def\csname paperresult@language.rate.gapchange.shared_own.lo95\endcsname{2.8}
\expandafter\def\csname paperresult@language.rate.gapchange.shared_own.hi95\endcsname{3.3}
\expandafter\def\csname paperresult@language.rate.gapchange.shared_own.prob\endcsname{above 0.999}
\expandafter\def\csname paperresult@language.rate.gapchange.shared_common.mean\endcsname{3.4}
\expandafter\def\csname paperresult@language.rate.gapchange.shared_common.lo95\endcsname{3.2}
\expandafter\def\csname paperresult@language.rate.gapchange.shared_common.hi95\endcsname{3.6}
\expandafter\def\csname paperresult@language.rate.gapchange.shared_common.prob\endcsname{above 0.999}
\expandafter\def\csname paperresult@language.cover.english.start2023\endcsname{96.3}
\expandafter\def\csname paperresult@language.cover.nonenglish.start2023\endcsname{100.0}
\expandafter\def\csname paperresult@language.cover.bilingual.start2023\endcsname{99.8}
\expandafter\def\csname paperresult@language.cover.english.start2025\endcsname{97.4}
\expandafter\def\csname paperresult@language.cover.nonenglish.start2025\endcsname{100.0}
\expandafter\def\csname paperresult@language.cover.bilingual.start2025\endcsname{99.9}
\expandafter\def\csname paperresult@language.cover.english.latest\endcsname{97.5}
\expandafter\def\csname paperresult@language.cover.nonenglish.latest\endcsname{100.0}
\expandafter\def\csname paperresult@language.cover.bilingual.latest\endcsname{99.9}
\expandafter\def\csname paperresult@language.cover.others.min\endcsname{99.7}
\expandafter\def\csname paperresult@language.shared.mass\endcsname{99.0}
\expandafter\def\csname paperresult@language.shared.n_profiles\endcsname{74}
\expandafter\def\csname paperresult@language.std.english.latest\endcsname{32.5}
\expandafter\def\csname paperresult@language.std.nonenglish.latest\endcsname{16.8}
\expandafter\def\csname paperresult@language.std.bilingual.latest\endcsname{21.4}
\expandafter\def\csname paperresult@composition.tld.com.adj_tier.shift\endcsname{-0.1}
\expandafter\def\csname paperresult@composition.tld.com.adj_tier.abs\endcsname{0.1}
\expandafter\def\csname paperresult@composition.tld.edu.adj_tier.shift\endcsname{1.5}
\expandafter\def\csname paperresult@composition.tld.edu.adj_tier.abs\endcsname{1.5}
\expandafter\def\csname paperresult@composition.tld.gov.adj_tier.shift\endcsname{4.3}
\expandafter\def\csname paperresult@composition.tld.gov.adj_tier.abs\endcsname{4.3}
\expandafter\def\csname paperresult@composition.tld.net.adj_tier.shift\endcsname{-0.3}
\expandafter\def\csname paperresult@composition.tld.net.adj_tier.abs\endcsname{0.3}
\expandafter\def\csname paperresult@composition.tld.org.adj_tier.shift\endcsname{1.5}
\expandafter\def\csname paperresult@composition.tld.org.adj_tier.abs\endcsname{1.5}
\expandafter\def\csname paperresult@composition.tld.other.adj_tier.shift\endcsname{-0.2}
\expandafter\def\csname paperresult@composition.tld.other.adj_tier.abs\endcsname{0.2}
\expandafter\def\csname paperresult@composition.tld.com.adj_lang.shift\endcsname{-2.6}
\expandafter\def\csname paperresult@composition.tld.com.adj_lang.abs\endcsname{2.6}
\expandafter\def\csname paperresult@composition.tld.edu.adj_lang.shift\endcsname{-1.1}
\expandafter\def\csname paperresult@composition.tld.edu.adj_lang.abs\endcsname{1.1}
\expandafter\def\csname paperresult@composition.tld.gov.adj_lang.shift\endcsname{-1.9}
\expandafter\def\csname paperresult@composition.tld.gov.adj_lang.abs\endcsname{1.9}
\expandafter\def\csname paperresult@composition.tld.net.adj_lang.shift\endcsname{0.8}
\expandafter\def\csname paperresult@composition.tld.net.adj_lang.abs\endcsname{0.8}
\expandafter\def\csname paperresult@composition.tld.org.adj_lang.shift\endcsname{-2.1}
\expandafter\def\csname paperresult@composition.tld.org.adj_lang.abs\endcsname{2.1}
\expandafter\def\csname paperresult@composition.tld.other.adj_lang.shift\endcsname{2.7}
\expandafter\def\csname paperresult@composition.tld.other.adj_lang.abs\endcsname{2.7}
\expandafter\def\csname paperresult@composition.tld.com.adj_len.shift\endcsname{-0.3}
\expandafter\def\csname paperresult@composition.tld.com.adj_len.abs\endcsname{0.3}
\expandafter\def\csname paperresult@composition.tld.edu.adj_len.shift\endcsname{-0.0}
\expandafter\def\csname paperresult@composition.tld.edu.adj_len.abs\endcsname{0.0}
\expandafter\def\csname paperresult@composition.tld.gov.adj_len.shift\endcsname{0.3}
\expandafter\def\csname paperresult@composition.tld.gov.adj_len.abs\endcsname{0.3}
\expandafter\def\csname paperresult@composition.tld.net.adj_len.shift\endcsname{0.0}
\expandafter\def\csname paperresult@composition.tld.net.adj_len.abs\endcsname{0.0}
\expandafter\def\csname paperresult@composition.tld.org.adj_len.shift\endcsname{-0.1}
\expandafter\def\csname paperresult@composition.tld.org.adj_len.abs\endcsname{0.1}
\expandafter\def\csname paperresult@composition.tld.other.adj_len.shift\endcsname{0.2}
\expandafter\def\csname paperresult@composition.tld.other.adj_len.abs\endcsname{0.2}
\expandafter\def\csname paperresult@language.std.maxshift\endcsname{1.0}
\expandafter\def\csname paperresult@language.fold.english.full.mean\endcsname{12.17}
\expandafter\def\csname paperresult@language.fold.english.full.lo95\endcsname{11.60}
\expandafter\def\csname paperresult@language.fold.english.full.hi95\endcsname{12.77}
\expandafter\def\csname paperresult@language.fold.nonenglish.full.mean\endcsname{7.94}
\expandafter\def\csname paperresult@language.fold.nonenglish.full.lo95\endcsname{7.55}
\expandafter\def\csname paperresult@language.fold.nonenglish.full.hi95\endcsname{8.38}
\expandafter\def\csname paperresult@language.fold.bilingual.full.mean\endcsname{41.67}
\expandafter\def\csname paperresult@language.fold.bilingual.full.lo95\endcsname{36.84}
\expandafter\def\csname paperresult@language.fold.bilingual.full.hi95\endcsname{47.15}
\expandafter\def\csname paperresult@language.foldratio.english_vs_nonenglish.early.mean\endcsname{1.34}
\expandafter\def\csname paperresult@language.foldratio.english_vs_nonenglish.early.lo95\endcsname{1.30}
\expandafter\def\csname paperresult@language.foldratio.english_vs_nonenglish.early.hi95\endcsname{1.37}
\expandafter\def\csname paperresult@language.foldratio.english_vs_nonenglish.late.mean\endcsname{1.15}
\expandafter\def\csname paperresult@language.foldratio.english_vs_nonenglish.late.lo95\endcsname{1.13}
\expandafter\def\csname paperresult@language.foldratio.english_vs_nonenglish.late.hi95\endcsname{1.17}
\expandafter\def\csname paperresult@language.relative.bilingual_to_english.start2023.mean\endcsname{0.18}
\expandafter\def\csname paperresult@language.relative.bilingual_to_english.start2023.lo95\endcsname{0.16}
\expandafter\def\csname paperresult@language.relative.bilingual_to_english.start2023.hi95\endcsname{0.20}
\expandafter\def\csname paperresult@language.contrast.bilingual_vs_nonenglish.start2023.prob\endcsname{0.000}
\expandafter\def\csname paperresult@language.relative.bilingual_to_english.start2025.mean\endcsname{0.36}
\expandafter\def\csname paperresult@language.relative.bilingual_to_english.start2025.lo95\endcsname{0.35}
\expandafter\def\csname paperresult@language.relative.bilingual_to_english.start2025.hi95\endcsname{0.38}
\expandafter\def\csname paperresult@language.contrast.bilingual_vs_nonenglish.start2025.prob\endcsname{0.000}
\expandafter\def\csname paperresult@language.relative.bilingual_to_english.latest.mean\endcsname{0.61}
\expandafter\def\csname paperresult@language.relative.bilingual_to_english.latest.lo95\endcsname{0.59}
\expandafter\def\csname paperresult@language.relative.bilingual_to_english.latest.hi95\endcsname{0.64}
\expandafter\def\csname paperresult@language.contrast.bilingual_vs_nonenglish.latest.prob\endcsname{above 0.999}
\expandafter\def\csname paperresult@language.cross.bilingual_nonenglish.crawl\endcsname{CC-MAIN-2026-08}
\expandafter\def\csname paperresult@language.cross.bilingual_nonenglish.date\endcsname{February 2026}

\newcommand{\given}{\mid}

\newcommand{\calC}{\mathcal{C}}
\newcommand{\calS}{\mathcal{S}}
\newcommand{\calD}{\mathcal{D}}
\newcommand{\calL}{\mathcal{L}}
\newcommand{\calW}{\mathcal{W}}

\newcommand{\calJ}{\mathcal{J}}
\newcommand{\calB}{\mathcal{B}}
\newcommand{\calK}{\mathcal{K}}
\newcommand{\calY}{\mathcal{Y}}
\newcommand{\calX}{\mathcal{X}}
\newcommand{\Se}{\mathrm{se}}
\newcommand{\Sp}{\mathrm{sp}}
\newcommand{\Ber}{\mathsf{Bernoulli}}
\newcommand{\Normal}{\mathsf{N}}
\newcommand{\Beta}{\mathsf{Beta}}
\newcommand{\Unif}{\mathsf{Uniform}}
\newcommand{\Binom}{\mathsf{Binomial}}
\newcommand{\halfNormal}{\mathsf{N}_{+}}
\newcommand{\logit}{\mathrm{logit}}

\definecolor{modlik}{RGB}{222,235,247}   
\definecolor{prior}{RGB}{226,240,217}   
\newcommand{\module}[3]{\underset{\parbox{2.5cm}{\centering\scriptsize #3}}{\colorbox{#1}{$\displaystyle #2$}}}

\usepackage{soul}

\title{The uneven rise of AI-generated web content}

\author[1]{Adam Visokay\thanks{These authors contributed equally. Please direct correspondence to \href{emailto:avisokay@berkeley.edu}{avisokay@berkeley.edu} and \href{emailto:tylermc@uw.edu}{tylermc@uw.edu}. All authors were either organizers$^{,\dagger,}$ or participants of the 2026 Summer Institute in Computational Social Science at the University of Washington (\href{https://sicss.io/2026/uw/}{SICSS-UW}).}$^\dagger$}
\author[2]{Soumyakanti Pan$^{*,\dagger,}$}
\author[3]{Molly Griston}
\author[4]{Polina Iasakova}
\author[5]{Mark Nepf$^{\dagger,}$}
\author[1]{Denis Peskoff}
\author[6]{Riya Sinha}
\author[7]{Jikai Sun}
\author[2]{Isabelle Thiara$^{\dagger,}$}
\author[8]{Letao Wang}
\author[6]{Yiwei Wu}
\author[9]{Lauren Yan}
\author[2,10]{Tyler H. McCormick$^{\dagger,}$}

\affil[1]{School of Information, University of California, Berkeley}
\affil[2]{Department of Statistics, University of Washington}
\affil[3]{Department of Physics, University of Colorado, Boulder}
\affil[4]{Department of Statistics, Oregon State University}
\affil[5]{Evans School of Public Policy, University of Washington}
\affil[6]{Department of Information, University of Texas at Austin}
\affil[7]{Hubbard School of Journalism and Mass Communication, University of Minnesota}
\affil[8]{Department of Sociology, Washington University in St. Louis}
\affil[9]{Department of Family Medicine \& Community Health, University of Kansas Medical Center}
\affil[10]{Department of Sociology, University of Washington}

\date{}

\begin{document}

\maketitle
\vspace{-2.5em}   


\begin{abstract}
In less than five years, generative artificial intelligence (AI) has dramatically reshaped the public-facing web. Using a sample of more than 5 million crawlable webpages spanning 2019 to 2026, we show that the share containing AI-generated text rose from near zero before 2022 to nearly 25\% by August 2026. This rise has also been uneven. Our analyses reveal substantial differences in AI composition across language and domain categories (e.g., \texttt{.edu}, \texttt{.com}, or \texttt{.gov}). A higher fraction of English-language webpages contain AI-generated text compared to pages in other languages, and prevalence is substantially lower on \texttt{.gov} and \texttt{.edu} domains compared to \texttt{.com}.  These distinctions persist even after accounting for compositional effects.  We also find heterogeneity based on a page's position in the connectivity structure of the web.  AI-generated text is most common and has increased most rapidly in pages that are the least central in the graph, which drives much of the overall growth in AI-generated text. A slower increase has taken place for more central pages, with the most central tier emerging as the least likely to contain AI-generated text. Methodologically, we use multiple AI detection tools, and generate estimates using a Bayesian hierarchical model that incorporates corrections for the detectors' statistical properties (both false positives and false negatives) and allows for different detector performance across groups, while propagating uncertainty in detector error rates into prevalence estimates. Our work suggests that AI is a prominent, but varied, contributor to public information and internet-based communication.
\end{abstract}

\newpage
\section{Introduction}
\label{sec:intro}

The public web is both a main channel through which institutions, businesses, and individuals communicate and the largest source of text from which language models are trained \citep{brown2020gpt3, touvron2023llama, penedo2024fineweb, baack2024critical}. Changes in how text on the web is produced propagate to what people read and to how the next generation of language models learn \citep{shumailov2024collapse}. Generative artificial intelligence (AI) is now reshaping the text on webpages at a rate that is accelerating and in a manner that is highly heterogeneous.
Using a probability sample of 5.1 million pages from the crawlable web from 2019-2026, spanning every language and the full crawlable link graph, together with a Bayesian model that corrects for the false positives and false negatives of multiple AI detectors, we find that in August 2026 about 24 percent (95\% credible interval 22.9-24.4) of crawlable webpages contained AI-generated or AI-edited text, and the rate of increase has nearly doubled, from 4.6 percentage points per year between early 2023 and early 2025 to 8.1 percentage points per year since. About four-fifths of the increase in the prevalence of pages with \Result{outcome.definition} comes from those belonging to domains on the periphery of the web graph that are not highly connected to other well-connected domains, while pages belonging to the most central domains on the web are least likely to contain \Result{outcome.definition}. Further, generative AI has spread faster among English-language pages and .com domains, though we see rises across languages and domain categories, even after accounting for compositional effects.  These results have implications specific to each role the public web plays. First, the growth in AI-influenced text means that generative AI has an increasing role in setting the conventions for how information is presented to the public, and how companies or governments communicate with their citizens.  Second, because the crawlable web is a main input for training future language models, the rise of AI-generated or edited content means that earlier language models increasingly shape the conventions, style, and norms of newer ones.

For the purpose of this paper, we operationalize ``AI-generated text'' as text generated with or revised by language models.  To identify AI-generated text, we take a holistic approach that views AI detection as a measurement error problem.  We use multiple AI detectors and treat their output as imperfect indicators of text's true origin~\citep{bollen1989structural,jacobs2021measurement}.  Commercial and open-source detectors infer AI involvement by identifying textual patterns correlated with language models' writing styles.  AI signatures vary across contexts and also appear in human writing, so we expect that the degree of measurement error will depend on the circumstances. In our evaluations, for example, the open Pangram model \citep{thai2026editlens} has fewer false positives than Binoculars-Falcon \citep{hans2024spotting} on \texttt{.gov} and \texttt{.edu} domains but more on \texttt{.com} domains (more detail in the Appendix section \ref{appendix:detector_benchmarks}). To address this context-specific measurment error, we build a Bayesian hierarchical model that combines detector outputs with labeled benchmarks and pre-2022 control crawls to estimate prevalence jointly with sensitivity and specificity. The early crawls inform false-positive rates under an assumption of low prevalence, while four benchmark datasets (one with texts from 9 languages) inform sensitivity and its variation across sources. We allow error rates to differ across contexts and to shift after the control period.  The specificity adjustment is specific to each detector, language, centrality tier, top-level domain, and text length.  The sensitivity adjustment varies across detector and language.  Uncertainty about those rates enters the intervals for prevalence and its changes over time, allowing us to assess whether differences between groups and changes in their growth rates persist after accounting for detector error.  Several previous studies also account for these errors, but in more limited contexts than we consider~\citep{brooks2024,latona2024ai,liang2024,russell2025newspapers,elazar2026}. We discuss related work further in Appendix \ref{sec:related}.

Our fundamental goals are to (i) estimate trends over time in the fraction of public-facing webpages using AI-generated text and, (ii) to determine how the prevalence varies across domain centrality, language, and top-level domain. Related to our first goal of understanding trends in prevalence over time, studies of newly archived English-language pages document a substantial increase in detected AI use between 2022 and mid-2025 \citep{dolezal2026}. Graphite likewise finds a sharp initial increase among newly published English-language articles, followed by an approximately stable share over the five quarters from early 2025 through early 2026 \citep{graphite2025, graphite2026}. These studies discuss new content, however, rather than the total prevalence, which is our focus and is critical to understand how AI-generated text impacts users' experiences and the composition of the web overall.  Several other authors look at AI-generated text, but in specific contexts rather than the public facing web. \cite{russell2025newspapers}, for example, audit 186,000 articles from 1,500 American newspapers published in summer 2025 and find that roughly 9\% carry detected AI use, concentrated mostly in small local outlets. The Pew Research Center does address the overall prevalence of AI-generated text, but restricts their analysis to English-language pages and report only the share of pages flagged by a particular AI-detector \citep{pew2026}. As described above, individual AI detectors have context-specific false positive and false negative rates, meaning that a study reporting only the share of pages flagged as AI by a detector will inherit those biases. Pew also draws repeated random samples of English-language pages from Common Crawl \citep{pew2026}, which means that the sampling distribution approximately mirrors that of the population, leaving relatively few examples of webpages in small groups (such as the ones mostly highly connected to other domains), unlike our design which is stratified to ensure we seen enough pages with rare characteristics to make meaningful between-group comparisons. Overall, these studies provide compelling preliminary evidence that AI is reshaping the web, but have yet to answer the question of how prevalence changes over time across English and non-English sites in a way that adjusts for the inevitable bias that comes from bespoke detector performance.

Existing work also largely leaves open our second main question about the distribution of the spread of AI-generated text. The same overall prevalence can have vastly differing implications depending on where increases occur relative to the distribution of pages. Our work explores a domain’s position in the web’s link network, which has not been addressed in previous work. Specifically, we construct six PageRank~\citep{page1999pagerank} tiers computed based on the domain. Higher PageRank means a domain is connected to many other highly connected domains, and is a measure of the centrality of a domain in the network graph of the web. The most central tier of domains contains a few hundred platform, infrastructure and institutional domains, such as WordPress.org, Wikimedia, the Internet Archive and the European Union, while the least central tier includes small businesses, local organizations and personal sites that make up ninety percent of domains and about two-fifths of eligible pages. We ask how prevalence and growth differ between peripheral and highly connected domains and how much each contributes to the overall increase. We also expand the treatment of language presented by previous work, investigating trends in prevalence across in non-English languages.  Our modeling approach facilitates this additional dimension to our analysis by allowing us to incorporate language-specific error rates for multiple AI detectors.  We also explore heterogeneity among top-level domain types (e.g., \texttt{.gov}, \texttt{.com}, \texttt{.edu}). Limited existing evidence comes from two sources. Pew documents differences between top-level domains \citep{pew2026}, although this could again be biased by the performance properties of the detector. DeGenTWeb characterizes websites dominated by generated text, but restricts their analysis to publishing and commercial settings \citep{he2026degentweblookllmdominantwebsites}.  An additional contribution in our analysis is distinguishing differences among comparable pages from differences in the kinds of pages each group contains. Lower prevalence of AI-generated text on \texttt{.gov} domains compared to \texttt{.com}, for example, could partly reflect differences in language, text length, and centrality of those domains. Those characteristics can also affect detector performance, creating a further source of apparent differences. We therefore ask which differences in prevalence and growth persist after accounting for both page composition and detector error.

\section{Results}\label{sec:results}
In this section, we characterize the growth and distribution of AI-generated text across the crawl-observed web. We analyzed \Result{sample.n_pages} eligible pages from \Result{sample.n_domains} domains across \Result{sample.n_crawls} Common Crawl snapshots spanning \Result{sample.date_range}. We draw a two-stage stratified cluster sample from each crawl wherein domains are sampled by six strata defined by PageRank rank in the crawl's web graph, with the most central stratum taken in full, and pages are then randomly sampled within each selected domain (Section~\ref{sec:sampling}). Since a sampling design may not feasibly balance every combination of covariates (cell) in proportion to its share of the web, inference from the sample does not naturally extend to the full population of eligible pages. We therefore use poststratification, which extends inference from the sampled data to the full population \citep{kennedy_2020}. Poststratification computes a weighted average of the posterior distribution of prevalence within each cell, defined by language, text length, PageRank tier and top-level domain, where the weights are constructed from additional data on the number of eligible pages in each cell. The Bayesian hierarchical model corrects each cell's estimate for detector error, and poststratification combines these corrected estimates into the prevalence of any group of cells, such as a tier, a language or a TLD category, each with a credible interval that reflects uncertainty in the detectors' sensitivity and specificity.

We evaluated five detectors on labeled benchmarks and deploy two for the full analysis (Section~\ref{detection}). We scored each page with Open Pangram, a supervised classifier fine-tuned from Llama-3.2-3B. We also scored every English page with Binoculars with the Falcon-7B pair, a zero-shot perplexity-ratio detector (its non-English sensitivity on the multilingual benchmark of $0.151$ did not justify the compute to score non-English pages at scale). The two other detectors we evaluated were uniformly outperformed by the detectors we kept, and the errors of the commercial Pangram~4 are too closely tied to open Pangram's to count as an independent reading (Section~\ref{sec:cond_indep}), so we use it only to check the error rates of the detectors we deploy. Detector accuracy differs by language, so we estimate sensitivity and specificity separately for English and non-English text, and specificity also varies with page characteristics such as domain type and text length (Section~\ref{sec:modeling}). 

Our primary outcome is the prevalence of pages containing \Result{outcome.definition}, expressed as a proportion of eligible pages. Eligible pages are successfully captured pages with at least 500 characters of extracted main text, which excludes navigation, login and other boilerplate pages. The prevalence of a group, whether a PageRank tier, a language or a top-level domain (TLD) category, is the share of the eligible pages in that group, where a page inherits the tier and the TLD category of the domain it belongs to. We report the overall prevalence and by network position (PageRank tier), language, and TLD category. 

We estimate it using a Bayesian hierarchical model that accounts for heterogeneity in detector sensitivity and specificity, languages, centrality tiers, TLD categories and text lengths, and propagates the resulting uncertainty into every reported estimate and contrast. We also poststratify the posterior distribution so that the estimates reflect the population of all eligible pages in the Common Crawl web rather than the composition of only the sampled pages (Section~\ref{sec:modeling}).

\subsection{Overall prevalence and rate of change}

\begin{figure}[!htbp]
  \centering
  \includegraphics[width=0.9\linewidth]{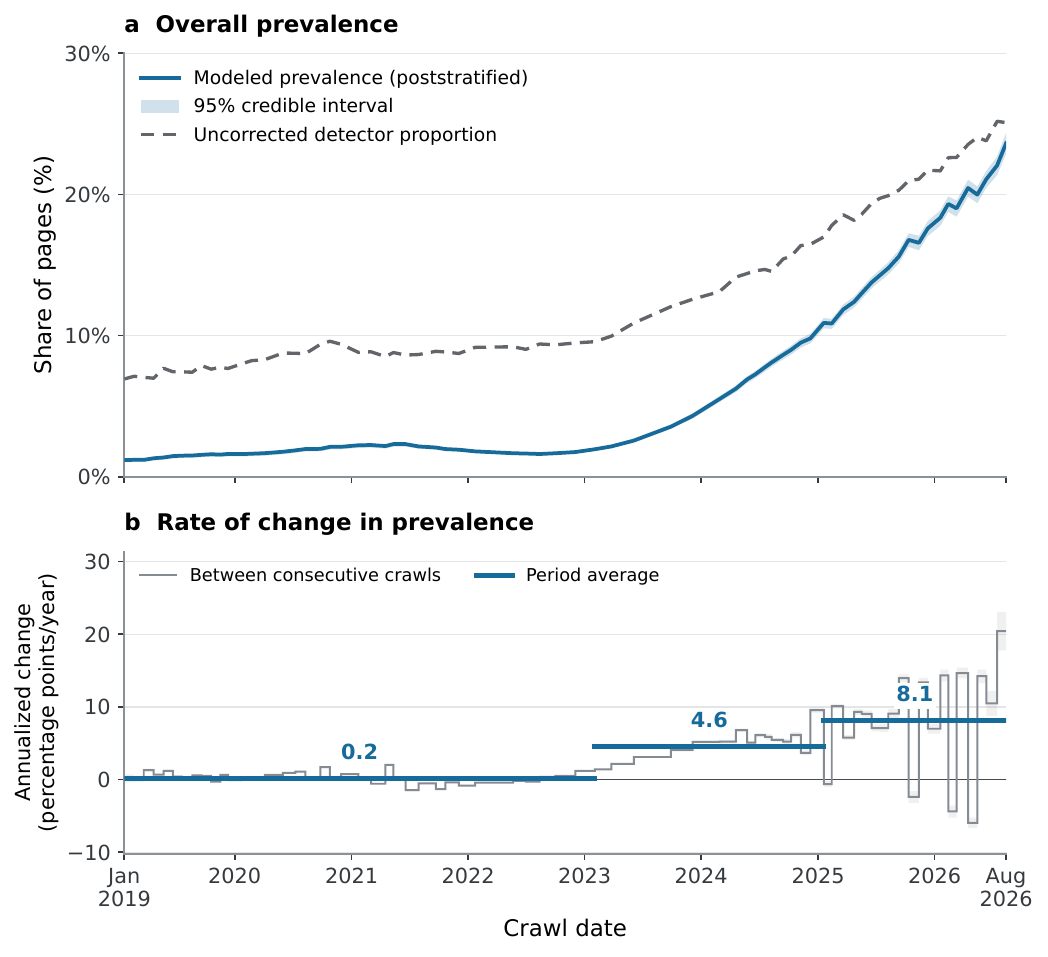}
  \caption{\textbf{Estimated overall prevalence and its rate of change of pages with \Result{outcome.definition} in Common Crawl.} \textbf{(a)} The dashed gray line shows Open Pangram flags divided by analyzed pages in each crawl. The blue curve is the poststratified posterior mean, with cells weighted by their number of eligible pages, and pointwise 95\% credible intervals shaded. The raw proportions are unweighted; their comparison with the poststratified estimate includes both detector correction and population weighting. \textbf{(b)} Gray segments show the change between consecutive crawl estimates divided by elapsed time, in percentage points per year (365.25 days). Blue segments give endpoint-based averages for the three comparison periods described in the text. Shading in panel b gives pointwise 95\% credible intervals calculated from matched joint posterior draws. All \Result{sample.n_crawls} crawl estimates are used. Dates are crawl midpoints; prevalence describes content present in each snapshot.}
  \label{fig:overall-prevalence}
\end{figure}

The fraction of pages with \Result{outcome.definition} has increased substantially since the widespread introduction of language models.  In the most recent crawl, in August 2026, we estimate that about one in four crawlable pages had AI-generated text (95\% credible interval 22.9-24.4).  Figure~\hyperref[fig:overall-prevalence]{\ref*{fig:overall-prevalence}{a}} demonstrates the estimated trend using our Bayesian hierarchical model and, for comparison, using an uncorrected fraction of pages flagged as positive using the open Pangram detector. Unsurprisingly, the estimated fraction based on our model was low before the introduction of commercial language models in late 2022 and early 2023, although the uncorrected detector puts the prevalence substantially higher, since it has not been adjusted for the detectors' sensitivity and specificity, and reflects the composition of the sample rather than the population.

Since the introduction of language models, the rate of change in the prevalence of pages with AI-generated text has also increased (Figure~\hyperref[fig:overall-prevalence]{\ref*{fig:overall-prevalence}{b}}). The average annual change was \Result{overall.rate.baseline.rate_pp_year} percentage points from \Result{overall.rate.baseline.start_date} to \Result{overall.rate.baseline.end_date}, \Result{overall.rate.early.rate_pp_year} percentage points from \Result{overall.rate.early.start_date} to \Result{overall.rate.early.end_date}, and \Result{overall.rate.late.rate_pp_year} percentage points from \Result{overall.rate.late.start_date} to \Result{overall.rate.late.end_date}. The difference between the last two rates was \Result{overall.rate.difference.rate_difference_pp_year} percentage points per year (95\% credible interval: \Result{overall.rate.difference.lo95_pp_year} to \Result{overall.rate.difference.hi95_pp_year}).  These results indicate that, not only are pages with AI-generated text common on the public-facing web, but that the rate at which they are becoming more common is growing substantially year-over-year, suggesting that the role of AI-generated text is growing and accelerating in the content that people see on the public-facing web and in text ingested for training by contemporary and future language models.

\subsection{Increasing prevalence comes mostly from the least connected sites}
We next analyze how AI-generated text varies with the network centrality of a page's domain. Centrality is measured by PageRank~\citep{page1999pagerank} in Common Crawl's hyperlink graph of the web, in which domains are nodes and a link from a page on one domain to a page on another is an edge. Domains are sorted into six PageRank tiers (Table~\ref{tab:strata}); each page takes the tier of its domain, and prevalence in a tier is the share of its eligible pages that contain AI-generated text. PageRank scores a domain by the number and importance of the domains that link to it, so that a link from a widely linked domain counts for more than a link from one that is less frequently linked to, and the score can be read as the share of time a surfer who follows links at random would spend on that domain.

\begin{figure}[!htbp]
  \centering
  \includegraphics[width=0.9\linewidth]{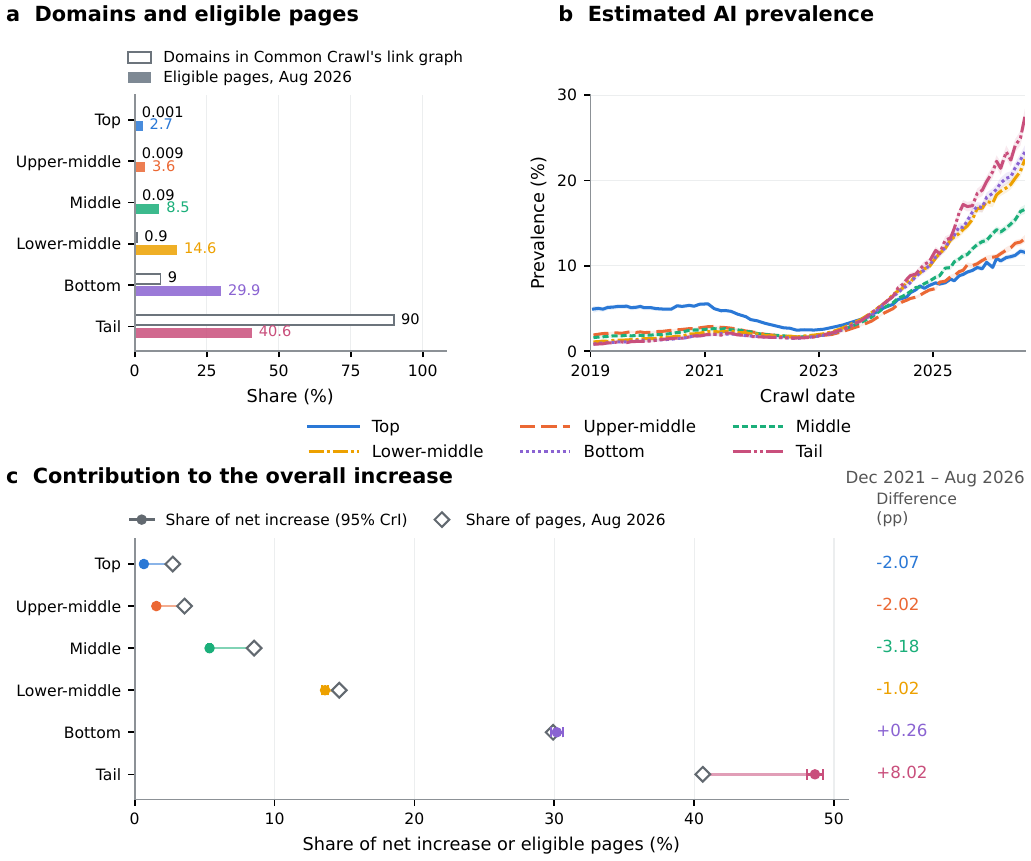}
  \caption{\small \textbf{Estimated prevalence and growth of pages with \Result{outcome.definition} across the Common Crawl's link network.} \textbf{(a)} Shares of domains in Common Crawl's link graph, set by the PageRank tier cutoffs in Table~\ref{tab:strata}, and estimated shares of eligible pages in August 2026 using the model's page-count weights. \textbf{(b)} Prevalence by PageRank tier, the share of the tier's eligible pages with fully AI-generated or AI-edited text. Lines show posterior means and shaded bands show 95\% credible intervals. \textbf{(c)} Each tier's share of the overall prevalence increase from December 2021 to August 2026 (circles; 95\% credible intervals), compared with its share of eligible pages in August 2026 (diamonds). With page share $w_{jk}$ and prevalence $p_{jk}$ of tier $j$ in crawl $k$, and overall prevalence $\bar\theta_k=\sum_j w_{jk}p_{jk}$, the growth share is $(w_{jk_1}p_{jk_1}-w_{jk_0}p_{jk_0})/(\bar\theta_{k_1}-\bar\theta_{k_0})$ for the first and last crawls $k_0$ and $k_1$ of the comparison, calculated in each posterior draw using the composition observed at each date. The difference denotes final page share subtracted from growth share in percentage points. These contributions combine within-tier prevalence change and changes in tier representation; the final page-share comparison does not separate them. Colors identify tiers from Top (most central) to Tail (least central); dates are crawl midpoints.}
  \label{fig:centrality-prevalence}
\end{figure}

The distribution of PageRank scores across domains is extremely skewed (Figure~\hyperref[fig:centrality-prevalence]{\ref*{fig:centrality-prevalence}{a}}). In August 2026, top-tier domains made up 0.001\% of domains and 2.7\% of eligible pages. They are mostly the platforms and infrastructure of the web itself, such as WordPress.org, Wix, Vimeo, Yahoo, Mozilla and the Internet Archive, together with a few global institutions and publishers such as the European Union, the IRS, Harvard and the Financial Times. The three middle tiers, which collectively represent 1\% of domains and 26.7\% of eligible pages, are the recognizable institutional web, including universities, national news outlets, government bodies, and established companies as well as regional organizations such as the Utah Bar Association; Larimer County, Colorado; and Volkswagen Japan. The two lowest tiers hold 99\% of domains and 70.5\% of eligible pages, having a median of 9 to 28 pages per domain across samples, and consist of small businesses, local organizations, and personal sites.  Examples of domains in these tiers include a Swedish rabbit breeder, a Louisville dog-walking service, a German hair salon, a financial planner in Argentina, a French roofing contractor, an Elks Lodge in Florida, a Zurich city council, and a pizza restaurant. 

We estimate that AI-generated text has reached every level of the web's link network, including pages in its most central domains (Figure~\hyperref[fig:centrality-prevalence]{\ref*{fig:centrality-prevalence}{b}}). Each page inherits the PageRank tier of the domain to which it belongs, so a tier consists of all eligible pages whose domain has a PageRank within the range that defines that tier (Table~\ref{tab:strata}). Hence, prevalence in a tier corresponds to the proportion of its eligible pages that contain \Result{outcome.definition}. Estimated prevalence was low across all tiers until 2022 (between 0.8\% and 4.9\% in 2019), and by the August 2026 snapshot, it had risen to between \Result{centrality.latest.top.mean}\% and \Result{centrality.latest.bottom.mean}\% in every tier, including \Result{centrality.latest.top.mean}\% (95\% credible interval \Result{centrality.latest.top.lo95}--\Result{centrality.latest.top.hi95}\%) in the most-central tier (Top). We estimate higher prevalence toward the periphery of the web, reaching \Result{centrality.latest.bottom.mean}\% (\Result{centrality.latest.bottom.lo95}--\Result{centrality.latest.bottom.hi95}\%) in the least-central tier (Tail). 

Moreover, tiers differ in the composition of their pages by TLD, language and length, and these attributes are themselves associated with prevalence. To assess whether the differences between tiers are confounded with these attributes, we compare the estimated prevalence in the tiers after standardizing them with respect to a common composition of TLD, language and length, observed as of August 2026. Upon standardization, estimated prevalence still increases monotonically across tiers, from 11.1\% in the Top to 26.7\% in the Tail, and no tier's standardized estimate differs from its original estimate by more than 1.2 percentage points. The Upper-middle tier, for example, rises from 13.2\% to 14.3\%, because only 34.7\% of its pages are on \texttt{.com}, where prevalence is highest, against 59.6\% in the Top, and 23.9\% are on \texttt{.edu}, \texttt{.gov} or \texttt{.org}, where prevalence is low. The differences in prevalence between tiers are therefore not attributable to differences in TLD, language and length composition.

Figure~\hyperref[fig:centrality-prevalence]{\ref*{fig:centrality-prevalence}{c}} reports each tier's contribution to the increase in overall prevalence between \Result{centrality.decomp.baseline_date} and \Result{centrality.decomp.latest_date}. The two least-central tiers, Bottom and Tail, hold 70.5\% of eligible pages in August 2026 and account for \Result{centrality.decomp.low_share_export_mean}\% of the net increase (conservative one-sided 95\% posterior lower bound: \Result{centrality.decomp.low_share_conservative_lower}\%). The contribution of a tier has two components, the change in prevalence within the tier and the change in the tier's share of eligible pages, as given by Equation~\eqref{eq:centrality-growth-decomposition}. Of the \Result{centrality.decomp.overall_change} percentage-point increase in overall prevalence, \Result{centrality.decomp.low_within} points are attributable to the change in prevalence within Bottom and Tail and \Result{centrality.decomp.low_composition} points to the change in their share of eligible pages (see Figure~\ref{fig:centrality-growth-decomposition} in the Appendix). The concentration of growth in the periphery is therefore driven by increasing prevalence within these tiers rather than by a shift in the distribution of pages across tiers. At a fixed distribution of pages across tiers, equal to the average over the two dates, prevalence increased by \Result{centrality.decomp.low_within_rate} percentage points within Bottom and Tail, against \Result{centrality.decomp.other_within_rate} points in the remaining tiers.

\subsection{AI-generated text is more common on English-language pages}

We next compare pages by language, using the language flag that Common Crawl records for each page (Section~\ref{sec:data_desc}). Pages are categorized into four groups: English-only, non-English-only (one or more languages detected, none of them English), bilingual (English and at least one other language detected), and unknown (no language recorded). In August 2026, an estimated \Result{language.share.english.latest}\% of eligible pages were English-only, \Result{language.share.nonenglish.latest}\% non-English-only, \Result{language.share.bilingual.latest}\% bilingual and \Result{language.share.unknown.latest}\% of unknown language. Bilingual pages are scored as English pages and unknown pages as non-English (Section~\ref{detection}).

\begin{figure}[!htbp]
  \centering
  \includegraphics[width=0.9\linewidth]{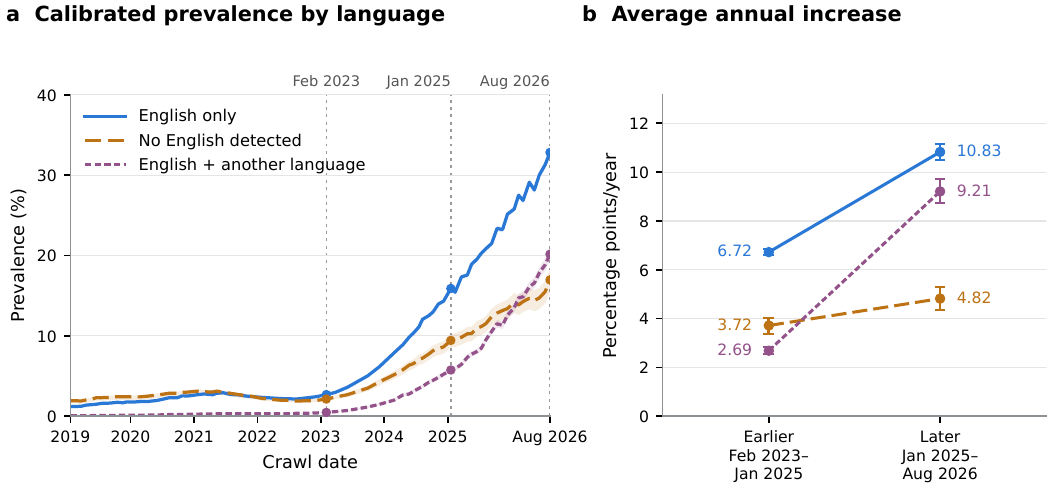}
  \caption{\small\textbf{Language differences in estimated prevalence and growth of pages with \Result{outcome.definition} in Common Crawl.} \textbf{(a)} Page-count-weighted prevalence of fully AI-generated or AI-edited text, with 95\% credible intervals. Groups are English only, no English detected, and English plus another detected language (bilingual); pages with unknown language are excluded from this figure. \textbf{(b)} Average annual increases from February 2, 2023 to January 19, 2025, and from that date to August 13, 2026. Changes use elapsed days (365.25 days per year) and are averages between crawl snapshots, not the date growth changed. Intervals are the 2.5th and 97.5th percentiles of matched posterior-draw contrasts.}
  \label{fig:language-prevalence}
\end{figure}

Prevalence is estimated to be substantially higher among English-language pages than among pages with no English detected, and the difference has widened over time. In the August 2026 snapshot, \Result{language.level.english.latest.mean}\% (\Result{language.level.english.latest.lo95}--\Result{language.level.english.latest.hi95}\%) of English-only pages were estimated to contain \Result{outcome.definition}, against \Result{language.level.nonenglish.latest.mean}\% (\Result{language.level.nonenglish.latest.lo95}--\Result{language.level.nonenglish.latest.hi95}\%) of non-English-only pages. The non-English-only category corresponds to all non-English languages in the sample. Prevalence among English-only pages exceeded that among non-English-only pages by \Result{language.contrast.english_vs_nonenglish.start2023.pp.mean} (\Result{language.contrast.english_vs_nonenglish.start2023.pp.lo95}--\Result{language.contrast.english_vs_nonenglish.start2023.pp.hi95}) percentage points in February 2023, \Result{language.contrast.english_vs_nonenglish.start2025.pp.mean} (\Result{language.contrast.english_vs_nonenglish.start2025.pp.lo95}--\Result{language.contrast.english_vs_nonenglish.start2025.pp.hi95}) in January 2025 and \Result{language.contrast.english_vs_nonenglish.latest.pp.mean} (\Result{language.contrast.english_vs_nonenglish.latest.pp.lo95}--\Result{language.contrast.english_vs_nonenglish.latest.pp.hi95}) in August 2026, a prevalence ratio of \Result{language.contrast.english_vs_nonenglish.latest.ratio.mean} (\Result{language.contrast.english_vs_nonenglish.latest.ratio.lo95}--\Result{language.contrast.english_vs_nonenglish.latest.ratio.hi95}; Figure~\hyperref[fig:language-prevalence]{\ref*{fig:language-prevalence}{a}}). Moreover, English-only prevalence increased faster than non-English-only prevalence both between February 2023 and January 2025 and between January 2025 and August 2026, and the difference in growth rates doubled from the first interval to the second (Figure~\hyperref[fig:language-prevalence]{\ref*{fig:language-prevalence}{b}}; Figure~\ref{fig:language-growth-gap} in the Appendix). In the first interval, English-only prevalence rose by \Result{language.rate.english.early.mean} (\Result{language.rate.english.early.lo95}--\Result{language.rate.english.early.hi95}) percentage points per year, against \Result{language.rate.nonenglish.early.mean} (\Result{language.rate.nonenglish.early.lo95}--\Result{language.rate.nonenglish.early.hi95}) for non-English-only pages; in the second interval, the rates were \Result{language.rate.english.late.mean} (\Result{language.rate.english.late.lo95}--\Result{language.rate.english.late.hi95}) and \Result{language.rate.nonenglish.late.mean} (\Result{language.rate.nonenglish.late.lo95}--\Result{language.rate.nonenglish.late.hi95}). The annual increase in English-only prevalence therefore exceeded that in non-English-only prevalence by \Result{language.rate.gap.early.mean} (\Result{language.rate.gap.early.lo95}--\Result{language.rate.gap.early.hi95}) percentage points per year in the first interval and by \Result{language.rate.gap.late.mean} (\Result{language.rate.gap.late.lo95}--\Result{language.rate.gap.late.hi95}) in the second. Bilingual pages (English and at least another language) showed a different pattern. Their prevalence rose by \Result{language.rate.bilingual.early.mean} percentage points per year (\Result{language.rate.bilingual.early.lo95}--\Result{language.rate.bilingual.early.hi95}) between February 2023 and January 2025 and by \Result{language.rate.bilingual.late.mean} (\Result{language.rate.bilingual.late.lo95}--\Result{language.rate.bilingual.late.hi95}) between January 2025 and August 2026, close to the English-only rate of \Result{language.rate.english.late.mean} in the second interval and the largest change in annual rate of the three groups (\Result{language.rate.bilingual.change.mean} percentage points per year, \Result{language.rate.bilingual.change.lo95}--\Result{language.rate.bilingual.change.hi95}). Bilingual prevalence was \Result{language.contrast.bilingual_vs_nonenglish.start2023.abs} percentage points below non-English-only prevalence in February 2023 and \Result{language.contrast.bilingual_vs_nonenglish.start2025.abs} below in January 2025, but exceeded it by \Result{language.contrast.bilingual_vs_nonenglish.latest.abs} (\Result{language.contrast.bilingual_vs_nonenglish.latest.pp.lo95}--\Result{language.contrast.bilingual_vs_nonenglish.latest.pp.hi95}) percentage points in August 2026, at \Result{language.level.bilingual.latest.mean}\% (\Result{language.level.bilingual.latest.lo95}--\Result{language.level.bilingual.latest.hi95}\%). See Appendices~\ref{sec:lang_detail_app} and~\ref{sec:sensitivity_app} for details.

These contrasts are not attributable to differences in composition between the language groups. At a common composition of centrality tier, TLD and length, prevalence in August 2026 is \Result{language.std.english.latest}\% for English-only pages, \Result{language.std.nonenglish.latest}\% for non-English-only pages and \Result{language.std.bilingual.latest}\% for bilingual pages (Appendix~\ref{sec:lang_detail_app}). The widening of the difference in growth rates is likewise not attributable to composition. We restrict the comparison to the \Result{language.shared.n_profiles} profiles, that is, combinations of centrality tier, TLD and length, that are observed for every language group at every endpoint, which hold \Result{language.shared.mass}\% of the page mass of the six most recent crawls, and assign every group the average composition of those crawls. On these profiles, the change in the difference in annual growth between English-only and non-English-only pages, later interval minus earlier, is \Result{language.rate.gapchange.shared_common.mean} percentage points per year (\Result{language.rate.gapchange.shared_common.lo95}--\Result{language.rate.gapchange.shared_common.hi95}) at the common composition, and \Result{language.rate.gapchange.shared_own.mean} (\Result{language.rate.gapchange.shared_own.lo95}--\Result{language.rate.gapchange.shared_own.hi95}) when each group keeps its own composition, against \Result{language.rate.gapchange.all.mean} (\Result{language.rate.gapchange.all.lo95}--\Result{language.rate.gapchange.all.hi95}) over all pages (Figure~\ref{fig:language-growth-gap} in the appendix). The estimates are similar and their intervals exclude zero in all three cases. 

\subsection{Prevalence of AI-generated text is lower on government and educational domains, but rising} 
\begin{figure}[!htbp]
  \centering
  \includegraphics[width=.9\linewidth]{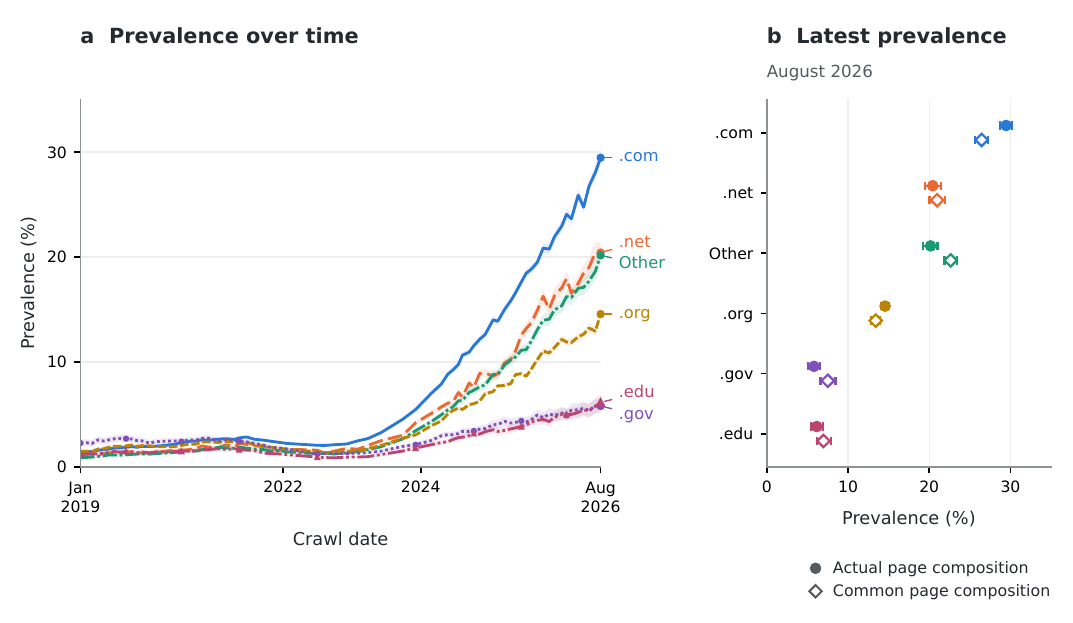}
  \caption{\small\textbf{Estimated prevalence of pages in Common Crawl with \Result{outcome.definition} over time and across domain categories at a common page composition.} \textbf{(a)} Estimated proportions of eligible archived pages containing fully AI-generated or AI-edited text, with posterior means and pointwise 95\% credible intervals. Eligible pages contain at least 500 extracted text characters. Dates are crawl midpoints; prevalence is weighted by modeled page counts. \textbf{(b)} Prevalence in \Result{domain.date.latest} under each category's actual page composition (filled circles) and a common composition (open diamonds), with 95\% credible intervals. Both series are model estimates. The common composition is the final crawl's page-count distribution of centrality, language, and length, summed across all six categories; identical weights are used for every category within each posterior draw. Combinations absent from a category are predicted by the fitted model. Represented combinations cover \Result{domain.coverage.gov}\% of the reference distribution for \texttt{.gov} and \Result{domain.coverage.edu}\% for \texttt{.edu}.}
  \label{fig:domain-prevalence}
\end{figure}

Estimated prevalence differs substantially across top-level domain (TLD) categories. We group top-level domains into six categories: \texttt{.com}, \texttt{.edu}, \texttt{.gov}, \texttt{.net}, \texttt{.org}, and all remaining top-level domains (other). Each page inherits the TLD category of the domain to which it belongs; for example, a page whose domain ends in \texttt{.edu} belongs to the \texttt{.edu} category. Hence, prevalence in a TLD category corresponds to the proportion of its eligible pages that contain \Result{outcome.definition}. For example, prevalence in the \texttt{.edu} category is the proportion of eligible pages on domains ending in \texttt{.edu} that contain AI-generated text. In the August 2026 snapshot, \Result{domain.latest.com.mean}\% (\Result{domain.latest.com.lo95}--\Result{domain.latest.com.hi95}\%) of \texttt{.com} pages were estimated to be AI-generated, compared with \Result{domain.latest.net.mean}\% of \texttt{.net} pages, \Result{domain.latest.org.mean}\% of \texttt{.org} pages and \Result{domain.latest.edu.mean}\% and \Result{domain.latest.gov.mean}\% of \texttt{.edu} and \texttt{.gov} pages (Figure~\hyperref[fig:domain-prevalence]{\ref*{fig:domain-prevalence}{a}}). Prevalence in the institutional categories also started later and increased more slowly: in \texttt{.edu} and \texttt{.gov}, it was only \Result{domain.earlier.edu.mean}\% (\Result{domain.earlier.edu.lo95}--\Result{domain.earlier.edu.hi95}\%) and \Result{domain.earlier.gov.mean}\% (\Result{domain.earlier.gov.lo95}--\Result{domain.earlier.gov.hi95}\%) as late as \Result{domain.date.earlier}, while \texttt{.com} had reached \Result{domain.earlier.com.mean}\% (\Result{domain.earlier.com.lo95}--\Result{domain.earlier.com.hi95}\%) by then. Between \Result{domain.date.baseline} and \Result{domain.date.latest} prevalence in \texttt{.edu} and \texttt{.gov} rose by only \Result{domain.growth.edu.mean} and \Result{domain.growth.gov.mean} percentage points in these two categories, respectively, against \Result{domain.growth.com.mean} in \texttt{.com}.

These differences are only partially driven by composition (Figure~\hyperref[fig:domain-prevalence]{\ref*{fig:domain-prevalence}{b}}). When every top-level domain category is assigned the same distribution of centrality, language and length, \texttt{.com} falls from \Result{domain.latest.com.mean}\% to \Result{domain.standardized.com.mean}\% (\Result{domain.standardized.com.lo95}--\Result{domain.standardized.com.hi95}\%), \texttt{.org} from \Result{domain.latest.org.mean}\% to \Result{domain.standardized.org.mean}\%, and \texttt{.gov} and \texttt{.edu} rise to \Result{domain.standardized.gov.mean}\% and \Result{domain.standardized.edu.mean}\%. For \texttt{.com} and \texttt{.org}, most of this shift comes from language. Adjusting for language alone lowers \texttt{.com} by \Result{composition.tld.com.adj_lang.abs} percentage points and \texttt{.org} by \Result{composition.tld.org.adj_lang.abs}, because their pages are more often in languages with higher prevalence, such as English. However, this composition effect is small relative to the differences themselves. The gap between \texttt{.com} and \texttt{.gov} narrows from \Result{domain.gap.gov.actual.mean} to \Result{domain.gap.gov.standardized.mean} percentage points (\Result{domain.gap.gov.standardized.lo95}--\Result{domain.gap.gov.standardized.hi95}), and the gap between \texttt{.com} and \texttt{.edu} from \Result{domain.gap.edu.actual.mean} to \Result{domain.gap.edu.standardized.mean} (\Result{domain.gap.edu.standardized.lo95}--\Result{domain.gap.edu.standardized.hi95}). Nevertheless, after standardization by page characteristics, \texttt{.com} remains the highest top-level domain category and the institutional domains the lowest. The standardized estimates for \texttt{.gov} and \texttt{.edu}, however, rest on model predictions for rare combinations of centrality tier, language, and length, as the represented combinations cover \Result{domain.coverage.gov}\% and \Result{domain.coverage.edu}\% of the common distribution, respectively.

\section{Discussion}
The public-facing web is critical infrastructure for the dissemination of information and the exchange of ideas. webpages inform people (accurately or otherwise) about many topics, deliver practical guidance, market commercial material, and are used by governments to communicate with their citizens. The web is an interconnected ecosystem built on trust, where both individuals and institutions rely on it for information \citep{shah2024envisioning}. Despite the recency of generative AI, this technology is already reshaping how that ecosystem's content is produced. Language models carry the model's own norms and conventions for writing and conveying information. When published to the web, those conventions shape what information people receive and how they receive it. When published by an institution, they are implicitly endorsed as part of how that organization speaks to its customers, stakeholders, or constituents. Establishing how much of the public web's text is now fully AI-generated or AI-edited across what kinds of pages is the first step toward understanding what this shift means for the people and institutions that rely on it. Further, since the crawlable web specifically is used as training data for future models, the composition of the web has implications for how future models will behave, both in terms of the conventions they use for writing, and in terms of the information that they provide.  In the extreme, as the composition of the web shifts towards more AI-generated text, we could reasonably expect to see future models that write increasingly less like humans and more like previous versions of themselves.  

Our framing of AI-detection as a measurement problem carries important implications for how we interpret the outputs. We distinguish observed indicators from the latent constructs they imperfectly capture \citep{bollen1989structural}, a distinction that the fairness literature has brought to bear on computational systems through the lens of construct validity \citep{jacobs2021measurement}. We frame detector outputs as proxies for AI involvement that are then used to assign authorship categories. Neither the measure nor its thresholds (human, AI-assisted, entirely AI) are equivalent to the underlying reality of authorship, which will only become more difficult to distinguish as detectors improve and AI-generated text becomes more common. Moreover, detectors operationalize AI involvement differently. Some measure text predictability; others, such as supervised classifiers, identify patterns in labeled examples; and informal human judgments rely on simple heuristics -- like em dashes. Given that each approach has blind spots, triangulation across distinct methods is both a principled measurement strategy and also a reminder that any given flag supports a much narrower claim than latent authorship. This is visible especially on a benchmark that records the known share of AI-generated tokens in each document, detectors miss some fully AI-generated text and flag some text that is mostly human, and they do so in different places and for different reasons, because each is measuring something slightly different (Figure~\ref{fig:aitdna_token_share}). A detector score is a type of stylized fact, a high-level summary of a much more complex phenomenon that focuses attention and shapes public debate and institutional judgments~\citep{hirschman2016stylized}. The stylized fact of a detector score proxies for \emph{evidence about authorship}, with real consequences. In October 2026, days after awarding its prize, the Caine Prize for African Writing placed the winning story under review after a journalist reported that a commercial AI-detector had flagged nearly all of it as AI-generated\footnote{\url{https://theatlantic.com/books/2026/10/caine-prize-winning-short-story-seems-be-ai/688967/}}. The author denies using AI, and no conclusive determination has been issued at the time of this writing. Treating detector outputs with undue epistemic authority risks turning uncertain measurements into consequential judgments about authorship, reputation, and professional legitimacy.

\section{Methods}

\subsection{Sampling}
\label{sec:sampling}

``The web'' is vast, changes constantly, and is not an enumerable population at any point in time. The sampling strategy, therefore, is central to what we measure. We take Common Crawl (\url{https://commoncrawl.org/}) as the best available operationalization of the crawlable web. Its coverage is neither complete nor uniform (see Limitations in Section~\ref{sec:limitations}), but this is also true of other publicly available alternatives, none of which supports a probability design. Since 2008 the \cite{CC_webgraph_stats} has regularly scanned the public web and published each scan as an open archive of page text and metadata, together with a URL index and a host- and domain- level webgraph. 

Three properties make Common Crawl a good candidate for operationalizing web content in this study. First, it represents a sampling frame to begin with. The per-crawl URL index enumerates all of the pages in a snapshot, so pages can be drawn with known inclusion probabilities, and the web graph supplies various centrality measures our exposure weighted estimand requires. Second, it is longitudinal, which lets us calibrate detector error rates on pre-ChatGPT crawls drawn from the same frame. This also allows us to compare the recent past to the present and carry on this analysis into the future as more crawls are published. Third, filtered text derived from the raw archive is used to pretrain and finetune the language models most widely used today \citep{wenzek2019ccnet, baack2024critical}, so its composition could have substantial implications for model collapse, depending on how it is weighted in the training process. 

We rely upon three specific artifacts for our analysis: Web ARChive (WARC) files holding per-page content, a columnar index which locates each capture across all crawls, and a domain-level hyperlink graph carrying centrality measures for each registered domain. Our primary unit of analysis is a crawl, which is a distinct window with a median duration of 13.1 days. We use all 71 crawls published from January 2019 to August 2026. The estimand's denominator is extractable-text pages, which must meet two criteria: an eligible page is (i) a page that Common Crawl returned successfully, and (ii) from which at least 500 characters of cleaned text were recovered by the text extraction tool \texttt{trafilatura}~\citep{barbaresi-2021-trafilatura}. Across the archive, 112.7 billion of 194.9 billion Common Crawl page captures (58\%) are eligible; the share per crawl ranges from 54.7\% to 63.1\% (Figure~\ref{fig:descriptives}). Non-text captures (too short, non-parsable text, or both) are excluded from the denominator, as the notion of distinguishing human- from fully AI-generated or AI-edited content is not meaningful for them. 

\subsubsection{Stratification and Sampling Allocation}

We draw 20,000 domains from each crawl in six strata cut at fixed quantiles of that crawl's PageRank ranking: (0,$10^{-5}$], ($10^{-5}$ ,$10^{-4}$], ($10^{-4}$,$10^{-3}$], ($10^{-3}$,$10^{-2}$], ($10^{-2}$,$10^{-1}$] and ($10^{-1}$,1]. We label these strata top, upper-middle, middle, lower-middle, bottom, and tail, respectively. We use quantiles as opposed to absolute rank cut-offs because the frame is not of fixed size over time, so our approach covers the same fractions of the frame across crawls.  We use PageRank for stratification because we know the distribution of PageRank to be highly skewed and we want to ensure we have a sufficient quantity of high PageRank pages for meaningful comparisons, which would be difficult to obtain by chance in reasonably sized samples. 

We could allocate samples based on the proportion of pages that fall within each tier, or based on the overall mass of PageRank in each tier.  Since the PageRank distribution is so skewed, these would yield different allocation strategies. The top stratum is about 900 domains but carries 30.2\% of all PageRank mass in our sample, for example, while the tail (lowest PageRank) contains about 38\% of all extractable-text pages.  We opt for a compromise and allocate $n_j \propto \sqrt{s_j}$ domains to stratum $j$, where $s_j$ denotes the mean of the page- and exposure- shares,
\begin{equation}
\label{eq:alloc}
  s_j = \tfrac{1}{2} s^{\text{page}}_j + \tfrac{1}{2} s^{\text{exp}}_j \;,
\end{equation}
with at least 700 draws per stratum so that every stratum remains usable as a post-stratification cell once crossed with top-level domain (TLD). By allocating according to $\sqrt{s_j}$ we intend to simultaneously improve the depth of sampling in the top centrality tier and the breadth of sampling in the bottom centrality tier. Because the top stratum is smaller than its nominal allocation, it is captured in full for all crawls, yielding a census with $\pi = 1$ in the domains that most influence the exposure estimand. Unused allocation in the top is redistributed to other strata in proportion to their nominal allocations. The stage-one inclusion probability is $\pi^{\text{dom}}_d = n_j / N_j$ with both quantities realized and recorded per row. Within a sampled domain, pages are drawn by reservoir sampling as the crawl index is scanned, which gives a simple random sample of the domain's captures and stage-two probability of $\pi^{\text{url}}_d = m_d / M_d$. The
page-level design weight is the product, $\pi_i = \pi^{\text{dom}} \cdot \pi^{\text{url}}$.

Population totals for poststratification come from a separate counting sample. PageRank mass is exact, as it is a property of domains and can be summed over the entire frame. Page counts need to be estimated in some cases, as fetching every page in a crawl is prohibitively expensive. The page count population totals for the three top strata are censused. For lower-middle, bottom and tail strata, page counts are estimated with a design-based standard error (the median relative SE on a crawl's extractable-text page count is 1.3pp). 
\begin{table}[t]
\centering
\caption{\textbf{Sampling strata and the two estimand denominators.} Strata are cut at
fixed quantiles of each crawl's PageRank ranking.}
\label{tab:strata}
\small
\setlength{\tabcolsep}{5pt}
\begin{threeparttable}
\begin{tabular}{@{}llrcrrr@{}}
\toprule
& PageRank & Nominal & & \multicolumn{2}{c}{Share of (\%)} & Extractable \\
\cmidrule(lr){5-6}
Stratum & quantile & allocation & Census & Pages & Exposure & rate \\
\midrule
Top          & $(0,10^{-5}]$        & 3,386 & \checkmark &  4.2 & 30.2 & 0.608 \\
Upper-middle  & $(10^{-5},10^{-4}]$  & 2,370 & \checkmark &  5.1 & 10.1 & 0.562 \\
Middle        & $(10^{-4},10^{-3}]$  & 2,740 & \checkmark &  9.4 & 11.2 & 0.584 \\
Lower-middle  & $(10^{-3},10^{-2}]$  & 3,000 &            & 15.0 & 11.8 & 0.617 \\
Bottom        & $(10^{-2},10^{-1}]$  & 3,748 &            & 28.3 & 15.4 & 0.610 \\
Tail          & $(10^{-1},1]$        & 4,756 &            & 38.0 & 21.3 & 0.561 \\
\bottomrule
\end{tabular}
\begin{tablenotes}[flushleft]\footnotesize
\item \emph{Nominal allocation} is the stratum's share of $n=20{,}000$ draws under
Equation~\ref{eq:alloc}, before the frame is consulted. It is not the realized draw for the top stratum, which holds only ${\sim}900$ domains against a nominal 3{,}386: all are taken ($\pi=1$) and the unused allocation is redistributed to the strata below in proportion to their nominal allocation. Realized counts for all six strata are recorded per crawl in the released sample. 
\item \emph{Share of pages} $s^{\text{page}}_j$ and \emph{share of exposure} $s^{\text{exp}}_j$ are each stratum $j$'s share of extractable-text pages and of PageRank mass on those pages, pooled over 71 crawls. Totals are exact for the three censused strata and estimated below them. 
\item \emph{Extractable rate} is the design-weighted share of text-MIME pages passing the 500-character screen, averaged over crawls.
\end{tablenotes}
\end{threeparttable}
\end{table}

\subsubsection{Data Descriptives}\label{sec:data_desc}

\begin{figure}[!htbp]
\centering
\includegraphics[width=0.9\linewidth]{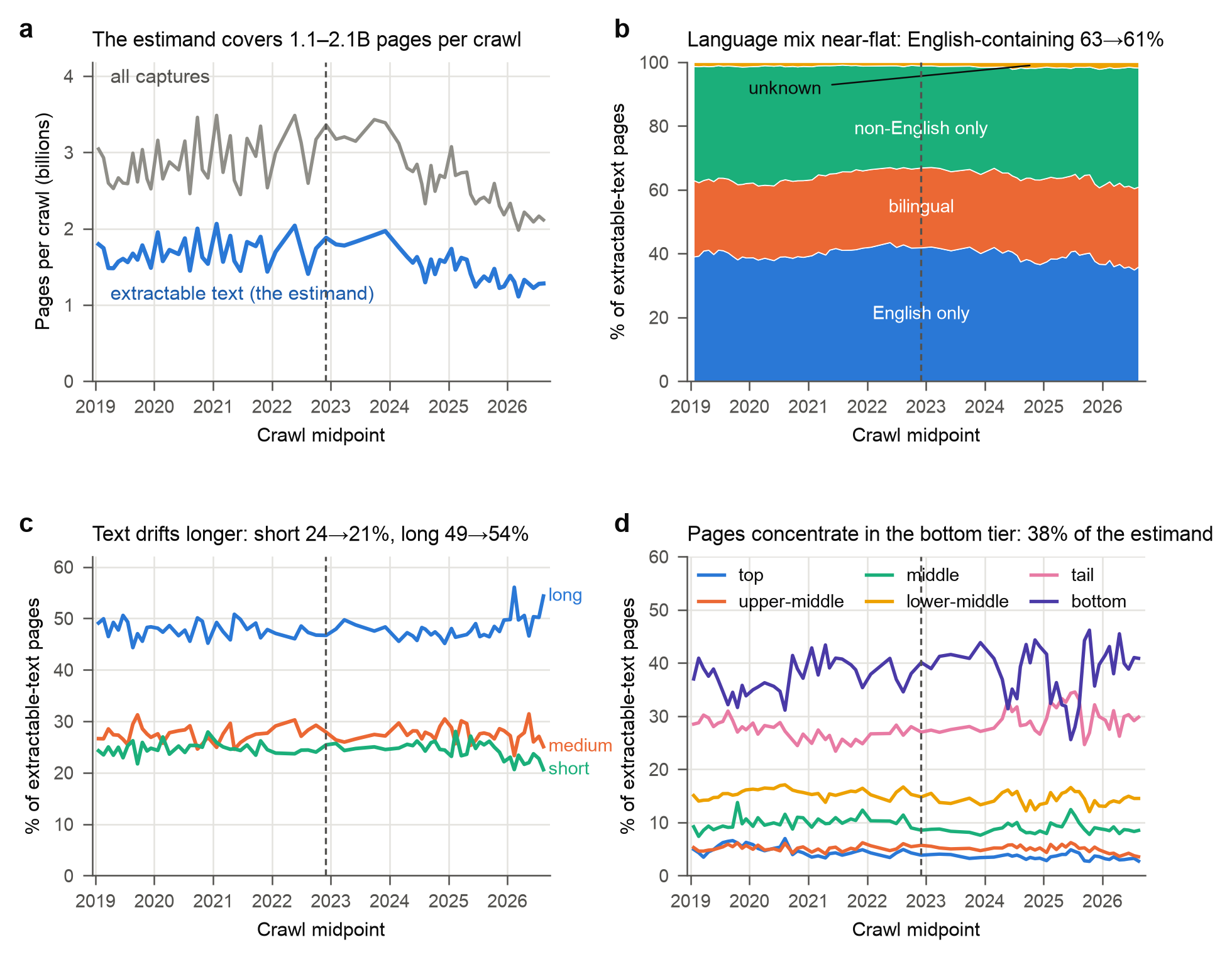}
\caption{This figure describes the full sample drawn from 71 Common Crawls. All quantities are population estimates formed from the per-crawl cell denominators, which scale the sample up by its inclusion probabilities; the dashed line marks ChatGPT's public launch (2022-11-30). \textbf{(a)}~Captures per crawl and the subset passing the 500-character extractable-text screen, which is the denominator of every rate we report. \textbf{(b)}~Language composition of that denominator over time as a four-way partition. \textbf{(c)}~Length composition on the word-count bins against which the primary detector's error rates are published. \textbf{(d)}~Share of extractable-text pages by PageRank stratum. Panels (b)--(d) are the three axes the measurement model conditions on, and the drift in (c) matters independently of prevalence: detectors are weakest on short text, so a shift toward longer pages raises measured sensitivity over the series.}
\label{fig:descriptives}
\end{figure}

Figure~\ref{fig:descriptives} summarizes the archive over which the estimates are taken. Across the 71 crawls it covers an average of 2.74 billion captures per crawl, of which an average of 1.59 billion pass the 500-character screen (range 1.11--2.06 billion, Figure~\ref{fig:descriptives}a). This results in a per-crawl average of 72,000 sampled pages, which are scored by our ensemble of AI detectors.

Common Crawl metadata includes a language indicator for each page for each $\text{language} \in \{\texttt{english\_only}, \texttt{bilingual}, \texttt{non\_english\_only}, \texttt{unknown}\}$. Between the first and last crawl we observe moderate drift: \texttt{english\_only} moves $39.0 \to 35.9$\%, \texttt{bilingual} $23.9 \to 25.0$\%, \texttt{non\_english\_only} $35.8 \to 37.3$\%, and \texttt{unknown} $1.2 \to 1.7$\%. The \texttt{bilingual} texts are evaluated by the English detectors, so the English fraction is $63.0 \to 61.0$\% of the estimand --- roughly three-fifths English, two-fifths not, throughout.

We bin text-length into short, medium and long categories with cuts 50-149, 150-299 and 300+ words, against which the primary text detector's published error rates are reported. The composition drifts towards longer page lengths over the study period (Figure~\ref{fig:descriptives}c): \emph{short} $24.3 \to 20.6$\%, \emph{medium} $26.6 \to 25.1$\%, \emph{long} $49.1 \to 54.3$\%. The direction matters because detectors are weakest on short text, such that a drift toward longer pages raises measured sensitivity over the series independently of any change in prevalence. We condition on length during the modeling stage for this reason.

The distribution of pages across centrality strata (Figure~\ref{fig:descriptives}d) differs from the distribution of PageRank mass, so the two estimands are given different weighting; the tail stratum holds 38.0\% of extractable-text pages against 21.3\% of PageRank mass, while the top holds 4.2\% of pages against 30.2\% of mass (Table~\ref{tab:strata}).

\subsection{AI Detection}
\label{detection}

We evaluated five detectors from three families on labeled benchmarks before moving forward with scoring the full sample. Two are open-weight checkpoints released with \emph{EditLens} \citep{thai2026editlens}. They are supervised classifers over ordinal levels of AI involvement (human, AI-edited, fully AI-generated): a fine-tuned RoBERTa-large (355M parameters) and Pangram's open model, a QLoRA adapter on Llama-3.2-3B. Two additional detectors use Binoculars \citep{hans2024spotting}, a zero-shot method that scores a text by the ratio of its perplexity to its cross-perplexity under an observer--performer pair of language models. We tried both Falcon-7B and Qwen2.5-7B as base models. The fifth is Pangram~4 \citep{glickenhaus2026pangram4technicalreport}, a commercial supervised detector accessed through a paid API. The open detectors operate at one threshold per language stratum, set to a 1\% false-positive rate on human text from the EditLens validation split (English) and from a held-out half of the multilingual benchmark (non-English).  We used a single non-English threshold because 55.5\% of the sample's non-English pages are in languages that appear in none of our validation sets, so further resolution in the calibration would not be possible. Pangram~4 is used as its own document-level call, counting a text as flagged when it is flagged as either fully AI-generated or mixed AI-use. We dropped the RoBERTa checkpoint, which the Pangram open model matched or outperformed on every out-of-domain benchmark and which detected almost no non-English AI-generated text (sensitivity $0.005$), and Binoculars-Qwen, whose English AUROC was $0.702$ and whose non-English sensitivity was $0.276$ at the pooled threshold. The prevalence model therefore uses the open Pangram model and Binoculars-Falcon. Pangram~4 is excluded because its errors correlate strongly on those of the open Pangram model (Section~\ref{sec:cond_indep}), and its cost limits it to validating the calibration approach (Section~\ref{sec:pangram4_validation}).

\subsubsection{Detectors used in this study} 
\label{sec:detectors_used} 
Two detectors score every page in the sample, routed by language stratum. The primary detector is the open Pangram checkpoint on Llama-3.2-3B, which we run on all \Result{sample.n_pages} text-extractable pages. For English pages it emits three classes (human, AI-edited and AI-generated). For non-English pages, where no labeled AI-edited text exists to support a middle class, it emits a binary call. We chose it because it is the most sensitive open detector able to detect text with partial AI involvement, flagging $94.2\%$ of the co-written texts in AITDNA against $34.1\%$ for Binoculars-Falcon; it is the only open detector with useful non-English sensitivity ($0.363$, against $0.151$ for Binoculars-Falcon); it ranks AI-generated text above human text most accurately on RAID (AUROC $0.939$, against $0.871$ for the next best open detector); and it runs at roughly three times the throughput of Binoculars-Falcon on the archive ($30.1$ against $10.6$ texts per second). On fully AI-generated English text, its sensitivity at the deployed threshold is no higher than that of Binoculars-Falcon ($0.720$ against $0.727$ on RAID).

The second detector is Binoculars with the Falcon-7B pair, which we run on English pages only. It is less sensitive than the primary detector to partially AI-generated text, but as a zero-shot perplexity method it shares no training data or architecture with a supervised classifier, which makes it the natural second indicator for a measurement model that assumes conditionally independent errors (Section~\ref{sec:cond_indep}). Its non-English sensitivity of $0.151$ does not justify the compute to score non-English pages at scale. 

\subsubsection{Empirical overlap of detector errors}\label{sec:cond_indep}
In the Bayesian hierarchical model (Section~\ref{sec:modeling}), we assume that detector outputs are conditionally independent given a page's true authorship status, human-written or AI-involved. That is, among human-written pages, the probability that two detectors both flag a page equals the product of their individual false-positive rates, and among AI-involved pages, the probability that both flag it equals the product of their individual sensitivities. We test both by comparing the observed joint flag rate with this product (Table~\ref{tab:detector_dependence}, for each pair of detectors on human-written and on AI-involved pages).

On human text the assumption holds. Across the 1.77 million English pages in the control crawls (Section~\ref{sec:calib_datasets}), where every flag is a false positive, the open Pangram model and Binoculars-Falcon flag $5{,}278$ pages in common against $5{,}422$ expected under independence (Pearson's $r = -0.002$). On AI-involved text their misses are moderately correlated: $r = 0.27$ on RAID ($0.26$ within text-length bins) and $0.18$ on AITDNA. Because the model treats a page that both detectors miss as stronger evidence of human authorship the model is biased toward \emph{understating} prevalence. 

Pangram~4 fails the assumption on AI-involved text. Unsurprisingly, its misses overlap with the open Pangram model's far more than with Binoculars-Falcon's: $\phi = 0.43$ on RAID ($0.45$ within length bins), $0.35$ on non-English text and $0.56$ on AITDNA. On RAID, Pangram~4 misses $36\%$ of the texts missed by the open model but misses only $4\%$ of the texts identified by the open model. Both detectors come from the same developer, and could share detection strategy or training data. Entering Pangram~4 as a third conditionally independent indicator would count the same evidence twice, overstating the information in concordant calls and narrowing the posterior without justification. We therefore exclude it from the prevalence model.

\subsubsection{Pangram~4 as a validation reference} 
\label{sec:pangram4_validation} 
Pangram~4 is billed per word, so we ran it on a probability subsample of the control period rather than the full sample: 670 pages drawn at random from each of the 30 control crawls (20,100 pages), with each input capped at 750 words. We also ran it on the RAID and multilingual benchmarks in full, on the 2,031 AI-edited texts of the EditLens test split, and on AITDNA. 

These results validate our calibration approach in two ways.
First, the results support treating the open model's flags in the control crawls as false positives, that is, as detector error rather than early machine text. Had those pages contained AI-generated text, Pangram~4 would have flagged many of them as well. It flagged almost none: of the 20,100 control pages it flagged only 2, neither among the 1,593 the open model flagged, and on those 1,593 pages it assigned no segment to AI or AI-assisted writing.

Second, they show that the open model's low non-English sensitivity is a property of that detector rather than of the benchmark. Pangram~4 reaches a sensitivity of $0.708$ on the same non-English texts for which the open model has sensitivity $0.363$, implying the benchmark contains detectable machine text that the open model misses. This supports treating the open model's non-English sensitivity as a floor (Section~\ref{sec:calib_datasets}) and its error rates as measured on, not limited by, the benchmarks.

\subsubsection{Datasets for calibrating false-negative and false-positive rates} 
\label{sec:calib_datasets} 
We measure false-positive and false-negative rates on different data, as false positives depend on the kind of human text a detector sees, while false negatives depend on the generator that produced the machine text.

\paragraph{False positives.} We treat text from language models released before the GPT-3.5 series as negligible on the web. Earlier generators such as GPT-2 and GPT-3 did produce some web text before 2022, but adoption was not widespread. Any such text in the control crawls is counted as a false positive, which overstates the false-positive rate and so biases prevalence estimates downward. Crawls with a midpoint before January 2022 (30 crawls, \texttt{CC-MAIN-2019-04} to \texttt{CC-MAIN-2021-49}) form the control period, in which flags are expected to be predominantly false positive. The remaining 41 crawls from the post-control period, 36 of which follow ChatGPT's release.

Because the control crawls are drawn from the same population as the measurement crawls, the false-positive rate is measured on the study population itself, and it can be resolved by the same cells used for poststratification. These cells differ substantially. For the open Pangram model, the control false-positive rate is $4.5\%$ on English-only pages and $28.5\%$ on bilingual pages, and on English pages it ranges from $9.0\%$ in the middle tier to $12.9\%$ in the top tier. These rates are higher than the 1\% false-positive rate at which the thresholds were set on benchmark human text. False positives also cluster within domains, consistent with the detector's being triggered by site templates rather than text content. Section~\ref{sec:modeling} specifies how these cells inform the modeling.

\paragraph{False negatives.} Measuring false negatives requires labeled AI-generated text, which is not readily available in our context. We therefore draw on four labeled corpora that cover complementary generators, domains and authorship settings. EditLens provides controlled pairs of human-written and fully AI-generated or AI-edited text, with outputs from GPT-4.1, Claude Sonnet 4, Gemini 2.5 Flash and Llama-3.3-70B. \emph{RAID} \citep{dugan2024} spans many generators and domains, largely from the 2023-era generation landscape. \emph{The Multilingual GenAI Content Detection corpus} \citep{wang2025genai} adds non-English human and AI-generated text across student essays, scientific papers, peer reviews, news and social media, with machine text from 2023-era models. \emph{AITDNA} \citep{dycke2026youraitextmine} provides 353 English essays and scientific texts from recruited authors: 95 written without AI assistance and 258 co-written with current language models, recording the share of tokens each model produced (median $77\%$). Unlike the other corpora, AITDNA measures human--AI collaboration rather than fully AI-generated output. All five detectors were scored on RAID and the multilingual corpus; the four open detectors were scored on the EditLens splits; and the two deployed detectors and Pangram~4 were scored on AITDNA. AITDNA is used to compare the detectors on partial AI involvement and does not enter the sensitivity likelihood, which is based on EditLens, RAID and the multilingual corpus. Figure~\ref{fig:benchmarks} compares the detectors on these corpora, and Appendix~\ref{appendix:detector_benchmarks} breaks down the results by text length, language, and degree of AI involvement.

\paragraph{Incorporating statistical properties into the model.} First, if an evaluation set overlaps with a detector's training data, we excluded it from that detector's sensitivity. The open Pangram checkpoints were fine-tuned on the EditLens training split, so their near-perfect results on the EditLens test splits are not a general sensitivity. For fully AI-generated English text, dropping those rows moves the open model's count-weighted sensitivity from $0.899$ to $0.725$; since sensitivity sits in the denominator of the prevalence correction, keeping them would understate prevalence by roughly 20\%. The exclusion is made per detector and evaluation set, not per evaluation set: EditLens is out-of-domain for Binoculars, which is not trained on labeled data.

Second, to retain signal from the disagreement across benchmarks, we fit a hierarchical prior over the per-source rates rather than pooling them into one number. The disagreement is large for both detectors: across out-of-domain sources, English sensitivity to any AI involvement ranges from $0.660$ to $1.000$ for the open Pangram model and from $0.363$ to $0.727$ for Binoculars-Falcon. The webpages themselves imply a lower sensitivity for Binoculars than these benchmarks, as discussed in Section~\ref{sec:calibration-prior} and Appendix~\ref{sec:youden_constraint}.

Third, we treat non-English sensitivity as a pooled estimate. The available labeled benchmarks do not provide sufficiently granular coverage across the non-English languages in our sample to estimate sensitivity separately for each language. Moreover, the available multilingual corpus covers only a subset of these languages (including Chinese, German, Russian, and Italian, but not French, Dutch, Japanese, or Spanish), and its labeled machine text comes from 2023-era generators rather than frontier models. We therefore pool non-English observations rather than impute language-specific performance, reporting a sensitivity of $0.363$ for the pooled group. This estimate should not be interpreted as representative of any individual language.

\subsection{Bayesian measurement error model}
\label{sec:modeling}
In this section, we introduce a Bayesian hierarchical model that estimates the prevalence of \Result{outcome.definition} in pages across the web while accounting for the measurement error inherent in the detectors that were used to score them. In order to calibrate the performance of the detectors, we explicitly model each detector's operating characteristics, i.e., sensitivity (probability of detecting AI given that the page is actually AI-generated) and specificity (probability of detecting human when the page is not AI-generated), which are assumed to be unknown, to be estimated jointly with AI prevalence. Pre-2022 crawls contain, by our assumption, little to no AI-generated text, so the flags they receive are treated as false positives. Beyond informing prevalence like any other crawl, they therefore serve as a negative control that identifies specificity on pages from the same population as later crawls. 

Pages are sampled in crawls $k \in \calK = \{1, \ldots, K\}$, with $K = 71$, from 2019 to 2026, where $t_k$ is the mid-point of crawl $k$, measured in months since the first crawl. At time $t_k$, each sampled page belongs to a cell $c \in \calC$, where $\calC$ denotes the collection of cells that occur in the data. All pages in the same cell and crawl are assumed exchangeable and share the same prevalence $\theta_{c,k}$. Each cell corresponds to a four-way combination of a language level $\ell[c] \in \calL$, a word-count (content length) level $w[c] \in \calW$, a PageRank-based domain centrality tier $j[c] \in \calJ$, and a top-level domain (TLD) $b[c] \in \calB$, shared by every page in $c$, i.e.,
\begin{equation*}
    c = \left( \ell[c], \, w[c], \, j[c], \, b[c] \right) \;.
\end{equation*}
Here, $\calL = \{\text{english-only}, \text{bilingual}, \text{non-english-only}, \text{unknown}\}$ is the language classification of the page, $\calW = \{\text{short}, \text{medium}, \text{long}\}$ is the content length, $\calJ$ = \{\text{top}, \text{upper-mid}, \text{mid}, \text{lower-mid}, \text{bottom}, \text{tail}\} is the domain centrality tiers based on pagerank, and $\calB$ = \{\texttt{.com}, \texttt{.edu}, \texttt{.gov}, \texttt{.net}, \texttt{.org}, \text{other}\} is the top-level domain. Suppose page $i$ belongs to the cell $c[i]$ and was sampled in crawl $k[i]$, for $i = 1, \ldots, n$. And, for each $i$, page $i$ inherits the attributes of its cell $c[i]$; the language level, word-count level, tier and TLD of page $i$ are given by $\ell[i] = \ell[c[i]]$, $w[i] = w[c[i]]$, $j[i] = j[c[i]]$ and $b[i] = b[c[i]]$, respectively.

Let $\calD = \{\text{open-pangram}, \text{binoculars-falcon}\}$ denote the collection of AI detectors, namely the open Pangram checkpoint fine-tuned on Llama-3.2-3B, which is run on all pages, and Binoculars with the Falcon-7B pair, which is run on english-stratum pages only, as detailed in Section~\ref{detection}. For each page $i$, let $\calD[i] \subseteq \calD$ be the subset of detectors that scored it, and, for $d \in \calD[i]$, let $Y_{id} \in \{0, 1\}$ be the detector output, with $Y_{id} = 1$ indicating that detector $d$ flags page $i$ as \Result{outcome.definition}. Let $Z_i \in \{0, 1\}$ be the latent true status of page $i$, with $Z_i = 1$ indicating that the page is indeed \Result{outcome.definition}. For each detector $d$, cell $c$, and crawl $k$, we denote its operating characteristics by $\psi_{d,c, k} = (\psi^{\Se}_{d, c, k}, \psi^{\Sp}_{d, c, k})$, the sensitivity and specificity, respectively, on pages in the cell-crawl pair $(c, k)$; how they depend on the cell and the crawl is specified in Section~\ref{sec:calibration-prior}. The detectors are routed by the language stratum $s[\ell[c]] \in \calS$, where $\calS = \{\text{English}, \text{non-English}\}$, which is coarser than the language levels in $\calL$, with $s[\ell] = \text{English}$ for bilingual and English-only pages and $s[\ell] = \text{non-English}$ for non-English-only and unknown pages, i.e., bilingual pages are sent to the English detectors. 

Following the definitions of sensitivity and specificity, we model the detector outputs $\{Y_{id} : d \in \calD[i]\}$ as conditionally independent given $Z_i$ as
\begin{equation}\label{eq:data-model}
\begin{split}
    Y_{id} \given Z_i, \psi_{d, c[i], k[i]} &\sim
    \begin{cases}
        \Ber \left( \psi^{\Se}_{d, c[i], k[i]} \right), & Z_i = 1\\
        \Ber \left( 1 - \psi^{\Sp}_{d, c[i], k[i]} \right), & Z_i = 0
    \end{cases}
    \;, \quad \forall d \in \calD[i] \;,\\
    Z_i \given \theta_{c[i], k[i]} &\sim \Ber (\theta_{c[i], k[i]}) \;.
\end{split}
\end{equation}
Marginalizing out the latent status $Z_i$ yields the mixture
\begin{equation}\label{eq:marginal}
\begin{split}
    p \left(Y_i \given \theta_{c[i], k[i]}, \psi_{c[i], k[i]} \right) &= \ \theta_{c[i], k[i]} \prod_{d \in \calD[i]} \left( \psi^{\Se}_{d, c[i], k[i]} \right)^{Y_{id}} \left( 1 - \psi^{\Se}_{d, c[i], k[i]} \right)^{1 - Y_{id}} \\
    & \qquad + (1 - \theta_{c[i], k[i]}) \prod_{d \in \calD[i]} \left( 1 - \psi^{\Sp}_{d, c[i], k[i]} \right)^{Y_{id}} \left( \psi^{\Sp}_{d, c[i], k[i]} \right)^{1 - Y_{id}} \;,
\end{split}
\end{equation}
where $Y_i = \{ Y_{id} : d \in \calD[i]\}$, and $\psi_{c[i], k[i]} = \{ \psi_{d, c[i], k[i]} : d \in \calD[i] \}$.
\begin{remark}
    Conditional independence of two detectors given $Z_i$ is an assumption; it is credible here because the two are of different kinds (supervised against perplexity-based). See Section~\ref{sec:cond_indep} for the empirical overlap.
\end{remark}

Subsequently, we model the prevalence $\theta_{c, k}$ of cell $c$ at crawl $k$ as
\begin{equation}\label{eq:theta_model}
    \logit(\theta_{c, k}) = \mu + \alpha^{\text{tier}}_{j[c]} + \alpha^{\text{tld}}_{b[c]} + \alpha^{\text{wc}}_{w[c]} + \alpha^{\text{lang}}_{\ell[c]} + \tau_{c, k}, \quad \forall c \in \calC, k \in \calK \;,
\end{equation}
where $\logit(x) = \ln \left( x / (1-x) \right)$ denotes the inverse of the standard logistic function. In \eqref{eq:theta_model}, $\mu$ denotes a global intercept, $\alpha^{\text{tier}}$, $\alpha^{\text{tld}}$, $\alpha^{\text{wc}}$ and $\alpha^{\text{lang}}$ are varying intercepts for centrality tier, TLD, word count and language, respectively, and $\tau_{c, k}$ is a temporal effect capturing the inhomogeneity of AI-prevalence across time, specified in Section~\ref{sec:temporal}. Variants of the model with interactions between pairs of attributes are described in Appendix~\ref{sec:sensitivity_app}. We take English-only, \texttt{.com} and long as the reference baseline levels of language, TLD and word count, respectively, so that $\mu$ is the logit prevalence of the baseline cell in a tier with zero effect at the first crawl and the varying intercepts of language, TLD and word count represent contrasts with respect to it. The tier effects have no baseline level and instead follow the zero-centered autoregressive prior in \eqref{eq:tier_prior}. Since the first crawl is a pre-2022 control crawl, the prevalence of the baseline cell is expected to be small, which we encode with
\begin{equation}\label{eq:mu_prior}
    \mu \sim \Normal \left(\logit(\varepsilon_\mu), s_{\mu}^2 \right), \quad \varepsilon_\mu = 0.01, \ s_{\mu} = 1\;.
\end{equation}
We present rationale and sensitivity analysis for the choice of $\varepsilon_\mu$ and $s_\mu$ in Section~\ref{sec:calibration-prior} and Appendix~\ref{sec:sensitivity_app}, respectively.

\subsubsection{Hierarchical priors on cell attributes}
Since centrality tiers form an ordered sequence in which contiguous tiers are expected to be more similar than distant ones, we follow \cite{gao_mrp_2021} and assign a structured prior to the tier effects as
\begin{equation}\label{eq:tier_prior}
\begin{split}
    \alpha^{\text{tier}}_{1} \given \rho_{\text{tier}}, \sigma_{\text{tier}} &\sim \Normal \left( 0, \sigma^2_{\text{tier}} /(1 - \rho^2_{\text{tier}}) \right) \;, \\
    \alpha^{\text{tier}}_{j} \given \alpha^{\text{tier}}_{j-1}, \rho_{\text{tier}}, \sigma_{\text{tier}} &\sim \Normal \left( \rho_{\text{tier}} \alpha^{\text{tier}}_{j-1}, \sigma^2_{\text{tier}} \right), \quad j = 2, \ldots, 6 \;, \\
    \rho_{\text{tier}} &\sim \Unif \left(0.1, 0.9 \right) \;,
\end{split}
\end{equation}
where $\alpha^{\text{tier}}_{1}$ corresponds to the effect of the most central tier (Top), and $\rho_{\text{tier}}$ controls the degree of borrowing of information between contiguous centrality tiers. Language levels and TLDs have no notion of adjacency, and while the word-count levels are ordered, with only three levels $\rho_{\text{wc}}$ would be weakly identified, so we treat these effects as a priori exchangeable 
\begin{equation}\label{eq:exchangeable_prior}
    \alpha^{\text{tld}}_b \given \sigma_{\text{tld}} \sim \Normal (0, \sigma^2_{\text{tld}}), \quad
    \alpha^{\text{wc}}_{w} \given \sigma_{\text{wc}} \sim \Normal (0, \sigma^2_{\text{wc}}), \quad
    \alpha^{\text{lang}}_{\ell} \given \sigma_{\text{lang}} \sim \Normal (0, \sigma^2_{\text{lang}})\;,
\end{equation}
where $b \in \bar{\calB} = \calB \setminus \{\texttt{.com}\}$, $w \in \bar{\calW} = \calW \setminus \{\text{long}\}$, and $\ell \in \bar{\calL} = \calL \setminus \{\text{English-only}\}$, respectively. Moreover, $\sigma^2_{\text{tld}}$, $\sigma^2_{\text{wc}}$, and $\sigma^2_{\text{lang}}$ denote the variability within top-level domains, languages, and word counts strata respectively. The hierarchy is completed by assigning a half-Normal prior on all the scale parameters
\begin{equation*}
    \sigma_{\text{tier}}, \sigma_{\text{tld}}, \sigma_{\text{wc}}, \sigma_{\text{lang}} \sim \halfNormal (0, 1) \;.
\end{equation*}

\begin{remark}
These random effects induce prior dependence between cells that share attributes. Two cells in the same centrality tier share $\alpha^{\text{tier}}$ and, through the autoregressive prior, are correlated with neighboring tiers; cells with the same TLD, language or word-count level share the corresponding effects. Hence, the more attributes two cells share, the more prior structure they share, and hence the more strongly correlated their prevalences are a priori.
\end{remark}

\subsubsection{Modeling temporal inhomogeneity}\label{sec:temporal}
The preceding sections described the dependence structure between cells within a single crawl. We extend this across crawls through the temporal effect $\tau_{c,k}$ of cell $c$ in crawl $k$,
\begin{equation}\label{eq:tau-model}
     \tau_{c, k} = \left( \beta^{\text{lang}}_{\ell[c]} + \beta^{\text{tier}}_{j[c]} + \beta^{\text{tld}}_{b[c]} \right) t_k + f(t_k), \quad \forall c \in \calC, k \in \calK \;,
\end{equation}
where $t_k$ denotes the timestamp (mid-point) associated with crawl $k$, $f(\cdot)$ is a trend shared by all cells, and $\beta^{\text{lang}}_{\ell}$, $\beta^{\text{tier}}_{j}$, and $\beta^{\text{tld}}_{b}$ are language-, tier-, and TLD- specific rates of change per month. Here $f$ denotes the temporal trend of the reference cell and the remaining slopes measure departures from it. Subsequently, we assign
\begin{equation*}
    \beta^{\text{lang}}_{\ell} \sim \Normal \left( 0, (1/12)^2 \right),\quad 
    \beta^{\text{tier}}_{j} \given \sigma_{\beta, \text{tier}} \sim \Normal \left( 0, \sigma_{\beta, \text{tier}}^2 \right),\quad
    \beta^{\text{tld}}_{b} \given \sigma_{\beta, \text{tld}} \sim \Normal \left( 0, \sigma_{\beta, \text{tld}}^2 \right)\;,
\end{equation*}
for $\ell \in \bar{\calL}$, $j \in \bar{\calJ} = \calJ \setminus \{ \text{mid} \}$, and $b \in \bar{\calB}$. We assign $\sigma_{\beta, \text{tier}}, \sigma_{\beta, \text{tld}} \sim \halfNormal \left( 0, (1/48)^2 \right)$. 

We specify $f$ as a penalized B-spline \citep{deBoor_1978, eilers_marx_1996} on a fixed knot vector $\kappa = (\kappa_0, \ldots, \kappa_{R+3})$,
\begin{equation}
    f(t) = \sum_{r = 1}^{R} \gamma_r \, \zeta_r (t; \kappa)\;,
\end{equation}
where $\{\zeta_r ( \cdot; \kappa )\}_{r = 1}^{R}$ is a cubic B-spline basis generated by $\kappa$ via the standard Cox--de Boor recursion, $\gamma_1 = 0$ to identify $f$ against $\mu$, with the remaining coefficients regularized by a second-order random-walk penalty
\begin{equation}
\begin{split}
    \gamma_2 \sim \Normal (0, 1), \quad
    \gamma_r \given \gamma_{r-1}, \gamma_{r-2}, \sigma_{\gamma} \sim \Normal (2 \gamma_{r-1} - \gamma_{r-2}, \sigma^2_{\gamma}), \ r = 3, \ldots, R \;,
\end{split}
\end{equation}
with $\sigma_{\gamma} \sim \halfNormal(0, 1)$. 
The knot vector $\kappa$ is fixed rather than estimated. Boundary knots are placed at the first and last crawl mid-points, and two interior knots at $\kappa_1 = \text{2022-01-01}$, when the GPT-3.5 series entered circulation, and $\kappa_2 = \text{2022-11-30}$, ChatGPT's public launch. One additional interior knot is placed every 10 crawls within each of the 3 resulting regimes---pre-GPT-3.5 control period $\{ t: t < \kappa_1 \}$, the intermediate epoch $\{ t: t \in[\kappa_1, \kappa_2)\}$, and post-ChatGPT era $\{ t : t \geq \kappa_2 \}$, which gives $R = 12$ basis functions for the 71 crawls.
\begin{remark}
    Although crawl $k$ spans a very short window of time rather than a single instant, this window is negligible relative to the full temporal domain under study, as each crawl lasts on the order of 2 to 3 weeks against a study period of several years. Hence, we represent crawl $k$ through a single representative timestamp $t_{k}$, its mid-point.
\end{remark}

\subsubsection{Prior for calibration of detectors}\label{sec:calibration-prior}

\paragraph{Control period.} Let $\calK_0 = \{k \in \calK : t_k < \kappa_1\}$, with $K_0 = |\calK_0| = 30$, denote the set of control crawls. These crawls precede the public availability of the generative models considered here, so their pages are expected to contain little to no \Result{outcome.definition}, and the flags they receive are therefore expected to be predominantly false positives. The control status is encoded through the informative prior on the global intercept $\mu$ specified above, instead of fixing $\theta_{c, k} = 0$, for all $c \in \calC$ and $k \in \calK_0$, so that the residual uncertainty about the presence of AI-generated text in these crawls, for instance from earlier generators, is retained in the model. Control pages are scored by the detectors like any other page. 
The prior on $\mu$ concerns the baseline cell at the first crawl only, while control cells differ in their flag rates, which could reflect lower specificity or higher prevalence. Since the effects in \eqref{eq:theta_model} are shrunk toward zero, the prevalence of a control cell stays close to that of the baseline cell, and the differences are attributed mainly to specificity. The second detector provides additional information to separate the two.

\paragraph{Specificity.} Specificity is identified from the control crawls, whose flags are expected to be predominantly false positives on pages of the study population itself. Since control flag rates also vary with the cell attributes---differently for the two detectors---we model the control-period specificity $\psi^{\Sp, 0}_{d, c}$ of detector $d$ in cell $c$ as
\begin{equation}\label{eq:control_sp}
    \logit (\psi^{\Sp, 0}_{d, c}) = \nu_{d, \ell[c]} + u^{\text{tier}}_{d, j[c]} + u^{\text{wc}}_{d, w[c]} + u^{\text{tld}}_{d, b[c]} \;,
\end{equation}
where $\nu_{d, \ell}$ is the logit control-period specificity of detector $d$ for the reference cell (mid tier, long, \texttt{.com}) of language $\ell$, and $u^{\text{tier}}_{d, j}$, $u^{\text{wc}}_{d, w}$ and $u^{\text{tld}}_{d, b}$ are the deviations from it associated with the tier, word-count level and TLD of the cell, respectively. For Binoculars, $\nu_{d, \ell}$ and the cell effects are defined only for english-only and bilingual pages. We assign the priors
\begin{equation}\label{eq:sp_prior}
\begin{split}
    \nu_{d, \ell} \sim \Normal (2.5, 2^2),\quad
    & u^{\text{tier}}_{d, j} \given \sigma_{d, \text{sp}, \text{tier}} \sim \Normal \left( 0, \sigma_{d, \text{sp}, \text{tier}}^2 \right), \ j \in \bar{\calJ} \;,\\
    u^{\text{wc}}_{d, w} \given \sigma_{d, \text{sp}, \text{wc}} \sim &\Normal \left( 0, \sigma_{d, \text{sp}, \text{wc}}^2 \right), \ w \in \bar{\calW}\;,\\
    u^{\text{tld}}_{d, b} \given \sigma_{d, \text{sp}, \text{tld}} \sim &\Normal \left( 0, \sigma_{d, \text{sp}, \text{tld}}^2 \right), \ b \in \bar{\calB} \;,
\end{split}
\end{equation} 
with $\sigma_{d, \text{sp}, \text{tier}}, \sigma_{d, \text{sp}, \text{wc}}, \sigma_{d, \text{sp}, \text{tld}} \sim \halfNormal(0, 1)$. The prior on $\nu_{d, \ell}$ is weakly informative, centered at a specificity of $\logit^{-1}(2.5) \approx 92\%$, with a dispersion that covers the empirical range of control flag rates across detectors and language levels. Note that here $\psi^{\Sp, 0}_{d, c}$ does not depend on the crawl, and its change after the control period is specified below. Appendix~\ref{sec:sensitivity_app} describes a reduced version of \eqref{eq:control_sp} in which fewer cell effects are included.

\paragraph{Sensitivity.} Sensitivity cannot be learned from the study population, which contains no labeled AI-generated text, and is therefore calibrated on benchmark corpora, which do not resolve the same language levels, centrality tier, word count or TLD (Section~\ref{sec:calib_datasets}). Since sensitivities measured on benchmark corpora need not transport to Common Crawl either, we therefore propagate the heterogeneity across corpora and allow the sensitivity to shift after the control period, which we elaborate below. For detector $d$ and language stratum $s \in \calS$, let $X_{d, s, q}$ be the number of texts flagged by the detector among the $N_{d, s, q}$ AI-involved texts of benchmark corpus $q$. The sensitivity $\pi_{d,s,q}$ of the detector on corpus $q$ varies across corpora around the control-period sensitivity $\psi^{\Se, 0}_{d, s} = \logit^{-1}(\eta_{d, s})$ of the detector in language stratum $s$, which is shared by every cell $c$ with $s[\ell[c]] = s$, so that
\begin{equation}\label{eq:se_prior}
\begin{split}
    X_{d, s, q} \given \pi_{d, s, q} &\sim \Binom \left( N_{d, s, q}, \, \pi_{d, s, q} \right) \;,\\
    \pi_{d, s, q} \given \eta_{d, s}, \phi_d &\sim \Beta \left( \phi_d \, \psi^{\Se, 0}_{d, s}, \, \phi_d \left( 1 - \psi^{\Se, 0}_{d, s} \right) \right), \quad \phi_d = \vartheta_d^{-2} \;, \\ 
    \eta_{d, s} &\sim \Normal (0, 1.5^2), \quad \vartheta_d \sim \halfNormal(0, 1) \;,
\end{split}
\end{equation}
where $\phi_d$ captures the heterogeneity across corpora, with smaller values indicating larger heterogeneity, which is thereby propagated into the model, instead of combining them into a single quantity. Integrating out $\pi_{d, s, q}$, the counts $X_{d, s, q}$ follow a beta-binomial distribution with mean $N_{d, s, q} \, \psi^{\Se, 0}_{d, s}$. When a stratum has a single benchmark corpus (Pangram, non-English), the corpus-level term is omitted, and $X_{d, s, q} \sim \Binom(N_{d, s, q}, \psi^{\Se, 0}_{d, s})$.

We do not constrain $\eta_{d, s}$ relative to the specificities. A common device to rule out the label-switching mode, in which the roles of flagged and unflagged are interchanged, is to require a positive Youden index \citep{youden_1950}, $\psi^{\Se} + \psi^{\Sp} - 1 > 0$, so that each detector performs better than chance, as is standard in latent class models without a gold standard \citep{hui_walter_1980, joseph_1995} and analogous to the identifiability constraints against label switching in mixture models \citep{stephens_2000, jasra_2005}. We instead check it on the fitted model. For each detector $d$ and cell $c$, we compute the Youden index of the control-period calibration, $J_{d, c} = \psi^{\Se, 0}_{d, s[\ell[c]]} + \psi^{\Sp, 0}_{d, c} - 1$, in every posterior draw, and report its distribution across the cells that occur in the data (Appendix~\ref{sec:model_diagnostics}). Pangram satisfies $J_{d, c} > 0$ in every cell and every draw (posterior means of $0.58$ to $0.97$ for English-stratum pages and $0.37$ for non-English pages). For Binoculars, which is applied to English-only and bilingual pages, the index is close to zero (posterior means between $-0.06$ and $0.04$), and it is positive in every draw for 68\% of the pages. Appendix~\ref{sec:youden_constraint} presents further discussion on a model that is constrained by $J_{d, c} > 0$.

\paragraph{Post-GenAI drift of detector performance.} The calibration above describes the performance of the detectors on the control crawls. Holding it constant over the later crawls may not be sufficient, because the detectors are applied unchanged to web text that is increasingly produced by newer generators and whose composition changes over time, while only the control pages inform specificity in domain. We therefore allow each operating characteristic to shift on the logit scale after the control period,
\begin{equation}\label{eq:drift}
\begin{split}
    \logit (\psi^{\Se}_{d, c, k}) &= \logit (\psi^{\Se, 0}_{d, s[\ell[c]]}) + \delta^{\Se}_{d, s[\ell[c]], k} \;,\\
    \logit (\psi^{\Sp}_{d, c, k}) &= \logit (\psi^{\Sp, 0}_{d, c}) + \delta^{\Sp}_{d, \ell[c], k} \;,
\end{split}
\end{equation}
where $\delta^{\Se}_{d, s, k} = \delta^{\Sp}_{d, \ell, k} = 0$ for all $k \in \calK_0$. Note that the operating characteristics appearing in \eqref{eq:data-model} are the drifted $\psi^{\Se}_{d,c,k}$ and $\psi^{\Sp}_{d,c,k}$. The drift is taken to be common to all cells within a language stratum for sensitivity, or a language level for specificity, as cell-level drift cannot be distinguished from cell-level prevalence. The drift series $\{\delta^{\Se}_{d, s, k} : k \in \calK\}$ and $\{\delta^{\Sp}_{d, \ell, k}: k \in \calK\}$ are assumed to be independent a priori, for each detector $d$ and each stratum $s$ or level $\ell$. We assume the shift to be static; it is constant after the last control crawl, i.e., $\delta^{\Se}_{d, s, k} = \delta^{\Se}_{d, s}$ and $\delta^{\Sp}_{d, \ell, k} = \delta^{\Sp}_{d, \ell}$ for all $k \geq K_0 + 1$, and for each detector $d \in \calD$,
\begin{equation}\label{eq:drift_static}
    \delta^{\Se}_{d, s} \sim \Normal(0, s_{\delta, \Se}^2), \forall s \in \calS \;, \qquad \delta^{\Sp}_{d, \ell} \sim \Normal(0, s_{\delta, \Sp}^2), \forall \ell \in \calL \;.
\end{equation}
This implies that we allow the operating characteristics to change once, at the onset of the generative-AI era, and remain at the shifted value. The scales $s_{\delta, \Sp}$ and $s_{\delta, \Se}$ are transport sensitivity parameters, with $s_{\delta, \Sp} = s_{\delta, \Se} = 0$ corresponding to exact transfer of the control calibration. We choose $s_{\delta, \Sp} = 0.5$ and $s_{\delta, \Se} = 0.25$ as the primary values. A shift in sensitivity cannot be separated from a change in prevalence after the control crawls except through its prior, so the scale for sensitivity is smaller, which holds the sensitivity on webpages nearer to its benchmark value. Appendix~\ref{sec:sensitivity_app} describes an alternative in which the shift evolves as a random walk, and varies different specification choices of the model hyperparameters.

\subsubsection{Posterior inference}

Let $\Theta$ collect the parameters of the prevalence model, which includes $\mu$, the random intercepts, slopes and spline coefficients along with their scale parameters. Let $\Psi^{\Sp} = (\nu, u, \sigma_{\text{sp}}, \delta^{\Sp})$ denote the parameters of the specificity model, including its control-period mean specificity for the reference cell, random intercepts and their scale parameters together with the drift, $\Psi^{\Se} = (\eta_{d, s}, \vartheta_d, \delta^{\Se})$ those of the sensitivity model, and let $\calY$ denote the detector outputs of the webpages and $\calX$ the counts of AI-involved benchmark texts that detectors flag. Marginalizing the latent status $Z$ yields the joint posterior
\begin{equation}\label{eq:joint_posterior}
\begin{split}
    p(\Theta, \Psi^{\Sp}, \Psi^{\Se} \given \calY, \calX) \propto \ &\module{modlik}{p(\calY \given \Theta, \Psi^{\Sp}, \Psi^{\Se})}{marginal web-page likelihood \eqref{eq:marginal}} \times \module{prior}{p(\Theta)}{prevalence\\ prior \eqref{eq:theta_model}} \times \module{prior}{p(\Psi^{\Sp})}{specificity prior\\ \eqref{eq:sp_prior}, \eqref{eq:drift}} \\
    &\quad \ \times \ \module{modlik}{p(\calX \given \Psi^{\Se})}{FNR benchmark likelihood, \eqref{eq:se_prior}} \ \times \ \module{prior}{p(\Psi^{\Se})}{sensitivity prior\\ \eqref{eq:se_prior}, \eqref{eq:drift}} \;,
\end{split}
\end{equation}
where the web-page likelihood is the product of \eqref{eq:marginal} over pages, with elements of $\Theta$ modeled by \eqref{eq:theta_model}, and $\psi$ of $(\Psi^{\Sp}, \Psi^{\Se})$ through \eqref{eq:drift}, and $p(\mathcal{X} \given \Psi^{\Se})$ is the benchmark likelihood of \eqref{eq:se_prior}. Specificity is informed by the control crawls through the web-page likelihood, whereas the machine-text benchmark counts, which belong to the sensitivity model, inform sensitivity only. Marginalizing out the latent status $Z$ is crucial, as it drastically reduces the dimension of the parameter space that an iterative sampling algorithm has to explore, since $Z$ has one entry per \Result{sample.n_pages} unique page. Integrating out $Z$ also yields the joint posterior density over continuous parameters only, which makes it possible to use gradient-based Hamiltonian Monte Carlo to sample from the full joint posterior. Pages in the same cell and crawl with the same detector outputs contribute identical terms to \eqref{eq:marginal}, so the page-level likelihood can be evaluated efficiently once per unique combination of cell, crawl, and detector output, which reduces the loop from around $5$ million pages, carrying $9$ million detector outputs, to about $50$ thousand unique combinations, during each step of the sampler.

\paragraph{Poststratification.} The model is first fitted to the sampled pages yielding posterior samples $\{ \theta^{(m)}_{c,k} : m = 1, \ldots, M \}$ which estimates the posterior distribution of the prevalence of cell $c$ in crawl $k$, for each $c$ and $k$. Because the sample does not allocate pages to cells in proportion to the web, averaging these draws over the sampled pages would give a biased estimate of the prevalence among the pages of the crawl. We therefore use poststratification, a standard technique in survey statistics \citep{little_1993, park_2004, kennedy_2020}. It reweights the cell-level posterior distributions of prevalence with respect to the share of each cell in the population we want to describe. Poststratification rests on two assumptions. First, the composition of the target population is known, that is, the number of eligible pages $N_{c,k}$ in every cell $c$ of crawl $k$ is available (exact for the Top, Upper-middle and Middle centrality tiers and estimated from a counting sample for the rest). Second, within a cell, the sampled pages are representative of the cell's unsampled pages, so that prevalence estimated from the former applies to the latter. For each crawl $k \in \calK$, let $\calC_k \subseteq \calC$ denote the cells with sampled pages in crawl $k$. Given $M$ posterior samples $\{\theta^{(m)}_{c,k} : m = 1, \ldots, M\}$, the poststratified posterior prevalence of crawl $k$ follows from samples
\begin{equation}\label{eq:poststrat}
    \bar{\theta}^{(m)}_k = \frac{\sum_{c \in \calC_k} w_{c,k} \, \theta^{(m)}_{c,k}}{\sum_{c \in \calC_k} w_{c,k}}, \quad m = 1, \ldots, M \;,
\end{equation}
where $w_{c,k}$ is the weight of cell $c$ in the target population in crawl $k$, and the choice of target determines the question answered. In particular, we consider two weighting schemes. The first is the pages of the crawl, in which $w_{c,k} = N_{c,k}$, where $N_{c,k}$ is the total number of pages in cell $c$ in crawl $k$ and $\bar{\theta}^{(m)}_k$ denotes posterior samples corresponding to the population-level prevalence of pages that contain \Result{outcome.definition}. The second acts as a proxy for the exposure of a random surfer, in which $w_{c,k}$ corresponds to the PageRank mass on the domains those pages in cell $c$ in crawl $k$ belong to, so that $\bar{\theta}^{(m)}_k$ corresponds to a posterior draw of the share of exposure that falls on  fully AI-generated or AI-edited pages. In this case, the prevalence is weighted by the link graph, giving more weight to central pages. 
The prevalence of a group of cells $\mathcal{G} \subseteq \calC$, such as the cells of centrality tier $j$, is obtained by restricting both sums in \eqref{eq:poststrat} to $\calC_k \cap \mathcal{G}$; we write $\bar{\theta}_{j,k}$ for tier $j$, and $w_{j,k}$ for the share of the weight of crawl $k$ that falls on tier $j$.

\paragraph{Contributions to the increase across centrality tiers.} For tier $j$ and crawl $k$, let $w_{jk}$ be the share of modeled eligible pages and $p_{jk}=\bar\theta_{j,k}$ their calibrated prevalence, so that the overall prevalence is $\bar\theta_k=\sum_j w_{jk}p_{jk}$. Let $k_0$ and $k_1$ be the crawls at the start and end of the comparison. We define a majority contribution as more than half of the net increase in overall prevalence. For Bottom and Tail together, $S_G=\sum_{j\in G}(w_{jk_1}p_{jk_1}-w_{jk_0}p_{jk_0})/(\bar\theta_{k_1}-\bar\theta_{k_0})$. Each tier's share is calculated in every posterior draw. The mean for Bottom and Tail together is the sum of their two means. If $L_j$ is the marginal 2.5th percentile of a tier's growth share, $\Pr(S_G\ge L_{\mathrm{Bottom}}+L_{\mathrm{Tail}})\ge0.95$ by the union bound. This provides a conservative one-sided posterior bound without assuming independence; adding the two marginal upper endpoints would not produce a central 95\% interval.

We separate changes in prevalence from changes in page shares using the symmetric decomposition of \cite{kitagawa_1955}:
\begin{equation}\label{eq:centrality-growth-decomposition}
w_{jk_1}p_{jk_1}-w_{jk_0}p_{jk_0}=\underbrace{\frac{w_{jk_0}+w_{jk_1}}{2}(p_{jk_1}-p_{jk_0})}_{\text{within-tier change}}+\underbrace{\frac{p_{jk_0}+p_{jk_1}}{2}(w_{jk_1}-w_{jk_0})}_{\text{changing page shares}}.
\end{equation}
The model treats page weights as fixed, so applying this identity to endpoint posterior means gives the exact posterior means of the additive components. Component intervals use paired joint draws. For a direct comparison of within-tier prevalence growth, we average the tier-specific changes using each tier's mean endpoint page share, normalized within Bottom and Tail and within the remaining tiers. This comparison holds the six-tier page distribution fixed; it does not follow the same domains or pages over time or identify a causal effect of centrality. The final-page-share comparison in Figure~\ref{fig:centrality-prevalence}c describes concentration relative to the final crawl and is not used to establish faster prevalence growth within tiers.

\paragraph{Computation.} We implement the model in Stan \citep{stan_manual_2026} and use \texttt{CmdStan} via the \texttt{CmdStanPy} interface, which translates the model to C++ and compiles it into an executable, and sample with the No-U-Turn sampler \citep{hoffman_gelman_2014}. Each of $4$ chains uses $1,000$ warmup iterations to adapt the step size and a diagonal mass matrix, with a target acceptance rate of $0.95$ and a maximum tree depth of $12$, and then draws $2,000$ samples, which gives $8,000$ posterior draws per model. Each run used 4 cores of an AMD EPYC 9654 processor and 64 GB of memory, with one core per chain, and the model was compiled with GCC 11.2.0. We assess convergence with the rank-normalized split-$\widehat{R}$ \citep{vehtari_rhat_2021}, bulk and tail effective sample sizes, divergent transitions, the energy Bayesian fraction of missing information and tree-depth saturation \citep{betancourt_2017}. For the reported model, the largest $\widehat{R}$ is $1.005$, the smallest bulk and tail effective sample sizes are $880$ and $736$, and no transition diverged or reached the maximum tree depth; the same holds for all the other fits (Appendix~\ref{sec:model_diagnostics}). Model variants are compared using Watanabe-Akaike Information Criteria \citep[WAIC,][]{watanabe_2010}. The comparison uses the page-level WAIC computed from all posterior draws, $\text{WAIC} = -2(\widehat{\text{lppd}} - \hat{p}_{\text{WAIC}})$, for which lower values indicate a better fit (Appendix~\ref{sec:model_comparison_app}).

\section{Limitations}\label{sec:limitations}

Relying on Common Crawl brings three limitations to the estimand. First, it covers only pages a crawler can reach: content behind logins, inside apps, or on sites that block crawlers is outside it, which excludes many of the major social media platforms even if its posts are publicly visible. Second, it covers only longer text. Current AI detectors are not reliable on very short passages; the open Pangram model, for example, requires at least 50 words, and every benchmark we use to measure detector error consists of texts of at least 500 characters. We therefore restrict the estimand to pages with at least 500 characters of extractable text (about 80 English words), which excludes standalone short-form writing such as individual posts, comments, and reviews. Third, in cases where short texts are collected on a single page (such as a comment thread or a list of reviews), the page can pass this threshold, but it is scored as one document, so its result describes the page as a whole rather than any individual post. Our estimates therefore describe long-form web text, and say nothing about the prevalence of \Result{outcome.definition} on social platforms. Common Crawl's crawling policy also changed over the study period to combat link spam, resulting in considerable variation in the number of unique domain nodes present in a given crawl \citep{CC_webgraph_stats}. However, just because a node exists in the web graph does not mean it is crawled. Link farmed domains with zero out-degree are not crawled, and therefore would not affect our analyses. A fourth limitation concerns the model specification rather than the data. Pages are sampled within domains, and pages of the same domain are likely to resemble each other in whether they contain \Result{outcome.definition}---potentially due to shared editorial practices---beyond what is explained by the language, centrality tier, TLD, and length of the page. We do not include domain-level random effects, which would have otherwise induced such correlation structure among pages of the same domain. They would add one parameter for each of the \Result{sample.n_domains} sampled domains to $\Theta$ in \eqref{eq:joint_posterior}, and would prevent the aggregation of pages into about $50$ thousand unique combinations on which the efficient evaluation of the likelihood relies, thus making estimation computationally prohibitive in our setting. The centrality tier and the TLD, which are domain-level attributes, absorb part of this heterogeneity, and we expect the estimates of prevalence by group to be little affected. In addition, domains contribute few pages: across the 71 crawls, the median domain has 6 sampled pages, 38\% of domains have at most four, and 14\% have a single page. Domain-specific random effects would then be only weakly informed by the data and hard to separate from the misclassification of the detectors. Domain-level random effects, perhaps with a lower-dimensional approximation, are a direction for future work.

\section*{Acknowledgements}

\noindent The authors gratefully acknowledge financial support from the Tech for Open Minds (TOM) initiative at SICSS-UW, which is supported by the Templeton World Charity Foundation.  THMc is partially supported by a María de Maeztu Visiting Professorship at El Centro de Estudios Monetarios y Financieros (CEMFI). Research reported in this publication was supported by the National Center for Advancing Translational Sciences of the National Institutes of Health under the Award Number TL1TR002368. The content is solely the responsibility of the authors and does not necessarily represent the official views of the National Institutes of Health. This work was also supported by the University of Washington eScience Institute's NSF funded CloudBank pilot projects (NSF 1925001 and 2505560) and an NSF ACCESS grant (CIS240554), and used computational and storage services of the Hyak Klone CPU clusters provided by the University of Washington and the eScience Institute. This work was also supported by Pangram's research credits. We are also grateful for the feedback and insights of the guest speakers at the 2026 Summer Institute in Computational Social Science at the University of Washington, as well as financial and in-kind support for the Summer Institute from the Center for Statistics \& the Social Sciences and the Department of Sociology at the University of Washington.

\section*{Author contributions}

\noindent A.V., S.P. and T.H.M. conceived and designed the study; A.V. led the sampling design and implementation, with M.N., R.S., I.T., Y.W. and L.W.; A.V. led the AI-detection evaluation and scoring, with M.G. and D.P.; S.P. led the statistical modeling, with P.I., L.Y. and J.S.; A.V., S.P. and T.H.M. wrote the paper; A.V. and T.H.M. acquired funding. All authors reviewed and approved the manuscript.

\section*{Conflict of interest}
\noindent While Pangram provided API credits, they did not direct any of our research decisions or review materials prior to release.

\section*{AI Use Statement}
We used AI coding agents to implement the sampling and extraction of WARC files from Common Crawl, to manage the Slurm jobs that ran AI detection on GPU, and to write and edit software for the modeling. We also used AI tools to design tables and figures, to obtain feedback on the project at many points, to identify relevant literature, and to edit the manuscript for readability. We have reviewed and validated all AI-assisted work and take full responsibility for the final content, including all text, claims, code, and figures produced with the aid of AI tooling. 

\section*{Ethics Statement}
None of the outputs from any AI-detection tool used here are intended to be used as conclusive evidence to establish authorship. All data used are publicly available and license permissions were followed at every stage. 

\section*{Reproducibility Statement} 
All web data come from Common Crawl (\url{https://commoncrawl.org}), using its crawl archives and index (\url{https://index.commoncrawl.org}) and its domain-level web graphs (\url{https://commoncrawl.org/web-graphs}), from which we draw the PageRank-stratified sample. Detector sensitivity was estimated on four labeled corpora, all publicly available: EditLens \citep{thai2026editlens}, data and code at \url{https://github.com/pangramlabs/EditLens}); RAID \citep{dugan2024} (\url{https://github.com/liamdugan/raid}); the Multilingual GenAI Content Detection corpus \citep{wang2025genai} (\url{https://github.com/mbzuai-nlp/COLING-2025-Workshop-on-MGT-Detection-Task1}); and AITDNA \citep{dycke2026youraitextmine} (\url{https://github.com/UKPLab/arxiv2026-aitdna}). The four open detectors we evaluated are publicly available. The two open Pangram models are the EditLens classifiers released by Pangram Labs: a PEFT adapter on Llama-3.2-3B (\url{https://huggingface.co/pangram/editlens_Llama-3.2-3B}; base model \url{https://huggingface.co/meta-llama/Llama-3.2-3B}), which we deployed on the full sample, and a RoBERTa-large classifier (\url{https://huggingface.co/pangram/editlens_roberta-large}). Binoculars is run from its reference implementation (\url{https://github.com/ahans30/Binoculars}) with two model pairs: Falcon-7B and Falcon-7B-Instruct (\url{https://huggingface.co/tiiuae/falcon-7b}, \url{https://huggingface.co/tiiuae/falcon-7b-instruct}), which we deployed on English-containing pages, and Qwen2.5-7B and Qwen2.5-7B-Instruct (\url{https://huggingface.co/Qwen/Qwen2.5-7B}, \url{https://huggingface.co/Qwen/Qwen2.5-7B-Instruct}). The commercial detector, Pangram~4, used as an independent check on the pre-2022 control crawls, was accessed through the Pangram API (\url{https://www.pangram.com/solutions/api}) in \textsc{October, 2026}. Upon publication we will release the code used to sample from Common Crawl and extract page text, the sampled page texts with their detector scores and labels, and the implementation of the Bayesian hierarchical model. 

\newpage
\bibliography{cite}

\newpage
\appendix

\section{Related work}
\label{sec:related}

Several related studies inform our approach, and together they identify some components required to produce a valid estimate of the prevalence of \Result{outcome.definition} on the open web. The first is a defined target population frame. \cite{graphite2026} draw a random sample of pages from Common Crawl, retain the English-language pages, and report that nearly half of these are primarily AI-generated; \cite{dolezal2026} find that about 35\% of new English-language pages archived by the Internet Archive were AI-generated or AI-assisted by mid-2025; and \cite{pew2026} draw a simple random sample of English-language pages from Common Crawl snapshots, so that flag rates in the sample generalize to that crawl. The second is an account of detector error. Studying the text in recent arXiv submissions, \cite{elazar2026} apply a Rogan--Gladen correction to Pangram labels, recovering prevalence from detector output with measured sensitivity and specificity; studying recent ICLR peer reviews and Wikipedia articles, \cite{latona2024ai} and \cite{brooks2024} subtract an estimated false-positive rate from flag rates to report a lower bound; and \cite{liang2024, liang2025} sidestep per-document labels altogether, inferring the share of LLM-modified text in scientific abstracts from population-level shifts in word frequencies. Our study combines both components, for English and non-English pages, and jointly models sampling and detector error into a valid statistical estimate with uncertainty.  

We also make a conceptual contribution: it is important to treat a detector score not as a label but as a noisy measurement of a latent construct, i.e., whether a page is fully AI-generated or AI-edited, that cannot be observed directly. Once AI-detection is framed this way, the components above can be seen as constituent parts of a single measurement model: the probability sample defines the population over which the latent quantity is to be estimated, labeled benchmarks identify the measurement error of each detector, and a statistical model carries both into one estimate with uncertainty. We evaluate six detectors on four benchmark collections and on 30 crawls that predate capable language models, with sensitivity and specificity estimated separately for English and non-English text and by text length. A Bayesian hierarchical model then estimates detector error jointly with prevalence and poststratifies to the population, so that each reported interval reflects both sampling and detection error. Because the sample is designed, the same model yields error-corrected estimates for subpopulations defined by network position, domain type and language, and lets us decompose change over time into shifts within these groups and shifts in their relative weight, so that we can speak to the composition of the trend and not only its headline level. We summarize our contributions relative to other relevant work in  (Table~\ref{tab:comparison}). 

\begin{table}[t]
\centering
\footnotesize
\begin{threeparttable}
\caption{Comparison of existing studies and this study}
\label{tab:comparison}
\setlength{\tabcolsep}{3pt}
\renewcommand{\arraystretch}{1.2}
\begin{tabularx}{\textwidth}{@{} >{\raggedright\arraybackslash}p{2.3cm}
                                 >{\raggedright\arraybackslash}X
                                 >{\raggedright\arraybackslash}X
                                 >{\raggedright\arraybackslash}X
                                 c c
                                 >{\raggedright\arraybackslash}p{2.1cm} @{}}
\toprule
 & & & & \multicolumn{2}{c}{Corrects for} & \\
\cmidrule(lr){5-6}
Study & Corpus & Sampling & Detection & FP & FN & Uncertainty \\
\midrule
\cite{graphite2026} & Common Crawl, English articles & Non-probability & Mean detector score & \checkmark & --- & None reported \\
\cite{dolezal2026} & Internet Archive, English captures & One URL per host & Pangram v3 & --- & --- & None reported \\
\cite{pew2026} & Common Crawl, English pages & Simple random, 10k pages per crawl & Open Pangram; Pangram 3.3 check & --- & --- & None reported \\
\cite{liang2024, liang2025}   & Scientific corpora & Near-census & Distributional estimator & \checkmark\tnote{a} & \checkmark\tnote{a} & Validation-based \\
\cite{elazar2026}  & arXiv & Near-census & Distributional + Pangram & \checkmark & \checkmark\tnote{b} & 95\% CI \\
\cite{brooks2024} & New English Wikipedia articles & Near-census & GPTZero + Binoculars & (\checkmark)\tnote{c} & --- & Lower bound \\
\cite{latona2024ai} & ICLR peer reviews & Near-census & GPTZero & \checkmark\tnote{d} & --- & Bootstrap; lower bound \\
\midrule
\textbf{This study} & Common Crawl, 71 crawls, 2019--2026 & Two-stage stratified probability sample & Two open-weight detectors & \checkmark\tnote{e} & \checkmark\tnote{e} & Full posterior\tnote{f} \\
\bottomrule
\end{tabularx}
\begin{tablenotes}[flushleft]
\footnotesize
\item FP = false positives; FN = false negatives; --- = not corrected.
\item[a] Error is handled inside the distributional estimator rather than by a separate correction.
\item[b] Rogan--Gladen adjustment using estimated sensitivity and specificity.
\item[c] Detection thresholds were set to a 1\% false-positive rate on pre-GPT-3.5 articles; the rate is controlled, not subtracted.
\item[d] A point estimate of the false-positive rate is subtracted.
\item[e] Specificity is estimated from 30 crawls predating capable LLMs and sensitivity from labeled benchmarks, both by language group, and estimated jointly with prevalence.
\item[f] Propagates both sampling uncertainty and uncertainty in detector error rates.
\end{tablenotes}
\end{threeparttable}
\end{table}

\section{AI Detector Benchmarks}
\label{appendix:detector_benchmarks}

This appendix summarizes the benchmark evaluations used to calibrate detector sensitivity (true-positive rates) and specificity (true-negative rates) for the measurement model.

\begin{figure}[p]
  \centering
  \includegraphics[width=\linewidth]{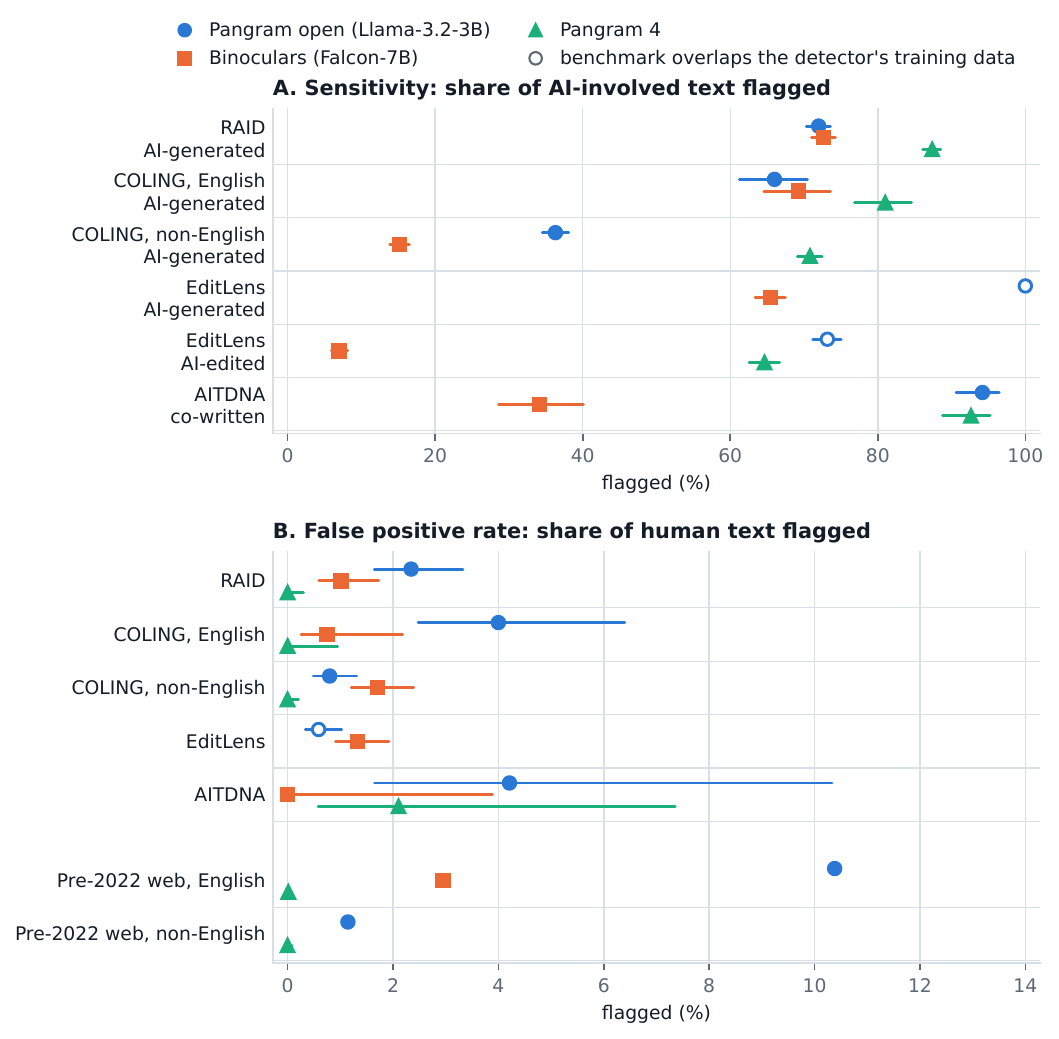}
  \caption{Detector performance at deployed thresholds. (A) Share of AI-involved text flagged on each labeled corpus, with 95\% Wilson intervals. (B) Share of human text flagged on the same corpora and on pre-2022 Common Crawl pages, which are the study population. Hollow markers indicate a corpus that overlaps the detector's training data; those rows are excluded from calibration. Pangram 4 was run on the AI-edited EditLens texts only, and on a random subsample of 20,100 control pages. Binoculars is shown on non-English benchmark text for comparison but is not run on non-English pages.}
  \label{fig:benchmarks}
\end{figure}

Figure~\ref{fig:benchmarks} reports performance at the deployed thresholds. Panel (A) shows the share of labeled AI-involved texts that each detector flags (sensitivity) on each benchmark corpus; panel (B) shows the share of labeled human texts flagged (1 $-$ specificity), including pre-2022 Common Crawl pages, which serve as an in-domain negative control. Hollow markers indicate benchmark rows that overlap a detector's training data and are excluded from calibration.

\begin{figure}[p]
  \centering
  \includegraphics[width=\linewidth]{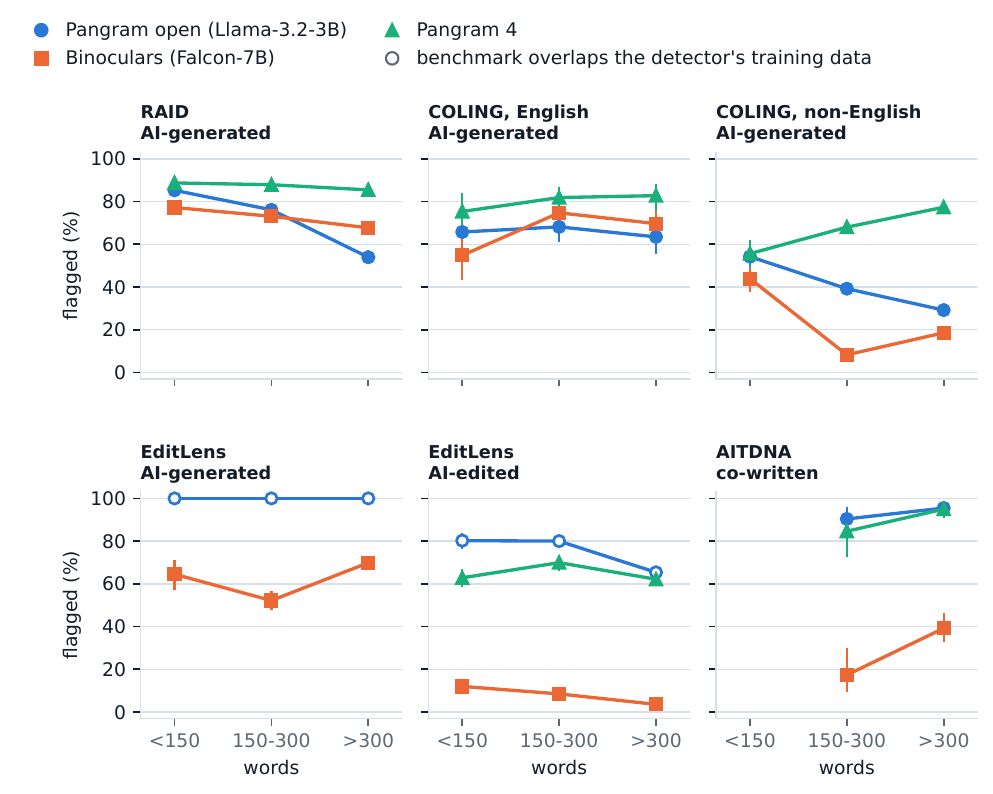}
  \caption{Sensitivity by text length at deployed thresholds. Each panel is one labeled corpus and positive class; the horizontal axis is text length in words, in the three bins used for calibration (under 150, 150 to 300, over 300). Vertical bars are 95\% Wilson intervals. Bins with fewer than 30 AI texts are omitted, which removes the shortest bin for AITDNA. Hollow markers indicate a corpus that overlaps the detector's training data; those rows are excluded from calibration. Pangram 4 was run on the AI-edited EditLens texts only.}
  \label{fig:bench-length}
\end{figure}

\begin{figure}[p]
  \centering
  \includegraphics[width=\linewidth]{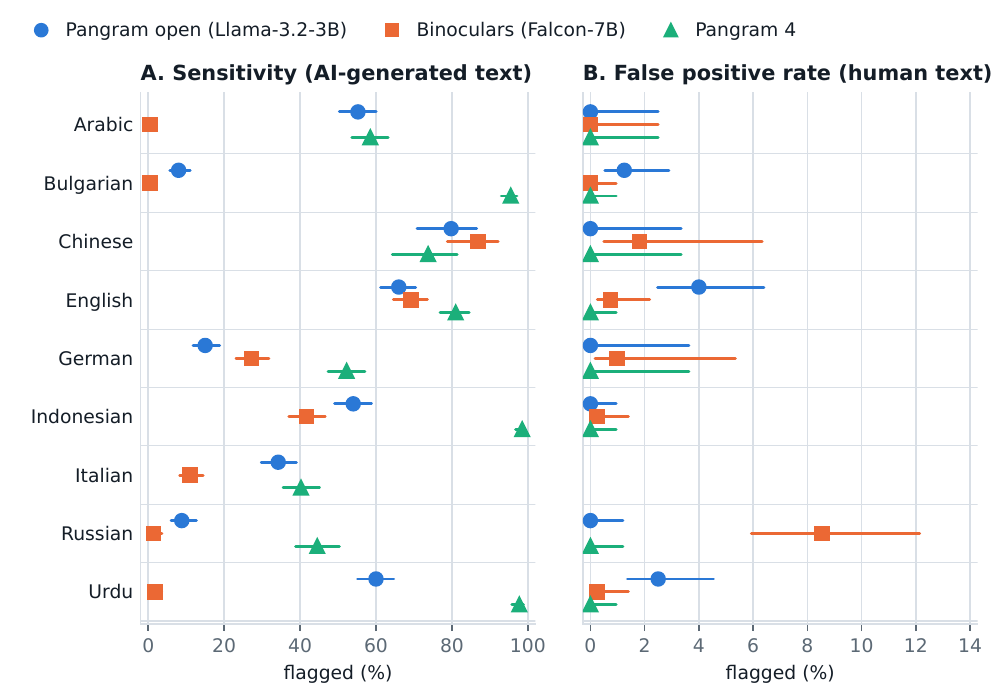}
  \caption{The COLING 2025 multilingual set by language, at deployed thresholds. (A) Share of AI-generated text flagged. (B) Share of human text flagged. Horizontal bars are 95\% Wilson intervals. English texts are scored at each detector's English threshold and all other languages at its non-English threshold. The set contains no human-written Italian text, so that row is empty in panel B. Binoculars is shown for comparison but is not run on non-English pages.}
  \label{fig:bench-language}
\end{figure}

\begin{figure}[p]
  \centering
  \includegraphics[width=0.75\linewidth]{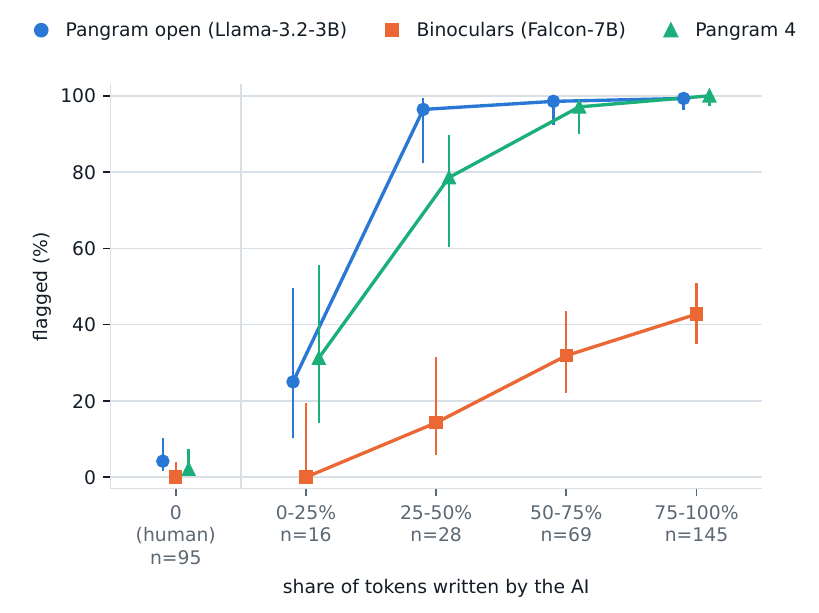}
  \caption{Detector flag rates by share of AI-generated tokens. Documents from the AITDNA benchmark are binned by the proportion of tokens written by an AI model, with human-written documents (share $=0$) at left; points give the percentage of documents in each bin that a detector flagged as fully AI-generated or AI-edited, bars give 95\% intervals, and $n$ is the number of documents per bin. Flag rates rise with the AI-generated share and the detectors diverge in where they err: the perplexity-based Binoculars flags fewer than half of the documents that are more than three-quarters fully AI-generated, while the supervised Pangram models flag a quarter or more of those with under a quarter AI-generated. Each detector measures a different operationalization of AI involvement, so none maps cleanly onto authorship.}
  \label{fig:aitdna_token_share}
\end{figure}

\subsection{Empirical overlap between AI-detectors}
Table~\ref{tab:fpr_by_context} demonstrates that neither Open Pangram nor Binoculars-Falcon achieves universal best performance across contexts. On English pages from crawls that predate capable language models, where every flag is a false positive, the open Pangram model wrongly flags fewer pages than Binoculars on \texttt{.gov} and \texttt{.edu} domains ($3.4\%$ against $4.7\%$, and $2.2\%$ against $4.9\%$) but more on \texttt{.com} domains ($5.2\%$ against $3.0\%$), and fewer short pages but more long ones. Sensitivity reverses in a similar way: the two detectors catch fully AI-generated benchmark text at similar rates ($0.720$ and $0.727$ on RAID), but the open Pangram model catches $94\%$ of text co-written with a language model, against $34\%$ for Binoculars.

The model of Section~\ref{sec:modeling} treats the outputs of the two detectors as conditionally independent given the true status of a page. Table~\ref{tab:detector_dependence} examines this assumption by comparing, within a true class, the number of texts that both detectors flag with the number expected if their errors were independent, and by reporting the correlation $\phi$ of the two binary flags.
\begin{table}[t]
\centering
\footnotesize
\begin{threeparttable}
\caption{Dependence between detectors' errors, within true class}
\label{tab:detector_dependence}
\setlength{\tabcolsep}{4pt}
\begin{tabular}{@{} l l r r r r r @{}}
\toprule
 & & & \multicolumn{2}{c}{Flagged by both} & & \\
\cmidrule(lr){4-5}
Data & Pair & $n$ & Observed & Expected\tnote{a} & $\phi$ & $\phi$ within length\tnote{b} \\
\midrule
\multicolumn{7}{@{}l}{\textit{Human text: pre-2022 control crawls (every flag is a false positive)}} \\
English, all pages             & LL--FA & 1{,}769{,}938 & 5{,}278 & 5{,}422 & $-0.002$ & --- \\
English, Pangram~4 subsample   & LL--P4 & 15{,}376      & 0       & 0.2     & $-0.004$ & --- \\
Non-English, Pangram~4 subsample & LL--P4 & 4{,}724     & 0       & 0.0     & ---      & --- \\
\addlinespace
\multicolumn{7}{@{}l}{\textit{AI-involved text: benchmarks}} \\
RAID, English                  & LL--P4 & 2{,}891 & 2{,}004 & 1{,}818 & 0.43 & 0.45 \\
RAID, English                  & LL--FA & 2{,}891 & 1{,}670 & 1{,}512 & 0.27 & 0.26 \\
RAID, English                  & FA--P4 & 2{,}891 & 1{,}937 & 1{,}836 & 0.24 & 0.23 \\
Multilingual, English          & LL--P4 & 400     & 242     & 214     & 0.38 & 0.38 \\
Multilingual, English          & LL--FA & 400     & 225     & 183     & 0.48 & 0.48 \\
Multilingual, 8 other languages & LL--P4 & 2{,}782 & 930    & 715     & 0.35 & 0.28\tnote{c} \\
AITDNA, co-written             & LL--P4 & 258     & 234     & 225     & 0.56 & --- \\
AITDNA, co-written             & LL--FA & 258     & 88      & 83      & 0.18 & --- \\
\bottomrule
\end{tabular}
\begin{tablenotes}[flushleft]
\footnotesize
\item LL = open Pangram (Llama-3.2-3B); FA = Binoculars-Falcon; P4 = Pangram~4. Flags are at each
detector's deployed threshold; Pangram~4 flags a text it calls AI or mixed. $\phi$ is the
correlation of the two binary flags; --- where a detector never flags or the set is too small to
split.
\item[a] Pages both detectors would flag if their errors were independent within the true class.
\item[b] Size-weighted mean of $\phi$ across three length bands (500--900, 900--1,800 and
$\geq$1,800 characters), removing correlation that arises only because both detectors find shorter
or longer text easier.
\item[c] Also within language; groups of fewer than 60 texts are excluded (2,512 of 2,782 texts).
\end{tablenotes}
\end{threeparttable}
\end{table}

\begin{table}[t]
\centering
\small
\begin{threeparttable}
\caption{False-positive rates on pre-2022 English pages, by context}
\label{tab:fpr_by_context}
\begin{tabular}{@{} l r r r @{}}
\toprule
 & & \multicolumn{2}{c}{Share of pages flagged} \\
\cmidrule(l){3-4}
Context & Pages & Open Pangram & Binoculars-Falcon \\
\midrule
All English-only pages & 1{,}336{,}683 & 4.5\% & \textbf{3.0\%} \\
\addlinespace
\multicolumn{4}{@{}l}{\textit{Top-level domain}} \\
\texttt{.com}   & 829{,}996 & 5.2\% & \textbf{3.0\%} \\
\texttt{.org}   & 198{,}951 & 4.1\% & \textbf{2.7\%} \\
\texttt{.net}   &  38{,}121 & \textbf{3.8\%} & 3.9\% \\
\texttt{.gov}   &  19{,}931 & \textbf{3.4\%} & 4.7\% \\
\texttt{.edu}   &  15{,}849 & \textbf{2.2\%} & 4.9\% \\
Other           & 233{,}835 & 3.0\% & 3.0\% \\
\addlinespace
\multicolumn{4}{@{}l}{\textit{Text length (characters)}} \\
500--900        & 270{,}147 & \textbf{3.7\%} & 6.9\% \\
900--1,800      & 338{,}865 & \textbf{2.2\%} & 3.4\% \\
$\geq$1,800     & 727{,}671 & 5.9\% & \textbf{1.3\%} \\
\addlinespace
\multicolumn{4}{@{}l}{\textit{Centrality tier}} \\
Top             & 247{,}784 & 7.9\% & \textbf{3.8\%} \\
Upper-middle    & 275{,}965 & 4.8\% & \textbf{2.9\%} \\
Middle          & 303{,}686 & 4.2\% & \textbf{2.7\%} \\
Lower-middle    & 256{,}880 & 3.3\% & \textbf{2.6\%} \\
Tail            & 180{,}196 & \textbf{2.7\%} & 2.9\% \\
Bottom          &  72{,}172 & \textbf{2.5\%} & 3.6\% \\
\bottomrule
\end{tabular}
\begin{tablenotes}[flushleft]
\footnotesize
\item Eligible English-only pages from the 30 control crawls (\texttt{CC-MAIN-2019-04} to
\texttt{CC-MAIN-2021-49}), which predate capable language models, so every flag is a false
positive. Each detector is at its deployed threshold. The lower rate in each row is in bold.
Shares are of sampled pages and are not design-weighted. Bilingual pages are excluded: the open
Pangram model flags $28.5\%$ of them against $2.8\%$ for Binoculars, which would dominate every row.
\end{tablenotes}
\end{threeparttable}
\end{table}

The first block uses human text, namely the pre-2022 control crawls, where every flag is a false positive. Among the $1.77$ million English control pages scored by both open Pangram and Binoculars-Falcon, $5{,}278$ are flagged by both against $5{,}422$ expected, and $\phi = -0.002$, so the false positives of the two detectors are essentially independent. The subsamples scored by Pangram~4 show no joint false positives, in line with the expected $0.2$ or fewer. The second block uses the benchmarks of AI-involved text. There the flags are positively dependent, with $\phi$ between $0.18$ and $0.56$ and more joint flags than independence implies. For the pair used in the model, open Pangram and Binoculars-Falcon, $\phi$ is $0.27$ on RAID, $0.48$ on the English multilingual set and $0.18$ on the co-written AITDNA texts. The correlations do not come from text length, since they remain similar within three length bands. Texts that one detector finds easy to recognize as AI-generated tend to be easy for the other as well.

The assumption therefore holds on the human side, which is the side that determines specificity and is identified from the control crawls. It does not hold on the AI side, where the dependence is positive. Since the sensitivities enter the likelihood as a product of the marginal sensitivities of each detector, the model assigns too small a probability to pages flagged by both detectors and too large a probability to pages flagged by exactly one, among pages that are AI-generated. We expect the effect on the prevalence estimate to be limited, because Pangram scores every page and carries most of the information, while Binoculars is applied to english-stratum pages only, but we have not quantified it. A model with a dependence parameter on the AI side is a direction for future work.

\section{Additional Results}

\begin{figure}[!htbp]
\centering
\includegraphics[width=0.9\linewidth]{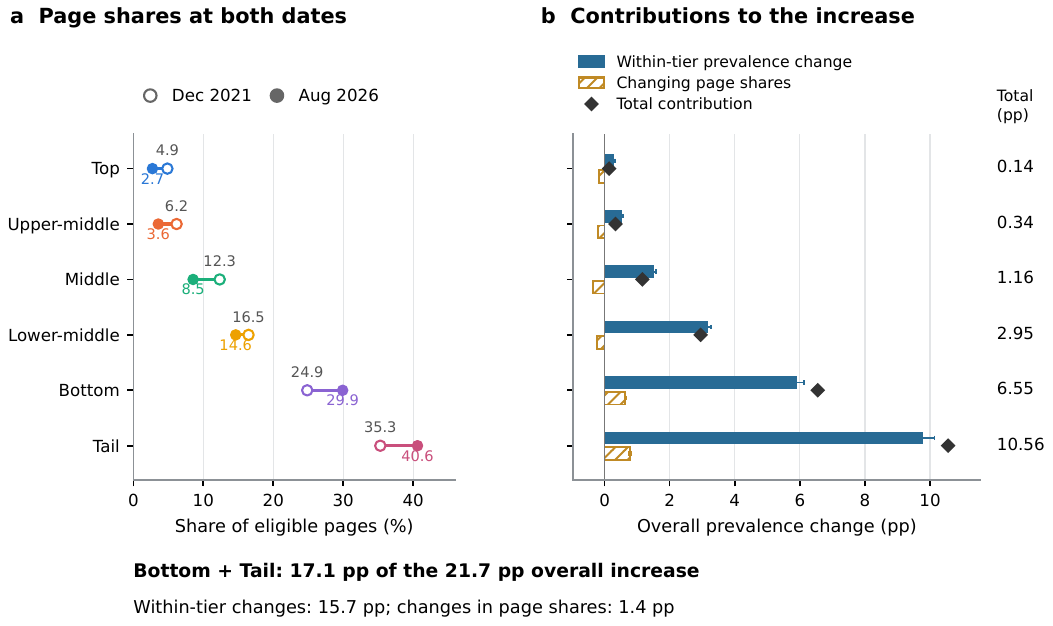}
\caption{\small \textbf{Separating estimated growth in prevalence of pages in Common Crawl with \Result{outcome.definition} from changes in the distribution of pages.} \textbf{(a)} Each tier's share of modeled eligible pages at the start and end of the December 2021 to August 2026 comparison. \textbf{(b)} Contribution to the overall prevalence increase, in percentage points. Blue bars show $\bar w_j(p_{jk_1}-p_{jk_0})$, the component from prevalence change within tiers. Hatched bars show $\bar p_j(w_{jk_1}-w_{jk_0})$, the component from changing page shares. Bars denote separate components measured from zero; diamonds show their sum $w_{jk_1}p_{jk_1}-w_{jk_0}p_{jk_0}$. Bars and diamonds show posterior means. Error bars show 95\% credible intervals from paired posterior draws. Overbars denote averages of the two endpoint values. The Bottom and Tail totals sum the two least-central tiers. Within-tier change includes changes in the pages and domains represented within each tier.}
\label{fig:centrality-growth-decomposition}
\end{figure}

\begin{figure}[!htbp]
  \centering
  \includegraphics[width=0.75\linewidth]{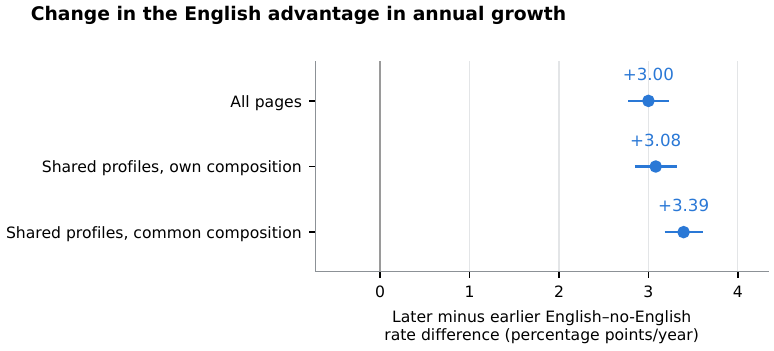}
  \caption{\textbf{Change in the English advantage in annual growth under alternative compositions.} Later-minus-earlier change in the English-only minus no-English difference in average annual increase, comparing February 2, 2023 to January 19, 2025 with January 19, 2025 to August 13, 2026, in percentage points per year; positive values indicate a larger English advantage in the later period. A profile is a combination of centrality tier, TLD and length. \emph{All pages}: all represented pages. \emph{Shared profiles, own composition}: the profiles observed for every language group at every endpoint, with each group keeping its own composition on them. \emph{Shared profiles, common composition}: the same profiles, with every group given the same composition (the average over the six most recent crawls). Points are posterior means and bars are 2.5th to 97.5th percentiles of matched posterior-draw contrasts.}
  \label{fig:language-growth-gap}
\end{figure}

\subsection{Prevalence pattern by text length}\label{sec:length_app}

\begin{figure}[t]
\centering 
\includegraphics[width=0.9\linewidth]{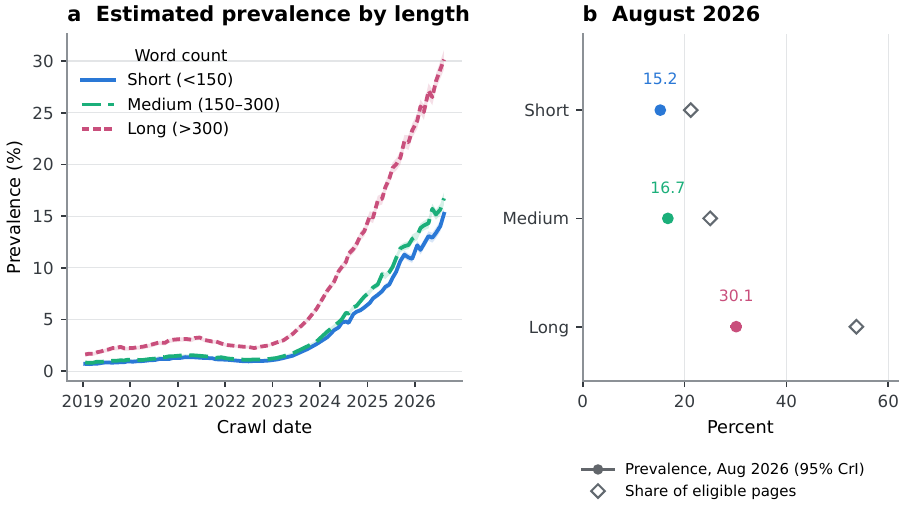} 
\caption{Prevalence of pages with \Result{outcome.definition} by word-count level. (a) Posterior mean and 95\% credible interval over the crawls. (b) Posterior mean and 95\% credible interval in August 2026 (circles), and the share of eligible pages in each word-count level (diamonds).}
\label{fig:length_prev}
\end{figure}

Pages are classified into three word-count levels (short, medium, long), which are part of the cell definition of the model. In the August 2026 snapshot, 30.2\% of long pages (29.2--31.1\%) were estimated to contain \Result{outcome.definition}, about twice that of short pages (15.2\%, 14.7--15.8\%) and medium pages (16.7\%, 16.1--17.3\%). Long pages hold 53.8\% of eligible pages, medium pages 25.0\% and short pages 21.2\%. After assigning each length level the same distribution of centrality, TLD and language, long pages were at 29.4\%, short pages at 15.7\% and medium pages at 17.4\%, so the difference is not explained by the composition of the groups. The sensitivity of the detectors to text length in the benchmark corpora is shown in Figure~\ref{fig:bench-length}. The model adjusts for text length through specificity, which varies with word count, but treats detector sensitivity as constant across the word-count levels, since the latter is available only by detector and language stratum. The comparison by length therefore relies on the assumption of constant sensitivity, unlike the comparisons by language, which have their own sensitivity. Figure~\ref{fig:length_prev} shows the estimates over time. Long pages have the highest prevalence at every crawl, and the difference from short and medium pages widens after 2023.

\subsection{Additional results by language}\label{sec:lang_detail_app}

\paragraph{Growth.} The prevalence of pages with \Result{outcome.definition} grew faster among English-only pages than among non-English-only pages, by \Result{language.rate.gap.early.mean} (\Result{language.rate.gap.early.lo95}--\Result{language.rate.gap.early.hi95}) percentage points per year in the first interval and by \Result{language.rate.gap.late.mean} (\Result{language.rate.gap.late.lo95}--\Result{language.rate.gap.late.hi95}) in the second, corresponding to an increment in growth rate by \Result{language.rate.gapchange.all.mean} (\Result{language.rate.gapchange.all.lo95}--\Result{language.rate.gapchange.all.hi95}) percentage points per year. This increase reflects the acceleration of English-only growth (\Result{language.rate.english.change.mean} points per year, \Result{language.rate.english.change.lo95}--\Result{language.rate.english.change.hi95}) much more than that of non-English-only growth (\Result{language.rate.nonenglish.change.mean}, \Result{language.rate.nonenglish.change.lo95}--\Result{language.rate.nonenglish.change.hi95}). The rates are averages over intervals fixed by calendar date; we do not estimate a change point.

\paragraph{Bilingual and unknown-language pages.} Prevalence of bilingual pages with \Result{outcome.definition} remained far below that of English-only pages, by \Result{language.contrast.bilingual_vs_english.start2023.abs} points in February 2023, \Result{language.contrast.bilingual_vs_english.start2025.abs} in January 2025 and \Result{language.contrast.bilingual_vs_english.latest.abs} in August 2026, although as a fraction of English-only prevalence it rose from \Result{language.relative.bilingual_to_english.start2023.mean} to \Result{language.relative.bilingual_to_english.latest.mean}. Language tags describe the captured page and can include navigation text or footers, so the bilingual category does not establish bilingual authorship. Pages of unknown language, 1--2\% of the sample, are not shown in Figure~\ref{fig:language-prevalence}.

\paragraph{Common composition.} The \Result{language.shared.n_profiles} shared combinations retain \Result{language.cover.english.start2023}\%, \Result{language.cover.english.start2025}\% and \Result{language.cover.english.latest}\% of English-only pages at the three endpoints and more than \Result{language.cover.others.min}\% of the pages of the other two groups. The comparison with the increase of \Result{language.rate.gapchange.shared_own.mean} (\Result{language.rate.gapchange.shared_own.lo95}--\Result{language.rate.gapchange.shared_own.hi95}) percentage points per year, when each group keeps its own composition on the same combinations, separates the effect of dropping the unshared combinations from that of reweighting.

\paragraph{Interpretation.} The fitted language categories are coarse. For example, non-English-only pools every non-English language, including but not limited to French, German, Spanish, Chinese, Japanese and Russian, and the detector evaluations that calibrate them cover Arabic, Bulgarian, English, German, Indonesian, Italian, Russian, Urdu and Chinese, with no human negatives for Italian. Non-English sensitivity is calibrated on machine text from 2023-era generators in only some of the languages (Section~\ref{sec:calib_datasets}), so the non-English levels carry more calibration uncertainty than the English ones. The evaluations reveal heterogeneous detector performance across languages, so the non-English-only estimate does not describe any individual non-English language. It serves only as a reference against which we compare the English-only prevalence estimates.

\paragraph{Language Comparisons Adjusted for Composition.}
Table~\ref{tab:lang_comp} presents the prevalence of pages with \Result{outcome.definition} for each language level as estimated and after assigning every level the common distribution of centrality, TLD and length, one dimension at a time and all three together. 
\begin{table}[t]
\centering
\caption{Estimated prevalence of pages in Common Crawl with \Result{outcome.definition} (\%) in the August 2026 snapshot by language level, unstandardized and standardized to the common distribution of centrality tier, TLD category and length level, each separately and jointly. The estimated prevalence uses each language level's own composition of these attributes. Entries are posterior means with 95\% credible intervals.}
\label{tab:lang_comp}
\small
\resizebox{0.95\textwidth}{!}{%
\begin{tabular}{lccccc}
\toprule
\multirow{2}{*}{Language level ($\calL$)} & \multirow{2}{*}{\shortstack{Estimated\\prevalence}} & \multicolumn{4}{c}{Standardized by} \\
\cmidrule(l){3-6}
 & & Centrality & TLD & Length & All three \\
\midrule
\multirow{2}{*}{English only} & 32.8 & 33.7 & 32.9 & 32.0 & 32.5 \\
 & {\scriptsize (32.2--33.4)} & {\scriptsize (33.1--34.3)} & {\scriptsize (32.3--33.4)} & {\scriptsize (31.4--32.6)} & {\scriptsize (31.9--33.0)} \\[4pt]
\multirow{2}{*}{No English detected} & 17.0 & 16.5 & 17.3 & 17.0 & 16.8 \\
 & {\scriptsize (15.4--18.4)} & {\scriptsize (15.0--17.9)} & {\scriptsize (15.7--18.7)} & {\scriptsize (15.4--18.4)} & {\scriptsize (15.3--18.2)} \\[4pt]
\multirow{2}{*}{English + another language} & 20.2 & 20.1 & 20.2 & 21.1 & 21.4 \\
 & {\scriptsize (19.3--21.0)} & {\scriptsize (19.3--21.0)} & {\scriptsize (19.4--21.0)} & {\scriptsize (20.3--22.0)} & {\scriptsize (20.5--22.2)} \\[4pt]
\multirow{2}{*}{Unknown language} & 27.8 & 29.5 & 26.8 & 23.7 & 26.4 \\
 & {\scriptsize (22.3--33.6)} & {\scriptsize (23.9--35.5)} & {\scriptsize (21.5--32.4)} & {\scriptsize (18.9--28.9)} & {\scriptsize (21.3--31.9)} \\[4pt]
\bottomrule
\end{tabular}}
\end{table}
Standardizing for centrality, TLD or length separately shifts no language level with a known language by more than \Result{language.std.maxshift} percentage points, so no single dimension accounts for the language differences, and standardizing for all three at once moves the three plotted groups by 0.3, 0.2 and 1.2 points. Pages with unknown language, about 1--2\% of the sample, shift more because their estimates are imprecise.

\section{Sensitivity analysis}\label{sec:sensitivity_app}
The reported model is the additive model of Section~\ref{sec:modeling}, with full specificity effects, a static shift of detector performance after the control crawls with scales $s_{\delta, \Sp} = 0.5$ for specificity and $s_{\delta, \Se} = 0.25$ for sensitivity, the prior $\mu \sim \Normal(\logit(0.01), 1)$ for the intercept, a temporal effect with $R = 12$ basis functions specifying the spline $f$, and no interactions. We fitted 16 variants that differ from the reported model in one or two of the components described below. Table~\ref{tab:sensitivity} lists the settings of all 17 fits together with their estimates, and Table~\ref{tab:waic} compares them by Watanabe-Akaike Information Criterion (WAIC).

\subsection{Model variants}\label{sec:model_variants_app}
\paragraph{Reduced specificity effects.} In the reduced structure, the control-period specificity of Pangram depends on word count and TLD, and that of Binoculars on word count only,
\begin{equation}\label{eq:control_sp_reduced}
    \logit (\psi^{\Sp, 0}_{d, c}) = \nu_{d, \ell[c]} + u^{\text{wc}}_{d, w[c]} + \mathbf{1}\{d = \text{open-pangram}\} \, u^{\text{tld}}_{d, b[c]} \;,
\end{equation}
that is, \eqref{eq:control_sp} without the tier effects for both detectors and without the TLD effects for Binoculars, with the priors in \eqref{eq:sp_prior} for the effects that remain.

\paragraph{Constrained by positive Youden index.} This model explicitly imposes the constraint $J_{d, c} > 0$ of Section~\ref{sec:calibration-prior} on the control-period calibration instead of post-hoc inspection, by requiring $\psi^{\Se, 0}_{d, s} > 1 - \psi^{\Sp, 0}_{d, c}$ in every observed cell $c$ with $s[\ell[c]] = s$. Writing $\eta_{d, s} = -\min_{c} \logit(\psi^{\Sp, 0}_{d, c}) + g_{d, s}$, this is the condition $g_{d, s} > 0$, and the prior on $\eta_{d, s}$ in \eqref{eq:se_prior} is truncated to this region. Appendix~\ref{sec:youden_constraint} describes the results.

\paragraph{Interactions.}\label{sec:interactions}
The effect of a domain-level attribute like the centrality tier or the TLD on prevalence is not necessarily the same for every language, and the effect of the centrality tier is not necessarily the same for every TLD. We therefore extend the model in Equation~\eqref{eq:theta_model} by adding two-way interaction terms between centrality tiers, TLD and languages,
\begin{equation}\label{eq:theta_model_interactions}
\begin{split}
    \logit(\theta_{c,k}) &= \mu + \alpha^{\text{tier}}_{j[c]} + \alpha^{\text{tld}}_{b[c]} + \alpha^{\text{wc}}_{w[c]} + \alpha^{\text{lang}}_{\ell[c]} \\
    & \qquad + \omega^{\text{tier} \times \text{lang}}_{j[c], \ell[c]} + \omega^{\text{tld} \times \text{lang}}_{b[c], \ell[c]} + \omega^{\text{tier} \times \text{tld}}_{j[c], b[c]} + \tau_{c,k}, \quad \forall c \in \calC, \ k \in \calK \;.
\end{split}
\end{equation}
Interactions are defined only for combinations of attribute levels that occur among the cells in $\calC$ and in which neither attribute is at its baseline level. Each observed interaction effect is given a centered prior with its own scale,
\begin{equation}\label{eq:interaction_prior}
\begin{split}
    \omega^{\text{tier} \times \text{lang}}_{j, \ell} \given \sigma_{\text{tier} \times \text{lang}} &\sim \Normal \left( 0, \sigma^2_{\text{tier} \times \text{lang}} \right), \quad (j, \ell) \in \bar{\calJ} \times \bar{\calL} \;,\\
    \omega^{\text{tld} \times \text{lang}}_{b, \ell} \given \sigma_{\text{tld} \times \text{lang}} &\sim \Normal \left( 0, \sigma^2_{\text{tld} \times \text{lang}} \right), \quad (b, \ell) \in \bar{\calB} \times \bar{\calL} \;,\\
    \omega^{\text{tier} \times \text{tld}}_{j, b} \given \sigma_{\text{tier} \times \text{tld}} &\sim \Normal \left( 0, \sigma^2_{\text{tier} \times \text{tld}} \right), \quad (j, b) \in \bar{\calJ} \times \bar{\calB} \;,\\
    \sigma_{\text{tier} \times \text{lang}}, \sigma_{\text{tld} \times \text{lang}}, \sigma_{\text{tier} \times \text{tld}} &\sim \halfNormal(0, 1) \;.
\end{split}
\end{equation}
Since the scale parameters are jointly estimated, the data determine how much the interactions are used. As the weight on the data increases with the number of pages in the combination, so effects for sparse combinations stay close to zero unless the data strongly indicate otherwise. With the interactions, the prior dependence between cells described in the remark of Section~\ref{sec:modeling} extends to cells that share a combination of attributes, as the interactions let the strength of one attribute's effect depend on the level of another.

\paragraph{No temporal effect.} This variant sets the temporal effect in \eqref{eq:theta_model} to zero, $\tau_{c, k} = 0$ for all $c \in \calC$ and $k \in \calK$, so that the prevalence of a cell does not change over time. There is then no trend $f$ and there are no language, tier or TLD slopes in \eqref{eq:tau-model}. The specificity, the sensitivity and the drift of Section~\ref{sec:calibration-prior} are unchanged, so that a change in the flag rates over the crawls can be explained only through the drift of the detectors.

\paragraph{Number of basis functions.} In the reported model, the knot vector of the spline $f$ in Section~\ref{sec:temporal} has the two interior knots $\kappa_1$ and $\kappa_2$ and one additional interior knot for every 10 crawls within each regime, which gives $R = 12$ basis functions. We vary the number of additional knots, with one for every 5, 15 and 20 crawls, which gives $R = 20$, $10$ and $8$ basis functions, and with no additional knots, which leaves the two knots $\kappa_1$ and $\kappa_2$ and gives $R = 6$. The prior on the spline coefficients $\gamma$ and the rest of the model are unchanged.

\paragraph{Scales of the drift and priors on early prevalence.} We also assess the sensitivity of the scale parameters that dictates the prior dispersion of the drift terms, as well as the prior on the global intercept $\mu$. The variants in the scale of the drift considers $s_{\delta, \Sp} = 0.25, 0.75$, and $s_{\delta, \Se} = 0.1, 0.5$, as given by \eqref{eq:drift_static}. Regarding the prior of the intercept, we consider $\varepsilon_{\mu} = 0.05$, and $s_{\mu} = 2$, as given by \eqref{eq:mu_prior}. Each variant is constructed by changing one of these values at a time, relative to the hyperparameters of the model reported in the main article.

\paragraph{No drift.} We also consider the model variant without a drift $\delta^{\Se}_{d, s, k} = \delta^{\Sp}_{d, \ell, k} = 0$ for all crawls. This setting corresponds to the model where we assume that the control-period calibration transports exactly to the later crawls ($s_{\delta, \Sp} = s_{\delta, \Se} = 0$).

\paragraph{Dynamic drift.} In contrast to the static drift in \eqref{eq:drift_static}, this model assumes that the drift follows a Gaussian first-order random walk started at zero after the last control crawl,
\begin{equation}\label{eq:drift_rw}
\begin{split}
    \delta^{\Se}_{d, s, K_0 + 1} &\sim \Normal(0, s_{\delta, \text{rw}}^2) \;, \quad \delta^{\Se}_{d, s, k} \given \delta^{\Se}_{d, s, k-1} \sim \Normal \left( \delta^{\Se}_{d, s, k-1}, s_{\delta, \text{rw}}^2 \right),\\
    \delta^{\Sp}_{d, \ell, K_0 + 1} &\sim \Normal(0, s_{\delta, \text{rw}}^2) \;, \quad \delta^{\Sp}_{d, \ell, k} \given \delta^{\Sp}_{d, \ell, k-1} \sim \Normal \left( \delta^{\Sp}_{d, \ell, k-1}, s_{\delta, \text{rw}}^2 \right),
\end{split}
\end{equation}
for $k = K_0 + 2, \ldots, K$. Thus, the operating characteristics evolve gradually, with uncertainty that grows with the distance from the control period. We take $s_{\delta, \text{rw}} = 0.1$ per crawl as the primary value. We estimated this model only on a pilot subsample of the data.

\subsection{Model comparison}\label{sec:model_comparison_app}

The reported model lets detector sensitivity and specificity drift once after the control crawls, with scale $s_{\delta, \Se} = 0.25$ for sensitivity and $s_{\delta, \Sp} = 0.5$ for specificity, lets specificity depend on word count, centrality tier and TLD for both detectors, and includes a temporal effect with $R = 12$ basis functions. We refitted the model under 16 alternative specifications on the full data (Tables~\ref{tab:sensitivity} and~\ref{tab:waic}). Each fit has four chains with 2,000 draws totalling 8,000 posterior samples, except the fit with $s_{\delta, \Sp} = 0.75$, which has four chains with 1,750 draws totalling 7,000 posterior samples.

The estimates depend on how detector performance is allowed to change after the control crawls and on the temporal effect, and we examine this across specifications that vary the structure of specificity, the scale of the drifts in sensitivity and specificity, the priors on early prevalence and on the intercept, the interactions among tier, TLD and language, the presence of a temporal effect and its number of basis functions. The August 2026 prevalence was between 22.3\% and 24.2\% under every specification that allows for drift and includes a temporal effect, including the reported \Result{overall.latest.mean}\%. It was 27.7\% when the drift is removed altogether, which fits the data much worse, and 5.2\% when the temporal effect is removed, which fits worse still. The level at the end of the series is therefore the quantity most sensitive to these assumptions, as we discuss in Section~\ref{sec:limitations}.

\begin{table}[t]
\centering
\caption{Settings and prevalence in the August 2026 snapshot under the reported model and 16 alternative specifications. Specificity effects are full (centrality tier, word count and TLD for both detectors) or reduced (word count for both detectors and TLD for Pangram); $R$ is the number of basis functions of the spline in the temporal effect (-- if there is no temporal effect); $s_{\delta, \Sp}$ and $s_{\delta, \Se}$ are the scales of the drifts in specificity and sensitivity after the control crawls (-- if there is no drift), and $\varepsilon_{\mu}$ and $s_{\mu}$ the location and scale of the prior of the intercept, $\mu \sim \Normal(\logit(\varepsilon_{\mu}), s_{\mu}^2)$. Prevalence is the posterior mean with its 95\% credible interval. Table~\ref{tab:waic} compares the fits by WAIC.}
\label{tab:sensitivity}
\small
\resizebox{0.95\textwidth}{!}{%
\begin{tabular}{lcccccccc}
\toprule
Model specification & Specificity model & Interactions & $R$ & $s_{\delta, \Sp}$ & $s_{\delta, \Se}$ & $\varepsilon_{\mu}$ & $s_{\mu}$ & Prevalence, \% \\
\midrule
\emph{Reported model} & Full & No & 12 & 0.5 & 0.25 & 0.01 & 1 & 23.6 (22.9--24.4) \\
Full $\Sp$, Youden constraint & Full & No & 12 & 0.5 & 0.25 & 0.01 & 1 & 22.6 (21.9--23.3) \\
Full $\Sp$, with interactions & Full & Yes & 12 & 0.5 & 0.25 & 0.01 & 1 & 23.9 (23.0--25.3) \\
Reduced $\Sp$ & Reduced & No & 12 & 0.5 & 0.25 & 0.01 & 1 & 22.9 (22.2--23.6) \\
$s_{\delta, \Sp} = 0.25$ & Reduced & No & 12 & 0.25 & 0.25 & 0.01 & 1 & 23.1 (22.4--23.8) \\
$\Sp$, $s_{\delta, \Sp} = 0.75$ & Reduced & No & 12 & 0.75 & 0.25 & 0.01 & 1 & 22.8 (22.2--23.5) \\
$s_{\delta, \Se} = 0.1$ & Reduced & No & 12 & 0.5 & 0.1 & 0.01 & 1 & 24.2 (23.6--24.9) \\
$s_{\delta, \Se} = 0.5$ & Reduced & No & 12 & 0.5 & 0.5 & 0.01 & 1 & 22.3 (21.6--23.0) \\
$\varepsilon_{\mu} = 0.05$ & Reduced & No & 12 & 0.5 & 0.25 & 0.05 & 1 & 22.9 (22.2--23.6) \\
$s_{\mu} = 2$ & Reduced & No & 12 & 0.5 & 0.25 & 0.01 & 2 & 22.9 (22.2--23.6) \\
Reduced $\Sp$, interactions & Reduced & Yes & 12 & 0.5 & 0.25 & 0.01 & 1 & 23.7 (22.9--24.7) \\
No post-2022 drift & Reduced & No & 12 & -- & -- & 0.01 & 1 & 27.7 (27.1--28.3) \\
No temporal effect & Full & No & -- & 0.5 & 0.25 & 0.01 & 1 & 5.2 (5.1--5.4) \\
Full $\Sp$, $R = 6$ & Full & No & 6 & 0.5 & 0.25 & 0.01 & 1 & 23.0 (22.3--23.7) \\
Full $\Sp$, $R = 8$ & Full & No & 8 & 0.5 & 0.25 & 0.01 & 1 & 23.8 (23.1--24.6) \\
Full $\Sp$, $R = 10$ & Full & No & 10 & 0.5 & 0.25 & 0.01 & 1 & 23.0 (22.2--23.9) \\
Full $\Sp$, $R = 20$ & Full & No & 20 & 0.5 & 0.25 & 0.01 & 1 & 22.5 (21.8--23.2) \\
\bottomrule
\end{tabular}}
\end{table}

\paragraph{Computing the WAIC.} We compare the fits by the Watanabe--Akaike information criterion \citep{watanabe_2010}, computed from the web-page likelihood \eqref{eq:marginal}, with the latent status $Z_i$ marginalized out. The benchmark counts $\calX$ are the same data in every fit and do not enter. Let $M$ be the number of posterior draws ($M = 8{,}000$, or $7{,}000$ for the fit with $s_{\delta, \Sp} = 0.75$), and let $\theta^{(m)}_{c, k}$ and $\psi^{(m)}_{d, c, k}$ be the prevalence and the operating characteristics implied by draw $m = 1, \ldots, M$ through \eqref{eq:theta_model} and \eqref{eq:drift}. Writing $p(Y_i \given \theta^{(m)}, \psi^{(m)}) = p(Y_i \given \theta^{(m)}_{c[i], k[i]}, \psi^{(m)}_{c[i], k[i]})$ for \eqref{eq:marginal} evaluated at draw $m$, the Monte Carlo estimates of the log pointwise predictive density and of the effective number of parameters are
\begin{equation*}
    \widehat{\text{lppd}} = \sum_{i = 1}^{n} \log \left( \frac{1}{M} \sum_{m = 1}^{M} p(Y_i \given \theta^{(m)}, \psi^{(m)}) \right) \;, \qquad
    \hat{p}_{\text{WAIC}} = \sum_{i = 1}^{n} \mathrm{Var}_{m} \left( \log p(Y_i \given \theta^{(m)}, \psi^{(m)}) \right) \;,
\end{equation*}
where $\mathrm{Var}_{m}$ is the sample variance over the $M$ draws, with divisor $M - 1$. That is, $\widehat{\text{lppd}}$ is the logarithm of the average likelihood over the draws, summed over pages, and $\hat{p}_{\text{WAIC}}$ is the variance of the log-likelihood over the draws, summed over pages. Then
\begin{equation*}
    \text{WAIC} = -2 \left( \widehat{\text{lppd}} - \hat{p}_{\text{WAIC}} \right) \;.
\end{equation*}
Pages in the same cell and crawl with the same detector outputs have the same likelihood, so each term is computed once per unique combination and weighted by the number of such pages, and all sums run over the $n \approx 5.1$ million pages. Lower WAIC indicates a better fit.

\begin{table}[t]
\centering
\caption{Page-level WAIC under the reported model and sixteen alternative specifications, computed from all posterior draws. $\Delta\widehat{\text{lppd}}$ and $\Delta$WAIC are differences from the reported model, whose values are $\widehat{\text{lppd}} = -2{,}157{,}455$, $\hat{p}_{\text{WAIC}} = 77.9$ and $\text{WAIC} = 4{,}315{,}065$; $\hat{p}_{\text{WAIC}}$ is shown as is. Since $\Delta\text{WAIC} = -2(\Delta\widehat{\text{lppd}} - \Delta\hat{p}_{\text{WAIC}})$, lower WAIC indicates a better fit and negative values of $\Delta$WAIC indicate a better fit than the reported model.}
\label{tab:waic}
\small
\resizebox{0.95\textwidth}{!}{%
\begin{tabular}{lrrr}
\toprule
Model specification & $\Delta\widehat{\text{lppd}}$ & $\hat{p}_{\text{WAIC}}$ & $\Delta$WAIC \\
\midrule
\emph{Reported model (full specificity effects)} & 0 & 77.9 & 0 \\
Full specificity effects, Youden constraint & $-274$ & 75.9 & $+544$ \\
Full specificity effects, with interactions & $+882$ & 129.4 & $-1{,}662$ \\
Reduced specificity effects & $-1{,}205$ & 62.8 & $+2{,}380$ \\
Reduced specificity effects, $s_{\delta, \Sp} = 0.25$ & $-1{,}221$ & 64.1 & $+2{,}414$ \\
Reduced specificity effects, $s_{\delta, \Sp} = 0.75$ & $-1{,}200$ & 63.0 & $+2{,}370$ \\
Reduced specificity effects, $s_{\delta, \Se} = 0.1$ & $-1{,}250$ & 63.1 & $+2{,}470$ \\
Reduced specificity effects, $s_{\delta, \Se} = 0.5$ & $-1{,}192$ & 63.3 & $+2{,}355$ \\
Reduced specificity effects, $\varepsilon_{\mu} = 0.05$ & $-1{,}205$ & 63.1 & $+2{,}380$ \\
Reduced specificity effects, $s_{\mu} = 2$ & $-1{,}205$ & 62.7 & $+2{,}379$ \\
Reduced specificity effects, with interactions & $-407$ & 103.7 & $+866$ \\
Reduced specificity effects, no drift after the control crawls & $-4{,}110$ & 58.1 & $+8{,}180$ \\
Full specificity effects, no temporal effect & $-36{,}343$ & 54.4 & $+72{,}639$ \\
Full specificity effects, $R = 6$ basis functions & $-338$ & 72.4 & $+666$ \\
Full specificity effects, $R = 8$ basis functions & $-128$ & 75.2 & $+250$ \\
Full specificity effects, $R = 10$ basis functions & $-26$ & 80.4 & $+57$ \\
Full specificity effects, $R = 20$ basis functions & $+171$ & 84.2 & $-330$ \\
\bottomrule
\end{tabular}}
\end{table}

Seven results stand out. First, allowing the specificity to depend on word count, centrality tier and TLD for both detectors lowers the WAIC by 2,380 relative to the reduced structure and moves the estimate by 0.7 percentage points, from 22.9\% to 23.6\%. Second, relative to the reduced structure, the scale of the specificity drift (0.25 and 0.75), the scale of the sensitivity drift (0.1 and 0.5) and the priors on early prevalence and on the intercept change the estimate by at most 1.3 percentage points, and differ in WAIC by less than 100, which the data cannot distinguish. Third, removing the drift altogether raises the estimate to 27.7\% and raises the WAIC by 5,800 relative to the reduced structure (8,180 relative to the reported model), so the drift is needed. Fourth, removing the temporal effect raises the WAIC by 72,639 relative to the reported model and lowers the estimate to 5.2\% (5.1--5.4\%). Without a temporal effect, the rise in the flag rates over the crawls can be explained only through the drift of the detectors, and the fit does so with a Youden index below zero for Pangram in every cell of English-stratum pages (Appendix~\ref{sec:model_diagnostics}), so that its calibration is not meaningful and we do not interpret its other estimates. Fifth, the WAIC improves with the number of basis functions: relative to $R = 12$ it is higher by 57, 250 and 666 for $R = 10$, $8$ and $6$, and lower by 330 for $R = 20$, while the effective number of parameters rises only from 72 to 84, and the August 2026 estimate stays between 22.5\% and 23.8\% in these four fits. These fits also move the crossing of bilingual above no-English prevalence (Section~\ref{sec:sens_crossing}) earlier, from February 2026 in the reported model to December 2025 in all four. We report $R = 12$, as the gain from $R = 20$ is about $6 \times 10^{-5}$ per page and changes the estimate by 1.1 percentage points. Sixth, adding the three blocks of interactions lowers the WAIC under both structures, by 1,514 relative to the reduced structure and by 1,662 relative to the reported model, and the effective number of parameters of the WAIC rises from 78 to 129 under the full structure. It moves the overall estimate by 0.3 percentage points under the full structure (from 23.6\% to 23.9\%) and by 0.8 under the reduced structure (from 22.9\% to 23.7\%). Relative to the reported model, the interactions leave the language, centrality-tier and length estimates within 1.2 percentage points, but move the estimates for \texttt{.net} from 20.4\% to 24.8\% and for \texttt{.org} from 14.6\% to 16.6\%, and widen the interval of the non-English-only level, from 15.4--18.4\% to 15.2--20.5\%. We report the additive model, because the gain in WAIC is small per page (about $3.2 \times 10^{-4}$) and the interactions lowers the effective sample sizes (smallest tail ESS of 215 against 736 for the reported model). Seventh, imposing a positive Youden index (Appendix~\ref{sec:youden_constraint}) raises the WAIC by 544 relative to the reported model, lowers the August 2026 estimate to 22.6\% (21.9--23.3\%) and moves the estimate for pages with no English detected from 17.0\% to 14.5\%.

Finally, in a pilot with 100 pages per cell, we replaced the single drift in specificity with a Gaussian random walk. The data did not distinguish the two models (WAIC difference of about 22), but the random walk absorbed the late rise in Pangram flags into the specificity and produced a decline in prevalence late in the series, with an end-of-series level between 9\% and 22\% depending on the scale of the walk. This potentially arises from an identifiability issue: after the control crawls, a rise in the flag rate can be explained either by a rise in prevalence or by a fall in specificity, and when specificity is allowed to drift too freely it absorbs the rise, so that prevalence is determined largely by the prior on the walk. Reducing the number of epochs over which the walk evolves did not resolve it. We therefore report the static drift and treat the end-of-series level as conditional on the assumption that detector specificity changes once.

\subsection{Robustness of the main findings}\label{sec:sens_findings}

\begin{figure}[t]
\centering 
\includegraphics[width=0.95\linewidth]{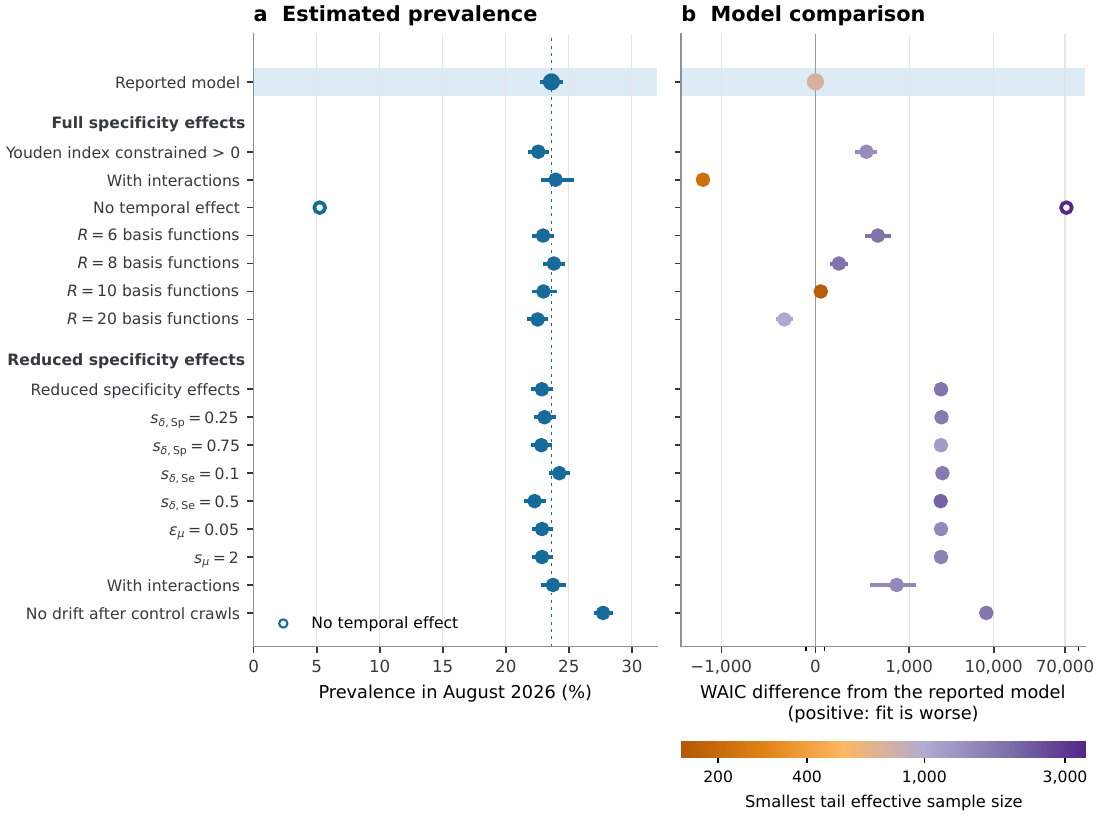}
\caption{Reported model and sixteen alternative specifications (settings in Table~\ref{tab:sensitivity}). (a) Estimated prevalence of pages with \Result{outcome.definition} in August 2026; points are posterior means and bars are 95\% credible intervals. (b) Difference in WAIC from the reported model; points are differences and bars are $\pm 1.96$ standard errors of the difference. The horizontal axis is symmetric logarithmic, and a positive difference indicates a worse fit. Point color is the smallest tail effective sample size of the fit over the monitored quantities (Table~\ref{tab:diagnostics}). The hollow marker is the fit without a temporal effect.}
\label{fig:fits_forest}
\end{figure}

Figure~\ref{fig:fits_forest} summarizes the August 2026 estimate and the fit of the reported model and the sixteen alternative specifications.

The substantive findings do not depend on which of the 16 alternative specifications is used, with two exceptions: removing the drift after the control crawls, which raises every level and which the data reject, and removing the temporal effect, which cannot reproduce the growth of prevalence (Tables~\ref{tab:sensitivity} and~\ref{tab:waic}). Across the 15 fits that allow drift and include a temporal effect, prevalence in August 2026 increases monotonically from the most to the least central tier in every fit, from 10.4--12.0\% in the Top tier to 25.8--28.3\% in the Tail (13.0\% and 31.8\% without the drift). \texttt{.com} is the highest of the six domain categories (28.4--30.5\%) and \texttt{.edu} and \texttt{.gov} the lowest (5.9--6.4\% and 5.5--6.3\%) in every fit, and long pages are 1.9 to 2.0 times as prevalent as short pages (1.8 without the drift). These estimates describe pages with at least 500 characters of extracted text. On a denominator of all English-language captures, with no text extraction, the same detector would flag roughly 13 to 19 percent, which brackets the 10 percent reported by Pew Research Center \citep{pew2026} (Appendix~\ref{sec:sensitivity_app}).

The language comparisons are more sensitive (Table~\ref{tab:sens_findings}). 
\begin{table}[t]
\centering
\caption{Language comparisons in the August 2026 snapshot under the reported model and sixteen alternative specifications (settings in Table~\ref{tab:sensitivity}). Each entry is the posterior mean, with its 95\% credible interval below, and prevalence is in percent. The English-only minus no-English difference is in percentage points; the change in the growth gap is the later minus the earlier annual growth rate of that difference (February 2023 to January 2025, January 2025 to August 2026), in percentage points per year; bilingual minus no English is the difference in percentage points. All intervals are computed within matched posterior draws.}
\label{tab:sens_findings}
\footnotesize
\resizebox{0.95\textwidth}{!}{%
\setlength{\tabcolsep}{5pt}
\begin{tabular}{p{5cm}*{6}{>{\centering\arraybackslash}p{1.6cm}}}
\toprule
Model specification & English only & No English & Bilingual & English minus no English & Change in growth gap & Bilingual minus no English \\
\midrule
\multirow{2}{4.9cm}{Reported model (full specificity effects)} & 32.8 & 17.0 & 20.2 & 15.8 & 3.0 & 3.2 \\
 & {\scriptsize \mbox{(32.2--33.4)}} & {\scriptsize \mbox{(15.4--18.4)}} & {\scriptsize \mbox{(19.3--21.0)}} & {\scriptsize \mbox{(14.4--17.4)}} & {\scriptsize \mbox{(2.8--3.2)}} & {\scriptsize \mbox{(1.6--4.9)}} \\[4pt]
\multirow{2}{4.9cm}{Full specificity effects, Youden constraint} & 32.7 & 14.5 & 19.9 & 18.2 & 3.0 & 5.4 \\ & {\scriptsize \mbox{(32.1--33.3)}} & {\scriptsize \mbox{(13.3--16.0)}} & {\scriptsize \mbox{(19.1--20.8)}} & {\scriptsize \mbox{(16.7--19.5)}} & {\scriptsize \mbox{(2.8--3.3)}} & {\scriptsize \mbox{(3.8--6.9)}} \\
\multirow{2}{4.9cm}{Full specificity effects, with interactions} & 32.7 & 17.4 & 21.0 & 15.2 & 2.7 & 3.6 \\
 & {\scriptsize \mbox{(32.1--33.3)}} & {\scriptsize \mbox{(15.2--20.5)}} & {\scriptsize \mbox{(20.1--22.0)}} & {\scriptsize \mbox{(12.2--17.4)}} & {\scriptsize \mbox{(2.4--3.0)}} & {\scriptsize \mbox{(0.5--5.9)}} \\[4pt]
\multirow{2}{4.9cm}{Reduced specificity effects} & 33.1 & 14.8 & 20.2 & 18.3 & 2.7 & 5.4 \\
 & {\scriptsize \mbox{(32.5--33.7)}} & {\scriptsize \mbox{(13.5--16.2)}} & {\scriptsize \mbox{(19.3--21.0)}} & {\scriptsize \mbox{(16.8--19.6)}} & {\scriptsize \mbox{(2.5--3.0)}} & {\scriptsize \mbox{(3.8--6.8)}} \\[4pt]
\multirow{2}{4.9cm}{Reduced specificity effects, $s_{\delta, \Sp} = 0.25$} & 33.0 & 15.4 & 20.1 & 17.7 & 2.7 & 4.8 \\
 & {\scriptsize \mbox{(32.5--33.6)}} & {\scriptsize \mbox{(14.0--16.9)}} & {\scriptsize \mbox{(19.3--21.0)}} & {\scriptsize \mbox{(16.1--19.1)}} & {\scriptsize \mbox{(2.4--3.0)}} & {\scriptsize \mbox{(3.1--6.3)}} \\[4pt]
\multirow{2}{4.9cm}{Reduced specificity effects, $s_{\delta, \Sp} = 0.75$} & 33.1 & 14.7 & 20.2 & 18.4 & 2.8 & 5.5 \\
 & {\scriptsize \mbox{(32.5--33.7)}} & {\scriptsize \mbox{(13.4--16.0)}} & {\scriptsize \mbox{(19.3--21.1)}} & {\scriptsize \mbox{(17.1--19.7)}} & {\scriptsize \mbox{(2.5--3.0)}} & {\scriptsize \mbox{(4.0--7.0)}} \\[4pt]
\multirow{2}{4.9cm}{Reduced specificity effects, $s_{\delta, \Se} = 0.1$} & 33.0 & 18.4 & 20.3 & 14.6 & 2.3 & 1.9 \\
 & {\scriptsize \mbox{(32.4--33.6)}} & {\scriptsize \mbox{(17.1--19.7)}} & {\scriptsize \mbox{(19.4--21.2)}} & {\scriptsize \mbox{(13.3--15.9)}} & {\scriptsize \mbox{(2.1--2.5)}} & {\scriptsize \mbox{(0.4--3.4)}} \\[4pt]
\multirow{2}{4.9cm}{Reduced specificity effects, $s_{\delta, \Se} = 0.5$} & 33.1 & 13.3 & 20.2 & 19.9 & 3.0 & 6.9 \\
 & {\scriptsize \mbox{(32.5--33.7)}} & {\scriptsize \mbox{(12.0--14.7)}} & {\scriptsize \mbox{(19.3--21.0)}} & {\scriptsize \mbox{(18.4--21.2)}} & {\scriptsize \mbox{(2.7--3.2)}} & {\scriptsize \mbox{(5.3--8.5)}} \\[4pt]
\multirow{2}{4.9cm}{Reduced specificity effects, $\varepsilon_{\mu} = 0.05$} & 33.1 & 14.8 & 20.2 & 18.3 & 2.7 & 5.4 \\
 & {\scriptsize \mbox{(32.5--33.7)}} & {\scriptsize \mbox{(13.6--16.2)}} & {\scriptsize \mbox{(19.3--21.0)}} & {\scriptsize \mbox{(16.8--19.5)}} & {\scriptsize \mbox{(2.5--3.0)}} & {\scriptsize \mbox{(3.8--6.8)}} \\[4pt]
\multirow{2}{4.9cm}{Reduced specificity effects, $s_{\mu} = 2$} & 33.1 & 14.8 & 20.2 & 18.3 & 2.7 & 5.4 \\
 & {\scriptsize \mbox{(32.5--33.7)}} & {\scriptsize \mbox{(13.6--16.2)}} & {\scriptsize \mbox{(19.3--21.0)}} & {\scriptsize \mbox{(16.8--19.5)}} & {\scriptsize \mbox{(2.5--3.0)}} & {\scriptsize \mbox{(3.8--6.8)}} \\[4pt]
\multirow{2}{4.9cm}{Reduced specificity effects, with interactions} & 32.7 & 17.5 & 20.0 & 15.2 & 2.4 & 2.5 \\
 & {\scriptsize \mbox{(32.1--33.3)}} & {\scriptsize \mbox{(15.7--19.6)}} & {\scriptsize \mbox{(19.0--21.0)}} & {\scriptsize \mbox{(13.1--17.1)}} & {\scriptsize \mbox{(2.1--2.7)}} & {\scriptsize \mbox{(0.2--4.5)}} \\[4pt]
\multirow{2}{4.9cm}{Reduced specificity effects, no drift after the control crawls} & 33.0 & 22.3 & 27.8 & 10.7 & 1.5 & 5.5 \\
 & {\scriptsize \mbox{(32.4--33.6)}} & {\scriptsize \mbox{(21.4--23.4)}} & {\scriptsize \mbox{(27.0--28.6)}} & {\scriptsize \mbox{(9.7--11.6)}} & {\scriptsize \mbox{(1.2--1.8)}} & {\scriptsize \mbox{(4.3--6.6)}} \\[4pt]
\multirow{2}{4.9cm}{Full specificity effects, no temporal effect} & 8.6 & 3.5 & 2.9 & 5.1 & $-$0.08 & $-$0.6 \\
 & {\scriptsize \mbox{(8.4--8.9)}} & {\scriptsize \mbox{(3.4--3.7)}} & {\scriptsize \mbox{(2.7--3.1)}} & {\scriptsize \mbox{(4.8--5.4)}} & {\scriptsize \mbox{($-$0.09,$-$0.07)}} & {\scriptsize \mbox{($-$0.9,$-$0.4)}} \\[4pt]
\multirow{2}{4.9cm}{Full specificity effects, $R = 6$ basis functions} & 32.2 & 15.8 & 20.2 & 16.4 & 2.1 & 4.4 \\
 & {\scriptsize \mbox{(31.7--32.7)}} & {\scriptsize \mbox{(14.3--17.5)}} & {\scriptsize \mbox{(19.4--21.0)}} & {\scriptsize \mbox{(14.7--18.0)}} & {\scriptsize \mbox{(1.9--2.3)}} & {\scriptsize \mbox{(2.6--6.1)}} \\[4pt]
\multirow{2}{4.9cm}{Full specificity effects, $R = 8$ basis functions} & 33.6 & 16.2 & 21.0 & 17.4 & 2.5 & 4.7 \\
 & {\scriptsize \mbox{(33.1--34.1)}} & {\scriptsize \mbox{(14.7--17.9)}} & {\scriptsize \mbox{(20.1--21.8)}} & {\scriptsize \mbox{(15.7--18.9)}} & {\scriptsize \mbox{(2.3--2.8)}} & {\scriptsize \mbox{(2.9--6.4)}} \\[4pt]
\multirow{2}{4.9cm}{Full specificity effects, $R = 10$ basis functions} & 32.8 & 15.4 & 20.1 & 17.3 & 2.7 & 4.7 \\
 & {\scriptsize \mbox{(32.2--33.3)}} & {\scriptsize \mbox{(13.9--17.4)}} & {\scriptsize \mbox{(19.3--21.0)}} & {\scriptsize \mbox{(15.3--18.9)}} & {\scriptsize \mbox{(2.4--2.9)}} & {\scriptsize \mbox{(2.7--6.4)}} \\[4pt]
\multirow{2}{4.9cm}{Full specificity effects, $R = 20$ basis functions} & 32.3 & 15.1 & 19.5 & 17.2 & 3.1 & 4.5 \\
 & {\scriptsize \mbox{(31.6--32.9)}} & {\scriptsize \mbox{(13.9--16.4)}} & {\scriptsize \mbox{(18.7--20.4)}} & {\scriptsize \mbox{(15.9--18.4)}} & {\scriptsize \mbox{(2.9--3.4)}} & {\scriptsize \mbox{(3.0--5.8)}} \\[4pt]
\bottomrule
\end{tabular}}
\end{table}
English-only prevalence is stable, between 32.2\% and 33.6\% in the fifteen fits that allow drift and include a temporal effect (33.0\% without the drift), and so is bilingual prevalence (19.5--21.0\%). Prevalence of pages with no English detected ranges from 13.3\% to 18.4\% across those fits, and the English-only advantage from 14.6 to 19.9 percentage points (a ratio of 1.8 to 2.5), but the advantage is positive in every specification with a temporal effect and so is the increase in the English advantage in annual growth between the two intervals (1.5 to 3.1 percentage points per year, with intervals that exclude zero in all sixteen fits with a temporal effect). Bilingual prevalence exceeds no-English prevalence in August 2026 in every one of these fits (by 1.9 to 6.9 points, with intervals that exclude zero). The fit without a temporal effect shows none of these patterns: it estimates 8.6\% for English-only pages and 3.5\% for pages with no English detected, and the change in the growth gap is -0.08 points per year (95\% credible interval -0.09 to -0.07).

\subsection{The level of non-English prevalence}\label{sec:sens_noneng}

The estimate for pages with no English detected is the quantity most affected by calibration. Among the specifications with the reduced specificity structure it falls from 18.4\% when the sensitivity drift is constrained to $s_{\delta, \Se} = 0.1$ to 14.8\% at 0.25 and 13.3\% at 0.5, while the English-only estimate does not move (33.0--33.1\%). The reported model, which adds cell-level specificity effects, gives 17.0\%, adding the interactions to it gives 17.4\% (15.2--20.5\%), imposing a positive Youden index gives 14.5\% (13.3--16.0\%), and the fits with 6 to 20 basis functions give 15.1\% to 16.2\%. Two features of the design explain why. Pages with no English detected are scored by one detector only (Pangram), whereas English-stratum pages are also scored by Binoculars, so for these pages a change in the flag rate after the control crawls cannot be separated from a change in the detector's operating characteristics except through the priors on the drifts; and non-English sensitivity is calibrated on 2023-era generators and treated as a lower bound (Section~\ref{sec:calib_datasets}).

The posterior drifts show the same pattern. In the reported model the logit drift of Pangram's sensitivity after the control crawls is $-0.90$ (posterior standard deviation 0.17) for English-stratum pages and $+0.78$ (0.10) for non-English pages, and the logit drift in Pangram's specificity for pages with no English detected is $+0.68$ (0.60). That last drift is the least well identified calibration parameter: its posterior standard deviation is 0.60 in the reported model against 0.27 under the reduced structure, and its posterior mean falls from 2.37 to 0.68 when the specificity effects are added; across the other fits with the reduced structure it ranges from 1.44 to 2.97, and it is 2.28 (0.28) when a positive Youden index is imposed. We therefore regard the English--no-English gap as robustly positive and growing but its magnitude as conditional on these assumptions.

\subsection{Timing of the bilingual crossing}\label{sec:sens_crossing}

Bilingual prevalence rises more steeply than no-English prevalence through 2025 and 2026. In the reported model the difference is $-2.7$ points in July 2025, $-0.8$ in December 2025, $+0.9$ in February 2026 and $+3.2$ in August 2026, so there is no discrete change to date, and the crawl at which the posterior mean difference first becomes non-negative depends on the specification, from August 2025 to April 2026 (Table~\ref{tab:sens_crossing}), and the difference never becomes non-negative in the fit without a temporal effect. 
\begin{table}[t]
\centering
\caption{When bilingual prevalence first exceeds no-English prevalence under each specification (settings in Table~\ref{tab:sensitivity}): the first crawl at which the posterior mean difference is non-negative, and the first crawl at which the difference is positive with posterior probability above 0.975.}
\label{tab:sens_crossing}
\small
\resizebox{0.95\textwidth}{!}{%
\begin{tabular}{lcc}
\toprule
Model specification & Mean difference non-negative & Credibly positive \\
\midrule
\emph{Reported model (full specificity effects)} & February 2026 & April 2026 \\
Full specificity effects, Youden constraint & October 2025 & February 2026 \\
Full specificity effects, with interactions & December 2025 & May 2026 \\
Reduced specificity effects & October 2025 & January 2026 \\
Reduced specificity effects, $s_{\delta, \Sp} = 0.25$ & October 2025 & February 2026 \\
Reduced specificity effects, $s_{\delta, \Sp} = 0.75$ & September 2025 & December 2025 \\
Reduced specificity effects, $s_{\delta, \Se} = 0.1$ & April 2026 & June 2026 \\
Reduced specificity effects, $s_{\delta, \Se} = 0.5$ & August 2025 & September 2025 \\
Reduced specificity effects, $\varepsilon_{\mu} = 0.05$ & October 2025 & January 2026 \\
Reduced specificity effects, $s_{\mu} = 2$ & October 2025 & January 2026 \\
Reduced specificity effects, with interactions & February 2026 & June 2026 \\
Reduced specificity effects, no drift after the control crawls & September 2025 & October 2025 \\
Full specificity effects, no temporal effect & never & never \\
Full specificity effects, $R = 6$ basis functions & December 2025 & February 2026 \\
Full specificity effects, $R = 8$ basis functions & December 2025 & February 2026 \\
Full specificity effects, $R = 10$ basis functions & December 2025 & February 2026 \\
Full specificity effects, $R = 20$ basis functions & December 2025 & February 2026 \\
\bottomrule
\end{tabular}}
\end{table}
In the reported model this is February 2026, and the difference is credibly positive from April 2026. The fits with 6, 8, 10 and 20 basis functions place both dates earlier, at December 2025 and February 2026. We do not attribute the crossing to any event: the series is monthly and smooth, and changes in the calibration of the non-English detectors and in the flexibility of the trend move its timing, as the variation across specifications shows.

\subsection{Positive Youden index constraint}\label{sec:youden_constraint}

Section~\ref{sec:calibration-prior} does not constrain the control-period sensitivities relative to the specificities. We also fitted the reported model with $J_{d, c} > 0$ imposed (Section~\ref{sec:model_variants_app}), which is the conventional device against label switching. The constraint is not active for Pangram: its sensitivity parameters lie 4.0 (English-stratum pages) and 4.5 (non-English pages) logit units above the bound, and its Youden indices in the least specific cell are nearly unchanged (0.58 and 0.36 against 0.58 and 0.37, Table~\ref{tab:diagnostics}). It is active for Binoculars. The posterior of the slack $g_{d, s}$ for Binoculars is concentrated at zero (median 0.001, 97.5th percentile 0.005), so that the index in the least specific cell is $0.000$, and the control-period sensitivity of Binoculars is 7.6\% rather than 4.6\% in the reported model. In other words, the web data place Binoculars at or just below chance in the least specific cell (posterior mean of $J$ of $-0.06$, Table~\ref{tab:diagnostics}), and the constraint holds it at the boundary. This sensitivity is below the benchmark true-positive rates for Binoculars, which range from 34\% to 73\% across the benchmark corpora (Section~\ref{detection}), whereas among the pages that Pangram flags in the most recent crawls, only 8--13\% are also flagged by Binoculars. The benchmark counts therefore pull the sensitivity up and the web pages pull it down; the sensitivity drift allows the post-control sensitivity to differ from the control-period value, and the estimated drift is positive (1.54 logit units).

\begin{figure}[t]
\centering \includegraphics[width=0.85\linewidth]{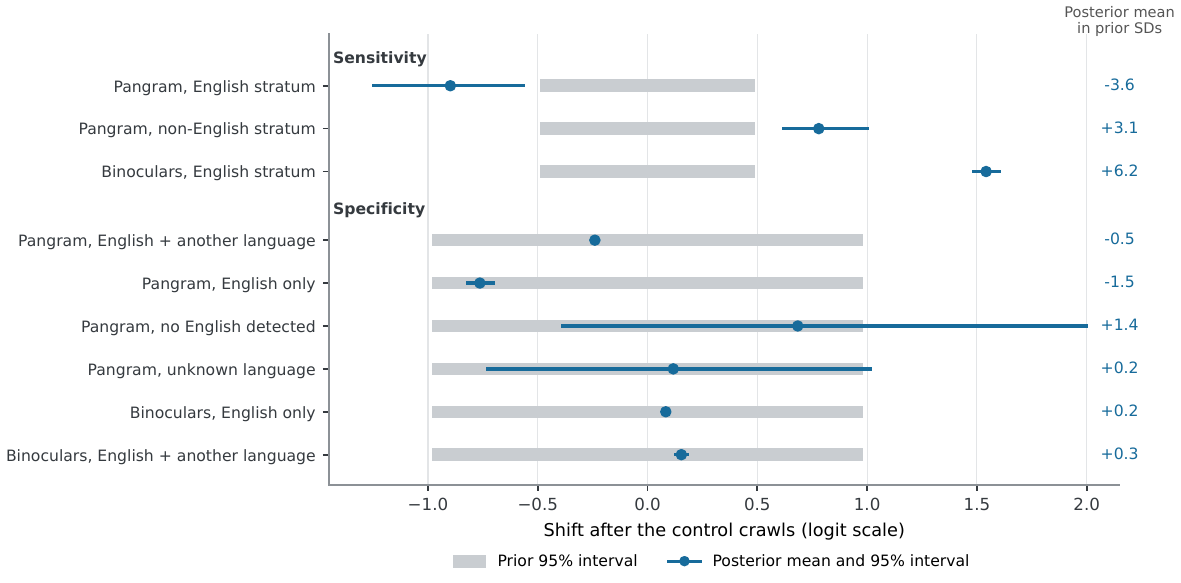} \caption{Prior and posterior of the shifts in sensitivity and specificity after the control crawls (logit scale) in the reported model. The grey bar is the central 95\% prior interval, and the blue point and bar are the posterior mean and 95\% credible interval. The right column gives the posterior mean in units of the prior standard deviation.}
  \label{fig:drift_prior}
\end{figure}

\begin{figure}[t]
\centering \includegraphics[width=0.95\linewidth]{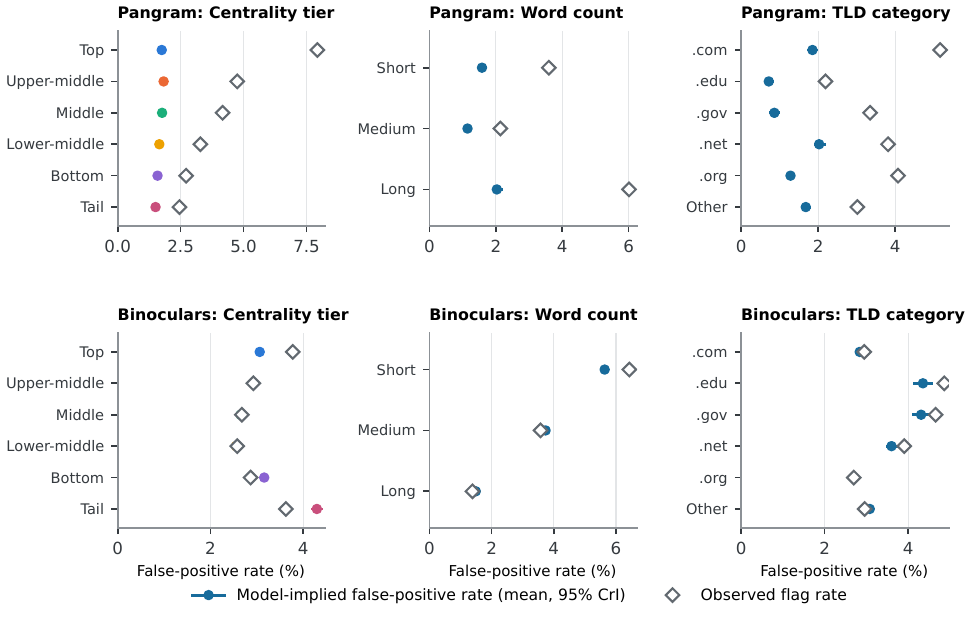} \caption{False-positive rate implied by the control-period specificity, by centrality tier, word-count level and TLD category, for Pangram (top) and Binoculars (bottom). Dots are the posterior mean and 95\% credible interval of the average over the English-only pages of the control crawls in each category, and diamonds are the observed flag rates of those pages.}
  \label{fig:control_sp}
\end{figure}

Figure~\ref{fig:drift_prior} shows the posterior of the shifts in the reported model against their priors: the shifts of the three sensitivities lie well outside the 95\% prior intervals, and the posterior means of the specificity shifts lie within them. For Binoculars the posterior mean of the sensitivity shift is about six prior standard deviations from zero, so the data, not the prior, determine its sensitivity after the control period. Figure~\ref{fig:control_sp} shows the false-positive rates implied by the specificity model. For Binoculars they reproduce the observed flag rates in the control crawls within each category, as expected if nearly all flags are false positives. For Pangram the observed flag rates are higher than the false-positive rates in every category, and the model attributes the difference to prevalence.

The constrained fit is worse by 544 in WAIC (Table~\ref{tab:waic}), and it changes some of the estimates (Tables~\ref{tab:sensitivity}, \ref{tab:sens_findings} and~\ref{tab:sens_crossing}). Prevalence in August 2026 is 22.6\% (21.9--23.3\%) instead of 23.6\%, English-only prevalence is unchanged (32.7\% against 32.8\%), and prevalence of pages with no English detected falls from 17.0\% to 14.5\%, so that the English-only advantage is 18.2 rather than 15.8 points, bilingual prevalence exceeds no-English prevalence by 5.4 rather than 3.2 points, and the crossing moves from February 2026 to October 2025. The change in the growth gap is the same (3.0 points per year). The posterior drifts shift as well: the logit drift of Binoculars' sensitivity falls from 1.54 to 1.03, that of Pangram's sensitivity for non-English pages rises from 0.78 to 1.26, and that of Pangram's specificity for pages with no English detected rises from 0.68 to 2.28, with its posterior standard deviation falling from 0.60 to 0.28. Binoculars is applied to English-stratum pages only, so the change in the non-English estimate must reach it through the structure shared across strata (the intercept, the cell effects and the temporal terms) and through the Pangram drifts, and we have not traced it further. All qualitative findings of Section~\ref{sec:sens_findings} hold under the constraint.

We do not impose the constraint in the reported model, for four reasons. First, the mirror mode that it is designed to exclude is not a concern here: the specificity prior is centered at 92\%, the control crawls precede the generative models so their flags are predominantly false positives, and Pangram, whose flags drive the estimates, has an index far from zero without the constraint. Second, for Binoculars the constraint does not remove an implausible mode but moves an estimate at the boundary: the unconstrained posterior has $J$ slightly below zero in the least specific cell and positive in every draw for 68\% of the pages, so Binoculars is close to uninformative in either fit, and the constraint raises the control-period sensitivity from 4.6\% to 7.6\% to satisfy a property of its specificities, although both values are far below the benchmark rates. Third, because the bound is the minimum of the cell-level specificities, the constraint makes the sensitivity a non-smooth function of the specificity effects, so that the flag rates of the least specific cell determine a parameter that is otherwise informed by the benchmark corpora and the webpages. Fourth, the data prefer the unconstrained model (WAIC difference of 544; the log predictive density of the webpages is lower by 274). We report the constrained fit as a sensitivity result. Its convergence is described in Appendix~\ref{sec:model_diagnostics}.

\subsection{Convergence Diagnostics}\label{sec:model_diagnostics}

All seventeen fits satisfy most of the usual convergence checks (Table~\ref{tab:diagnostics}): none has divergent transitions, transitions at the maximum tree depth or a low energy Bayesian fraction of missing information, the largest split-$\widehat{R}$ is 1.011 and the smallest bulk effective sample size is 464. The largest $\widehat{R}$, in the fit with $R = 6$ basis functions, is marginally above the cutoff of 1.01 that is often used, and belongs to the derived prevalence of English-only pages in the first crawl. In two fits the smallest tail effective sample size is below 400: 215 in the fit with full specificity effects and interactions and 164 in the fit with $R = 10$ basis functions. Both belong to derived prevalences of non-English pages in early crawls, not to model parameters, and are 736 or more in all other fits.

Table~\ref{tab:diagnostics} also reports the Youden index of Section~\ref{sec:calibration-prior}, evaluated at the control-period calibration in the least specific observed cell of each stratum. For Pangram its 95\% credible interval is above zero in every fit except the fit without a temporal effect, with posterior means between 0.55 and 0.61 for english-stratum pages and between 0.35 and 0.39 for non-english pages. In the fit without a temporal effect it is $-0.39$ for english-stratum pages and 0.28 for non-english pages. For Binoculars it is negative in every draw of every fit that allows drift and does not impose the constraint, $-0.06$ with full specificity effects (including the fits with other numbers of basis functions), $-0.08$ without a temporal effect and $-0.01$ with reduced specificity effects, and positive (0.10) only in the fit without drift; in the fit that imposes a positive index it is $0.00$ (0.00, 0.00) by construction, because the constraint binds (Appendix~\ref{sec:youden_constraint}). Computed in every observed cell (206 English-stratum and 188 non-English cells), the index of Pangram is positive in every cell and every draw in all fits except the fit without a temporal effect, where it is negative in every cell of english-stratum pages (from $-0.39$ to 0.00). For Binoculars it ranges from $-0.06$ to $0.04$ across cells in the reported model, with a page-weighted mean of $0.01$, and is positive in every draw for 68\% of pages (the same in the fits with other $R$, 76\% with reduced specificity effects, 49\% without a temporal effect, and 100\% without drift, where it ranges from 0.10 to 0.14).

The fit that imposes a positive Youden index mixes as well as the reported model by the usual measures: its largest $\widehat{R}$ is 1.004 (1.005 in the reported model), its smallest bulk effective sample size is 813 (880) and its smallest tail effective sample size is 1,399 (736). The slack of Binoculars, which sits at its boundary, has bulk and tail effective sample sizes of 9,322 and 4,333. The cost is in the sampler. The mean step size falls from 0.011 to 0.008, the mean number of leapfrog steps per iteration rises by a quarter, from 431 to 541, the mean tree depth rises from 8.5 to 9.0, and 1.3\% of the transitions reach a tree depth of at least 10 (0.04\% in the reported model), with a maximum of 11 against a limit of 12. There are no divergent transitions and no transitions at the maximum tree depth. We attribute the longer trajectories to the boundary and to the minimum over cells in the bound (Appendix~\ref{sec:youden_constraint}).
\begin{table}[t]
\centering
\caption{Convergence diagnostics and Youden index by fit. Draws are the posterior draws over four chains. $\widehat{R}$ is the largest rank-normalized split-$\widehat{R}$, and ESS the smallest bulk and tail effective sample size, over all parameters and derived quantities. $J$ is the posterior mean (95\% credible interval) of the Youden index of the control-period calibration in the least specific observed cell of the stratum, for Pangram in english-stratum and non-English pages and for Binoculars in English-stratum pages (the constraint binds for Binoculars in the fit with a positive Youden index). No fit has divergent transitions, transitions at the maximum tree depth or a low energy Bayesian fraction of missing information.}
\label{tab:diagnostics}
\footnotesize
\resizebox{0.95\textwidth}{!}{%
\setlength{\tabcolsep}{3pt}
\begin{tabular}{p{4.7cm}*{4}{>{\centering\arraybackslash}p{1.1cm}}*{3}{>{\centering\arraybackslash}p{2cm}}}
\toprule
Specification & Draws & $\widehat{R}$ & ESS bulk & ESS tail & $J$, Pangram, English & $J$, Pangram, non-English & $J$, \mbox{Binoculars}, English \\
\midrule
\multirow{2}{4.7cm}{Reported model (full specificity effects)} & \multirow{2}{*}{8{,}000} & \multirow{2}{*}{1.005} & \multirow{2}{*}{880} & \multirow{2}{*}{736} & 0.58 & 0.37 & $-$0.06 \\
 & & & & & {\scriptsize \mbox{(0.57, 0.59)}} & {\scriptsize \mbox{(0.35, 0.39)}} & {\scriptsize \mbox{($-$0.06, $-$0.05)}} \\[4pt]
\multirow{2}{4.7cm}{Full specificity effects, Youden constraint} & \multirow{2}{*}{8{,}000} & \multirow{2}{*}{1.004} & \multirow{2}{*}{813} & \multirow{2}{*}{1{,}399} & 0.58 & 0.36 & 0.00 \\ & & & & & {\scriptsize \mbox{(0.57, 0.59)}} & {\scriptsize \mbox{(0.34, 0.37)}} & {\scriptsize \mbox{(0.00, 0.00)}} \\[4pt]
\multirow{2}{4.7cm}{Full specificity effects, with interactions} & \multirow{2}{*}{8{,}000} & \multirow{2}{*}{1.007} & \multirow{2}{*}{484} & \multirow{2}{*}{215} & 0.57 & 0.35 & $-$0.06 \\
 & & & & & {\scriptsize \mbox{(0.56, 0.58)}} & {\scriptsize \mbox{(0.33, 0.37)}} & {\scriptsize \mbox{($-$0.06, $-$0.05)}} \\[4pt]
\multirow{2}{4.7cm}{Reduced specificity effects} & \multirow{2}{*}{8{,}000} & \multirow{2}{*}{1.008} & \multirow{2}{*}{1{,}057} & \multirow{2}{*}{1{,}830} & 0.60 & 0.36 & $-$0.01 \\
 & & & & & {\scriptsize \mbox{(0.59, 0.61)}} & {\scriptsize \mbox{(0.34, 0.38)}} & {\scriptsize \mbox{($-$0.01, $-$0.01)}} \\[4pt]
\multirow{2}{4.7cm}{Reduced specificity effects, $s_{\delta, \Sp} = 0.25$} & \multirow{2}{*}{8{,}000} & \multirow{2}{*}{1.003} & \multirow{2}{*}{1{,}044} & \multirow{2}{*}{1{,}721} & 0.60 & 0.37 & $-$0.01 \\
 & & & & & {\scriptsize \mbox{(0.58, 0.61)}} & {\scriptsize \mbox{(0.35, 0.38)}} & {\scriptsize \mbox{($-$0.01, $-$0.01)}} \\[4pt]
\multirow{2}{4.7cm}{Reduced specificity effects, $s_{\delta, \Sp} = 0.75$} & \multirow{2}{*}{7{,}000} & \multirow{2}{*}{1.007} & \multirow{2}{*}{868} & \multirow{2}{*}{1{,}191} & 0.60 & 0.36 & $-$0.01 \\
 & & & & & {\scriptsize \mbox{(0.59, 0.61)}} & {\scriptsize \mbox{(0.34, 0.38)}} & {\scriptsize \mbox{($-$0.01, $-$0.01)}} \\[4pt]
\multirow{2}{4.7cm}{Reduced specificity effects, $s_{\delta, \Se} = 0.1$} & \multirow{2}{*}{8{,}000} & \multirow{2}{*}{1.004} & \multirow{2}{*}{1{,}038} & \multirow{2}{*}{1{,}730} & 0.58 & 0.37 & $-$0.01 \\
 & & & & & {\scriptsize \mbox{(0.57, 0.59)}} & {\scriptsize \mbox{(0.35, 0.38)}} & {\scriptsize \mbox{($-$0.01, $-$0.00)}} \\[4pt]
\multirow{2}{4.7cm}{Reduced specificity effects, $s_{\delta, \Se} = 0.5$} & \multirow{2}{*}{8{,}000} & \multirow{2}{*}{1.005} & \multirow{2}{*}{1{,}206} & \multirow{2}{*}{2{,}118} & 0.61 & 0.36 & $-$0.01 \\
 & & & & & {\scriptsize \mbox{(0.60, 0.62)}} & {\scriptsize \mbox{(0.35, 0.38)}} & {\scriptsize \mbox{($-$0.01, $-$0.01)}} \\[4pt]
\multirow{2}{4.7cm}{Reduced specificity effects, $\varepsilon_{\mu} = 0.05$} & \multirow{2}{*}{8{,}000} & \multirow{2}{*}{1.009} & \multirow{2}{*}{779} & \multirow{2}{*}{1{,}473} & 0.60 & 0.36 & $-$0.01 \\
 & & & & & {\scriptsize \mbox{(0.59, 0.61)}} & {\scriptsize \mbox{(0.34, 0.38)}} & {\scriptsize \mbox{($-$0.01, $-$0.01)}} \\[4pt]
\multirow{2}{4.7cm}{Reduced specificity effects, $s_{\mu} = 2$} & \multirow{2}{*}{8{,}000} & \multirow{2}{*}{1.008} & \multirow{2}{*}{925} & \multirow{2}{*}{1{,}579} & 0.60 & 0.36 & $-$0.01 \\
 & & & & & {\scriptsize \mbox{(0.59, 0.61)}} & {\scriptsize \mbox{(0.34, 0.38)}} & {\scriptsize \mbox{($-$0.01, $-$0.01)}} \\[4pt]
\multirow{2}{4.7cm}{Reduced specificity effects, with interactions} & \multirow{2}{*}{8{,}000} & \multirow{2}{*}{1.004} & \multirow{2}{*}{724} & \multirow{2}{*}{1{,}462} & 0.60 & 0.35 & $-$0.01 \\
 & & & & & {\scriptsize \mbox{(0.58, 0.61)}} & {\scriptsize \mbox{(0.33, 0.37)}} & {\scriptsize \mbox{($-$0.01, $-$0.01)}} \\[4pt]
\multirow{2}{4.7cm}{Reduced specificity effects, no drift after the control crawls} & \multirow{2}{*}{8{,}000} & \multirow{2}{*}{1.005} & \multirow{2}{*}{904} & \multirow{2}{*}{1{,}786} & 0.55 & 0.39 & 0.10 \\
 & & & & & {\scriptsize \mbox{(0.54, 0.56)}} & {\scriptsize \mbox{(0.38, 0.41)}} & {\scriptsize \mbox{(0.10, 0.10)}} \\[4pt]
\multirow{2}{4.7cm}{Full specificity effects, no temporal effect} & \multirow{2}{*}{8{,}000} & \multirow{2}{*}{1.002} & \multirow{2}{*}{2{,}538} & \multirow{2}{*}{3{,}428} & $-$0.39 & 0.28 & $-$0.08 \\
 & & & & & {\scriptsize \mbox{($-$0.39, $-$0.39)}} & {\scriptsize \mbox{(0.27, 0.30)}} & {\scriptsize \mbox{($-$0.08, $-$0.07)}} \\[4pt]
\multirow{2}{4.7cm}{Full specificity effects, $R = 6$ basis functions} & \multirow{2}{*}{8{,}000} & \multirow{2}{*}{1.011} & \multirow{2}{*}{931} & \multirow{2}{*}{1{,}855} & 0.58 & 0.37 & $-$0.06 \\
 & & & & & {\scriptsize \mbox{(0.57, 0.59)}} & {\scriptsize \mbox{(0.35, 0.38)}} & {\scriptsize \mbox{($-$0.06, $-$0.05)}} \\[4pt]
\multirow{2}{4.7cm}{Full specificity effects, $R = 8$ basis functions} & \multirow{2}{*}{8{,}000} & \multirow{2}{*}{1.004} & \multirow{2}{*}{1{,}074} & \multirow{2}{*}{1{,}852} & 0.58 & 0.36 & $-$0.06 \\
 & & & & & {\scriptsize \mbox{(0.57, 0.59)}} & {\scriptsize \mbox{(0.35, 0.38)}} & {\scriptsize \mbox{($-$0.06, $-$0.05)}} \\[4pt]
\multirow{2}{4.7cm}{Full specificity effects, $R = 10$ basis functions} & \multirow{2}{*}{8{,}000} & \multirow{2}{*}{1.010} & \multirow{2}{*}{464} & \multirow{2}{*}{164} & 0.58 & 0.36 & $-$0.06 \\
 & & & & & {\scriptsize \mbox{(0.57, 0.59)}} & {\scriptsize \mbox{(0.34, 0.38)}} & {\scriptsize \mbox{($-$0.06, $-$0.05)}} \\[4pt]
\multirow{2}{4.7cm}{Full specificity effects, $R = 20$ basis functions} & \multirow{2}{*}{8{,}000} & \multirow{2}{*}{1.004} & \multirow{2}{*}{658} & \multirow{2}{*}{1{,}028} & 0.58 & 0.36 & $-$0.06 \\
 & & & & & {\scriptsize \mbox{(0.57, 0.59)}} & {\scriptsize \mbox{(0.34, 0.38)}} & {\scriptsize \mbox{($-$0.06, $-$0.05)}} \\[4pt]
\bottomrule
\end{tabular}}
\end{table}

\end{document}